\documentstyle[11pt,psfig]{article}

\title{\Large AN INTRODUCTION TO COLLECTIVE INTELLIGENCE}

\author{David H. Wolpert\\ 
NASA Ames Research Center\\ 
Moffett Field, CA 94035\\ 
dhw@ptolemy.arc.nasa.gov\\
\and
Kagan Tumer\\ 
NASA Ames Research Center\\ 
Caelum Research \\
Moffett Field, CA 94035\\ 
kagan@ptolemy.arc.nasa.gov
\and
{\tt http://ic.arc.nasa.gov/ic/people/kagan/coin\_pubs.html}\\
Tech Report: NASA-ARC-IC-99-63}

\begin{document}

\maketitle

\begin{abstract}

\nocite{dros98} % drossels emergent complexity from scales \\
\nocite{fivr97} % competitive markov decision processes (book)  \\
\nocite{youn93} % convention in econ \\
\nocite{litt94} 

This paper surveys the emerging science of how to design a ``COllective
INtelligence'' (COIN). A COIN is a large multi-agent system where:
 
\noindent
i) There is little to no centralized communication or control.

\noindent
ii) There is a provided world utility function that rates the
possible histories of the full system.

\noindent

In particular, we are interested in COINs in which each agent runs a
reinforcement learning (RL) algorithm.  The conventional approach to
designing large distributed systems to optimize a world utility does
not use agents running RL algorithms. Rather, that approach begins
with explicit modeling of the dynamics of the overall system, followed
by detailed hand-tuning of the interactions between the components to
ensure that they ``cooperate'' as far as the world utility is
concerned. This approach is labor-intensive, often results in highly
nonrobust systems, and usually results in design techniques that have
limited applicability.

\noindent
In contrast, we wish to solve the COIN
design problem implicitly, via the ``adaptive'' character of the RL
algorithms of each of the agents. This approach introduces an
entirely new, profound design problem: Assuming the RL algorithms are
able to achieve high rewards, what reward functions for the individual
agents will, when pursued by those agents, result in high world
utility? In other words, what reward functions will best ensure that we do
not have phenomena like the tragedy of the commons, Braess's paradox,
or the liquidity trap?

\noindent
Although still very young, research specifically concentrating on the
COIN design problem has already resulted in successes in artificial
domains, in particular in packet-routing, the leader-follower problem,
and in variants of Arthur's El Farol bar problem. It is expected that
as it matures and draws upon other disciplines related to COINs, this
research will greatly expand the range of tasks addressable by human
engineers. Moreover, in addition to drawing on them, such a fully developed science of
COIN design may provide much insight into other already established
scientific fields, such as economics, game theory, and population
biology.  
\end{abstract}

\section{INTRODUCTION}
\label{sec:intro}

Over the past decade or so two separate developments have occurred in
computer science whose intersection promises to open a vast new area
of research, an area extending far beyond the current boundaries of
computer science. The first of these developments is the growing
realization of how useful it would be to be able to control
distributed systems that have little (if any) centralized
communication, and to do so ``adaptively'', with minimal reliance on
detailed knowledge of the system's small-scale dynamical behavior.
The second development is the maturing of the discipline of
reinforcement learning (RL).  This is the branch of machine learning
that is concerned with an agent who periodically receives ``reward''
signals from the environment that partially reflect the value of that
agent's private utility function. The goal of an RL algorithm is to
determine how, using those reward signals, the agent should update its
action policy to maximize its
utility~\cite{kali96,sutt88,wada92}. (Until our detailed discussions
below, we will use the term ``reinforcement learning'' broadly, to
include any algorithm of this sort, including ones that rely on
detailed Bayesian modeling of underlying Markov processes
~\cite{russ98,caka94,frko98}.

Intuitively, one might hope that RL would help us solve
the distributed control problem, since RL is adaptive, and, in
particular, since it is not restricted to domains having sufficient
breadths of communication. However, by itself, conventional
single-agent RL does not provide a means for controlling large,
distributed systems. This is true even if the system $does$ have
centralized communication.  The problem is that the space of possible
action policies for such systems is too big to be searched.  We might
imagine as a variant using a large set of agents, each controlling
only part of the system. Since the individual action spaces of such
agents would be relatively small, we could realistically deploy
conventional RL on each one. However, now we face the central question
of how to map the world utility function concerning the overall system
into private utility functions for each of the agents. In particular,
how should we design those private utility functions so that each
agent can realistically hope to optimize its function, and at the same
time the collective behavior of the agents will optimize the world
utility?

We use the term ``COllective INtelligence'' (COIN) to refer to any
pair of a large, distributed collection of interacting computational
processes among which there is little to no centralized communication
or control, together with a `world utility' function that rates the
possible dynamic histories of the collection.  The central COIN design
problem we consider arises when the computational processes run RL
algorithms: How, without any detailed modeling of the overall system,
can one set the utility functions for the RL algorithms in a COIN to
have the overall dynamics reliably and robustly achieve large values
of the provided world utility? The benefits of an answer to this
question would extend beyond the many branches of computer science,
having major ramifications for many other sciences as well.
Section~\ref{sec:back} discusses some of those benefits.
Section~\ref{sec:lit} reviews previous work that has bearing on the
COIN design problem. Section~\ref{sec:math} section constitutes the
core of this chapter. It presents a quick outline of a promising
mathematical framework for addressing this problem in its most general
form, and then experimental illustrations of the prescriptions of that
framework. Throughout, we will use italics for emphasis, single quotes
for informally defined terms, and double quotes to delineate
colloquial terminology.

\section{Background}
\label{sec:back}
There are many design problems that involve distributed computational
systems where there are strong restrictions on centralized
communication (``we can't all talk''); or there is communication with a
central processor, but that processor is not sufficiently powerful to
determine how to control the entire system (``we aren't smart enough'');
or the processor is powerful enough in principle, but it is not clear
what algorithm it could run by itself that would effectively control
the entire system (``we don't know what to think''). Just a few of the
potential examples include:

i) Designing a control system for constellations of communication
satellites or for constellations of planetary exploration vehicles
(world utility in the latter case being some measure of quality of
scientific data collected);

ii) Designing a control system for routing over a communication
network (world utility being some aggregate quality of service
measure)

iii) Construction of parallel algorithms for solving numerical
optimization problems (the optimization problem itself constituting
the world utility);

iv) Vehicular traffic control, {\em e.g.}, air traffic control, or
high-occupancy toll-lanes for automobiles. (In these problems the
individual agents are humans and the associated utility functions must
be of a constrained form, reflecting the relatively inflexible kinds
of preferences humans possess.);

v) Routing over a power grid;

vi) Control of a large, distributed chemical plant;

vii) Control of the elements of an amorphous computer;

viii) Control of the elements of a `noisy' phased array radar;

ix) Compute-serving over an information grid. 

Such systems may be best controlled with an artificial COIN. However,
the potential usefulness of deeper understanding of how to tackle the
COIN design problem extends far beyond such engineering
concerns. That's because the COIN design problem is an inverse
problem, whereas essentially all of the scientific fields that are
concerned with naturally-occurring distributed systems analyze them
purely as a ``forward problem.'' That is, those fields analyze what
global behavior would arise from provided local dynamical laws, rather
than grapple with the inverse problem of how to configure those laws
to induce desired global behavior.  (Indeed, the COIN design problem
could almost be defined as decentralized adaptive control theory for
massively distributed stochastic environments.) It seems plausible that
the insights garnered from understanding the inverse problem would
provide a trenchant novel perspective on those fields. Just as
tackling the inverse problem in the design of steam engines led to the
first true understanding of the macroscopic properties of physical
bodes (aka thermodynamics), so may the cracking of the COIN design
problem may improve our understanding of many naturally-occurring
COINs. In addition, although the focuses of those other fields are not
on the COIN design problem, in that they are related to the COIN
design problem, that problem may be able to serve as a ``touchstone''
for all those fields. This may then reveal novel connections between
the fields.

As an example of how understanding the COIN design problem may provide
a novel perspective on other fields, consider countries with
capitalist human economies. Although there is no intrinsic world
utility in such systems, they can still be viewed from the perspective
of COINs, as naturally occurring COINs. For example, one can declare
world utility to be a time average of the Gross Domestic Product (GDP)
of the country in question.  (World utility per se is not a
construction internal to a human economy, but rather something defined
from the outside.) The reward functions for the human agents in this
example could then be the achievements of their personal goals
(usually involving personal wealth to some degree).

Now in general, to achieve high world utility in a COIN it is
necessary to avoid having the agents work at cross-purposes. Otherwise
the system is vulnerable to economic phenomena like the Tragedy of the
Commons (TOC), in which individual avarice works to lower world
utility~\cite{hard68}, or the liquidity trap, where behavior that
helps the entire system when employed by some agents results in poor
global behavior when employed by all agents ~\cite{krug94}.  One way
to avoid such phenomena is by modifying the agents' utility
functions. In the context of capitalist economies, this kind of effect
can be achieved via punitive legislation that modifies the rewards the
agents receive for engaging in certain kinds of activity.  A real
world example of an attempt to make just such a modification was the
creation of anti-trust regulations designed to prevent monopolistic
practices.

In designing a COIN we usually have more freedom than anti-trust
regulators though, in that there is no base-line ``organic'' private
utility function over which we must superimpose legislation-like
incentives. Rather, the entire ``psychology'' of the individual agents
is at our disposal when designing a COIN.  This obviates the need for
honesty-elicitation (`incentive compatible') mechanisms, like
auctions, which form a central component of conventional economics.
Accordingly, COINs can differ in certain crucial respects from human
economies. The precise differences --- the subject of current research
--- seem likely to present many insights into the functioning of
economic structures like anti-trust regulators.

To continue with this example, consider the usefulness, as far as the
world utility is concerned, of having (commodity, or especially fiat)
money in the COIN.  Formally, from a COIN perspective, the use of
`money' for trading between agents constitutes a particular class of
couplings between the states and utility functions of the various
agents. For example, if one agent's `bank account' variable goes up in
a `trade' with another agent, then a corresponding `bank account'
variable in that other agent must decrease to compensate. In addition
to this coupling between the agents' states, there is also a coupling
between their utilities, if one assume that both agents will prefer to
have more money rather than less, everything else being equal. However
one might formally define such a `money' structure, we can consider
what happens if it does (or does not) obtain for an arbitrary
dynamical system, in the context of an arbitrary world utility. For
some such dynamical systems and world utilities, a money structure
will improve the value of that world utility. But for the same
dynamics, the use of a money structure will simultaneously induce {\it
low levels} of other world utilities (a trivial example being a world
utility that equals the negative of the first one). This raises a host
of questions, like how to formally specify the most general set of
world utilities that benefits significantly from using money-based
private utility functions. If one is provided a world utility that is
not a member of that set, then an ``economics-like'' configuration of
the system is likely to result in poor performance. Such a
characterization of how and when money helps improve world utilities of
various sorts might have important implications for conventional human
economics, especially when one chooses world utility to be one of the
more popular choices for social welfare function. (See
~\cite{stst98,duoc98} and references therein for some of the standard
economics work that is most relevant to this issue.)

There are many other scientific fields that are currently under
investigation from a COIN-design perspective. Some of them are, like
economics, part of (or at least closely related to) the social
sciences. These fields typically involve RL algorithms under the guise
of human agents. An example of such a field is game theory, especially
game theory of bounded rational players.  As illustrated in our money
example, viewing such systems from the perspective of a non-endogenous
world utility, {\it i.e.}, from a COIN-design perspective, holds the
potential for providing novel insight into them. (In the case of game
theory, it holds the potential for leading to deeper understanding of
many-player inverse stochastic game theory.)

However there are other scientific fields that might benefit from a
COIN-design perspective even though they study systems that don't even
involve RL algorithms. The idea here is that if we viewed such systems
from an ``artificial'' teleological perspective, both in concentrating
on a non-endogenous world utility and in casting the nodal elements of
the system as RL algorithms, we could learn a lot about the form of
the `design space' in which such systems live. (Just as in economics,
where the individual nodal elements {\it are} RL algorithms,
investigating the system using an externally imposed world utility
might lead to insight.) Examples here are ecosystems (individual
genes, individuals, or species being the nodal elements) and cells
(individual organelles in Eukaryotes being the nodal elements). In
both cases, the world utility could involve robustness of the desired
equilibrium against external perturbation, efficient exploitation of
free energy in the environment, etc.

\section{Review of Literature Related to COINs}
\label{sec:lit}

The following list elaborates what we mean by a COIN:

1) There are many processors running concurrently, performing actions
that affect one another's behavior.

2) There is little to no centralized personalized communication, 
{\em i.e.}, little to no behavior in which a small subset of the
processors not only
communicates with all the other processors, but communicates
differently with each one of those other processors. Any single
processor's ``broadcasting'' the same information to all other
processors is not precluded.

3) There is little to no centralized personalized control, {\em i.e.},
little to no behavior in which a small subset of the processors not only
controls all the other processors, but controls each one of those
other processors differently. ``Broadcasting'' the same control signal
to all other processors is not precluded.

4) There is a well-specified task, typically in the form of
extremizing a utility function, that concerns the behavior of the
entire distributed system. So we are confronted with the inverse
problem of how to configure the system to achieve the task.

The following elements characterize the sorts of approaches to COIN
design we are concerned with here: 

5) The approach for tackling (4) is scalable to very large numbers of
processors.

6) The approach for tackling (4) is very broadly applicable. In
particular, it can work when little (if any) ``broadcasting'' as in
(2) and (3) is possible.

7) The approach for tackling (4) involves little to no hand-tailoring.

8) The approach for tackling (4) is robust and adaptive, with minimal
need to ``get the details exactly right or else,'' as far as the
stochastic dynamics of the system is concerned.

9) The individual processors are running RL algorithms. Unlike the
other elements of this list, this one is not an {\it a priori}
engineering necessity. Rather, it is a reflection of the fact that RL
algorithms are currently the best-understood and most mature
technology for addressing the points (8) and (9).

There are many approaches to COIN design that do not have every one of
those features.  These approaches constitute part of the overall field
of COIN design. As discussed below though, not having every feature in
our list, no single one of those approaches can be extended to cover
the entire breadth of the field of COIN design. (This is not too
surprising, since those approaches are parts of fields whose focus is
not the COIN design problem per se.) 

The rest of this section consists of brief presentations of some of
these approaches, and in particular characterizes them in terms of our
list of nine characteristics of COINs and of our desiredata for their
design. Of the approaches we discuss, at present it is probably the
ones in Artificial Intelligence and Machine Learning that are most
directly applicable to COIN design. However it is fairly clear how to
exploit those approaches for COIN design, and in that sense relatively
little needs to be said about them. In contrast, as currently
employed, the toolsets in the social sciences are not as immediately
applicable to COIN design. However, it seems likely that there is more
yet to be discovered about how to exploit them for COIN design.
Accordingly, we devote more space to those social science-based
approaches here.

We present an approach that holds promise for covering all nine of our
desired features in Section~\ref{sec:math}.

\subsection{AI and Machine Learning}

There is an extensive body of work in AI and machine learning that is
related to COIN design. Indeed, one of the most famous speculative
works in the field can be viewed as an argument that AI should be
approached as a COIN design problem ~\cite{mins88}. Much work of a
more concrete nature is also closely related to the problem of COIN
design. 

\subsubsection{Reinforcement Learning}
\label{sec:control}
As discussed in the introduction, the maturing field of reinforcement
learning provides a much needed tool for the  types of problems 
addressed by COINs.
Because RL generally provides 
model-free\footnote{There exist some model-based variants of traditional
RL. See for example ~\cite{atke98}.}
and ``online'' learning features, it is ideally suited for
the distributed environment where a ``teacher'' is not available
and the agents need to learn successful strategies based on
``rewards'' and ``penalties'' they receive from the overall system
at various intervals. It is even possible 
for the learners to use those rewards to modify {\em how} they 
learn~\cite{sczh97a,sczh97b}.

Although work on RL dates back to Samuel's checker
player~\cite{samu59}, relatively recent theoretical ~\cite{wada92} and
empirical results~\cite{crba96,tess92} have made RL one of the most
active areas in machine learning. Many problems ranging from
controlling a robot's gait to controlling a chemical plant to
allocating constrained resource have been addressed with considerable
success using RL~\cite{haba97,hugr97,mubo98,prwu97,zhdi91}.  In
particular, the RL algorithms $TD(\lambda)$ (which rates potential
states based on a {\em value function})~\cite{sutt88} and
$Q$--learning (which rates action-state pairs)~\cite{wada92} have been
investigated extensively.  A detailed investigation of RL is available
in~\cite{kali96,suba98,wada92}.

Although powerful and widely applicable, solitary RL algorithms will
not perform well on large distributed heterogeneous problems in
general.  This is due to the very big size of the action-policy space
for such problems.  In addition, without centralized communication and
control, how a solitary RL algorithm could run the full system at all,
poorly or well, becomes a major concern.\footnote{One possible
solution would be to run the RL off-line on a simulation of the full
system and then convey the results to the components of the system at
the price of a single centralized initialization ({\em e.g.},
~\cite{mola98}). In general though, this approach will suffer from
being extremely dependent on ``getting the details right'' in the
simulation.} For these reasons, it is natural to consider deploying
many RL algorithms rather than a single one for these large
distributed problems.  We will discuss the coordination issues such an
approach raises in conjunction with multi-agent systems in
Section~\ref{sec:mas} and with learnability in COINs in
Section~\ref{sec:math}.

\subsubsection{Distributed Artificial Intelligence} 
The field of Distributed Artificial Intelligence (DAI) has arisen as
more and more traditional Artificial Intelligence (AI) tasks have migrated
toward parallel implementation.  The most direct approach to such
implementations is to directly parallelize AI production systems or the
underlying programming
languages ~\cite{forg82,rikn91}.  An alternative and more challenging
approach is to use distributed computing, where not only are the individual
reasoning, planning and scheduling AI tasks parallelized, but there are
{\em different modules} with different such tasks, concurrently working
toward a common goal~\cite{huhn87,iygh95,legh99}.

In a DAI, one needs to ensure that the task has been modularized in a
way that improves efficiency. Unfortunately, this usually requires a
central controller whose purpose is to allocate tasks and process the
associated results. Moreover, designing that controller in a
traditional AI fashion often results in brittle
solutions. Accordingly, recently there has been a move toward both
more autonomous modules and fewer restrictions on the interactions
among the modules~\cite{sale95}.

Despite this evolution, DAI maintains the traditional AI concern with
a pre-fixed set of {\em particular} aspects of intelligent behavior
({\em e.g.} reasoning, understanding, learning etc.) rather than on their
{\em cumulative} character. As the idea that intelligence may have more
to do with the interaction among components started to take 
shape~\cite{broo91a,broo91b}, focus shifted to concepts ({\em e.g.}, 
multi-agent systems) that better incorporated that idea~\cite{jesy98}.

\subsubsection{Multi-Agent Systems}
\label{sec:mas}
The field of Multi-Agent Systems (MAS) is concerned with the
interactions among the members of such a set of agents
~\cite{brad97,goro99,jesy98,sen97,syca98}, as well as the inner workings of
each agent in such a set ({\em e.g.}, their learning
algorithms)~\cite{bout99a,bout99b,bosh97}.  As in computational ecologies and
computational markets (see below), a well-designed MAS is one that
achieves a global task through the actions of its components.  The associated
design steps involve~\cite{jesy98}:
\begin{enumerate}
\item{}
Decomposing a global task into distributable subcomponents,
yielding tractable tasks for each agent;
\item{}
Establishing communication channels that provide sufficient
information to each of the agents for it to achieve its task, but are
not too unwieldly for the overall system to sustain; and
\item{}
Coordinating the agents in a way that ensures that they
cooperate on the global task, or at the very least does not allow them to pursue
conflicting strategies in trying to achieve their tasks.
\end{enumerate}

Step (3) is rarely trivial; one of the main difficulties encountered
in MAS design is that agents act selfishly and artificial cooperation
structures have to be imposed on their behavior to enforce
cooperation~\cite{axel97}. An active area of research, which holds
promise for addressing parts the COIN design problem, is to determine
how selfish agents' ``incentives'' have to be engineered in order to
avoid the tragedy of the commons (TOC) ~\cite{shta97}. (This work
draws on the economics literature, which we review separately below.)
When simply providing the right incentives is not sufficient, one can
resort to strategies that actively induce agents to cooperate rather
than act selfishly. In such cases coordination~\cite{sese99},
negotiations~\cite{krau97}, coalition
formation~\cite{sala98,sale97,zlro99} or contracting~\cite{ansa98}
among agents may be needed to ensure that they do not work at cross
purposes.

Unfortunately, all of these approaches share with DAI and its
offshoots the problem of relying excessively on hand-tailoring, and
therefore being difficult to scale and often nonrobust. In addition,
except as noted in the next subsection, they involve no RL, and
therefore the constituent computational elements are usually not as
adaptive and robust as we would like.

\subsubsection{Reinforcement Learning-Based Multi-Agent Systems}
Because it
neither requires explicit modeling of the environment nor having
a ``teacher'' that provides the ``correct'' actions, the approach of
having the individual agents in a MAS use
RL is well-suited for MAS's
deployed in domains where one has little knowledge about the
environment and/or other agents.
There are two main approaches to designing such MAS's:  \\
(i) One has `solipsistic agents' that don't know about each other
and whose RL rewards are given by the performance
of the entire system (so the
joint actions of all other agents form an ``inanimate background'' 
contributing to the reward signal each agent receives); \\ 
(ii) One has `social agents' that explicitly model each other and
take each others' actions into account. 

\noindent
Both (i) and (ii) can be viewed as ways to (try to) coordinate the
agents in a MAS in a robust fashion.

\noindent
{\bf Solipsistic Agents:}
MAS's with solipsistic agents have been successfully applied to a
multitude of
problems~\cite{crba96,guho97,hohu98,sacr95,scsh95}. Generally, these
schemes use RL algorithms similar to those discussed in
Section~\ref{sec:control}.  However much of this work lacks a
well-defined global task or broad applicability ({\em e.g.},
~\cite{sacr95}).  More generally, none of the work with solipsistic
agents scales well. (As illustrated in our experiments on the ``bar
problem'', recounted below.) The problem is that each agent must be able to
discern the effect of its actions on the overall performance of the
system, since that performance constitutes its reward signal.  As the
number of agents increases though, the effects of any one agent's
actions (signal) will be swamped by the effects of other agents
(noise), making the agent unable to learn well, if at all. (See the
discussion below on learnability.) In addition, of course, solipsistic
agents cannot be used in situations lacking centralized calculation
and broadcast of the single global reward signal.

\noindent {\bf Social agents:}
MAS's whose agents take the actions of other agents into account
synthesize RL with game theoretic concepts ({\em e.g.}, Nash
equilibrium). They do this to try to ensure that the overall system
both moves toward achieving the overall global goal and avoids often
deleterious oscillatory
behavior~\cite{clbo98,fule93,huwe96,huwe98b,huwe98a}. To that end, the agents
incorporate internal mechanisms that actively model the behavior of
other agents. In Section~\ref{sec:bar}, we discuss a situation where
such modeling is necessarily self-defeating. More generally, this
approach usually involves extensive hand-tailoring for the problem at
hand.

\subsection{Social Science--Inspired Systems}
\label{sec:economics}

Some human economies provides examples of naturally occurring systems
that can be viewed as a (more or less) well-performing COIN. The field
of economics provides much more though.  Both empirical economics
({\em e.g.}, economic history, experimental economics) and theoretical
economics ({\em e.g.}, general equilibrium theory~\cite{arde54},
theory of optimal taxation~\cite{mirr74}) provide a rich literature on
strategic situations where many parties interact. In fact, much of the
entire field of economics can be viewed as concerning how to maximize
certain constrained kinds of world utilities, when there are certain
(very strong) restrictions on the individual agents and their
interactions, and in particular when we have limited freedom in
setting either the utility functions of those agents or modifying
their RL algorithms in any other way.

In this section we summarize just two economic concepts, both of which
are very closely related to COINs, in that they deal with how a large
number of interacting agents can function in a stable and efficient
manner: general equilibrium theory and mechanism design. We then
discuss general attempts to apply those concepts to distributed
computational problems. We follow this with a discussion of game
theory, and then present a particular celebrated toy-world problem
that involves many of these issues.

\subsubsection{General Equilibrium Theory} 

Often the first version of ``equilibrium'' that one encounters in
economics is that of supply and demand in single markets: the price of
the market's good is determined by where the supply and demand curves
for that good intersect.  In cases where there is interaction among
multiple markets however, even when there is no production but only
trading, one cannot simply determine the price of
each market's good individually, as both the supply and demand for
each good depends on the supply/demand of other goods. Considering the
price fluctuations across markets leads to the concept of `general
equilibrium', where prices for each good are determined in such a way
to ensure that all markets `clear'~\cite{arde54,star97}.  Intuitively,
this means that prices are set so the total supply of each good is
equal to the demand for that good.\footnote{More formally, each agent's
utility is a function of that agent's allotment of all the possible
goods. In addition, every good has a price.  (Utility functions are
independent of money.) Therefore, for any set of prices for the goods,
every agent has a `budget', given by their initial allotment of
goods. We pool all the agents' goods together.  In the `tatonnement'
(single step)
version of market clearing, we next allocate the goods back among the
agents in such a way that each agent is given a total value of goods
(as determined by the prices) equal to that agent's budget (as
determined by the prices and by that agent's initial allotment). As a
(formally identical) alternative, we can have a two-step process, in
which first each agent is given funds equal to its budget, and then
each agent decides how to use those funds to purchases goods from the
central pool. In either case, the `market clearing' prices are those
prices for which exactly all of the goods in the pool are reallocated
back among the agents (no ``excess supply''), and for which each agent
views its allocation of goods as optimizing its utility, subject to
its budget and to those prices for the goods (no ``excess
demand''). Similar definitions hold for a `production' rather than
`endowment' economy.}  The existence of such an equilibrium, proven in
~\cite{arde54}, was first postulated by Leon Walras~\cite{walr74}. A
mechanism that calculates the equilibrium ({\em i.e.},
`market-clearing') prices now bears his name: the Walrasian auctioner.

In general, for an arbitrary goal for the overall system, there is no
reason to believe that having markets clear achieves that goal. In
other words, there is no {\it a priori} reason why the general
equilibrium point should maximize one's provided world utility
function. However, consider the case where one's goal for the overall
system is in fact that the markets clear. In such a context, examine
the case where the interactions of real-world agents will induce the
overall system to adopt the general equilibrium point, so long as
certain broad conditions hold. Then if we can impose those conditions,
we can cause the overall system to behave in the manner we
wish. However general equilibrium theory is not sufficient to
establish those ``broad conditions'', since it says little about
real-world agents. In particular, general equilibrium theory suffers
from having no temporal aspect ({\em i.e.}, no dynamics) and from assuming
that all the agents are perfectly rational.

Another shortcoming of general equilibrium theory as a model of
real-world systems is that despite its concerning prices, it does not
readily accommodate the full concept of money~\cite{frha90}.  Of the
three main roles money plays in an economy (medium of exchange in
trades, store of value for future trades, and unit of account) none
are essential in a general equilibrium setting. The unit of account
aspect is not needed as the bookkeeping is performed by the Walrasian
auctioner. Since the supplies and demands are matched directly there
is no need to facilitate trades, and thus no role for money as a
medium of exchange. And finally, as the system reaches an equilibrium
in one step, through the auctioner, there is no need to store value
for future trading rounds~\cite{osst90}.

The reason that money is not needed can be traced to the fact that
there is an ``overseer'' with global information who guides the
system. If we remove the centralized communication and control exerted
by this overseer, then (as in a real economy) agents will no longer
know the exact details of the overall economy. They will be forced to
makes guesses as in any learning system, and the differences in those
guesses will lead to differences in their actions~\cite{kurz94,kurz96}. 

Such a decentralized learning-based system more closely resembles a
COIN than does a conventional general equilibrium system. In contrast
to general equilibrium systems, the three main roles money plays in a
human economy are crucial to the dynamics of such a decentralized
system~\cite{bano98}. This comports with the important effects in
COINs of having the agents' utility functions involve money (see
Background section above).

\subsubsection{Mechanism Design}

Even if there exists centralized communication so that we aren't
considering a full-blown COIN, if there is no centralized Walras-like
control, it is usually highly non-trivial to induce the overall system
to adopt the General Equilibrium point. One way to try to do so is via
an auction. (This is the approach usually employed in computational
markets --- see below.) Along with optimal taxation and public good
theory~\cite{krpe97}, the design of auctions is the subject of the
field of mechanism design. More generally, mechanism design is
concerned with the incentives that must be applied to any set of
agents that interact and exchange goods~\cite{mirr74,vick61} in order
to get those agents to exhibit desired behavior. Usually that desired
behavior concerns pre-specified utility functions of some sort for
each of the individual agents. In particular, mechanism design is
usually concerned with incentive schemes which induce `(Pareto)
efficient' (or `Pareto optimal') allocations in which no agent can be
made better off without hurting another agent~\cite{fule98,futi91}.

One particularly important type of such an incentive scheme is an
auction. When many agents interact in a common environment often
there needs to be a structure that supports the exchange of goods or
information among those agents. Auctions provide one such
(centralized) structure for managing exchanges of goods.  For example,
in the English auction all the agents come together and `bid' for a
good, and the price of the good is increased until only one bidder
remains, who gets the good in exchange for the resource bid. As
another example, in the Dutch auction the price of a good is decreased
until one buyer is willing to pay the current price.

All auctions perform the same task: match supply and demand.  As such,
auctions are one of the ways in which price equilibration among a set
of interacting agents (perhaps an equilibration approximating general
equilibrium, perhaps not) can be achieved.  However, an auction
mechanism that induces Pareto efficiency does not necessarily maximize some
other world  utility.  For example, in a transaction in an English
auction both the seller and the buyer benefit. They may even have
arrived at an allocation which is efficient. However, in that the
winner may well have been willing to pay more for the good, such an
outcome may confound the goal of the market designer, if that
designer's goal is to maximize revenue. This point is returned to
below, in the context of computational economics.

\subsubsection{Computational Economics}
`Computational economies' are schemes inspired by economics, and more
specifically by general equilibrium theory and mechanism design
theory, for managing the components of a distributed computational
system. They work by having a `computational market', akin to an
auction, guide the interactions among those components.  Such a market
is defined as any structure that allows the components of the system
to exchange information on relative valuation of resources (as in an
auction), establish equilibrium states ({\em e.g.}, determine market
clearing prices) and exchange resources ({\em i.e.}, engage in trades).

Such computational economies can be used to investigate real economies
and biological systems~\cite{blea97,boja97,boja98,keha98}.  They can
also be used to design distributed computational systems. For example,
such computational economies are well-suited to some distributed
resource allocation problems, where each component of the system can
either directly produce the ``goods'' it needs or acquire them through
trades with other components.  Computational markets often allow for
far more heterogeneity in the components than do conventional resource
allocation schemes.  Furthermore, there is both theoretical and
empirical evidence suggesting that such markets are often able to
settle to equilibrium states.  For example, auctions find prices that
satisfy both the seller and the buyer which results in an increase in
the utility of both (else one or the other would not have agreed to
the sale).  Assuming that all parties are free to pursue trading
opportunities, such mechanisms move the system to a point where all
possible bilateral trades that could improve the utility of both
parties are exhausted.

Now restrict attention to the case, implicit in much of computational
market work, with the following characteristics: First, world utility
can be expressed as a monotonically increasing function $F$ where each
argument $i$ of $F$ can in turn be interpreted as the value of a
pre-specified utility function $f_i$ for agent $i$.  Second, each of
those $f_i$ is a function of an $i$-indexed `goods vector' $x_i$ of
the non-perishable goods ``owned'' by agent $i$. The components of
that vector are $x_{i,j}$, and the overall system dynamics is
restricted to conserve the vector $\sum_i x_{i,j}$. (There are also
some other, more technical conditions.)  As an example, the resource
allocation problem can be viewed as concerning such vectors of
``owned'' goods.

Due to the second of our two conditions, one can integrate a
market-clearing mechanism into any system of this sort. Due to the
first condition, since in a market equilibrium with non-perishable
goods no (rational) agent ends up with a value of its utility function
lower than the one it started with, the value of the world utility
function must be higher at equilibrium than it was initially.  In
fact, so long as the individual agents are smart enough to avoid all
trades in which they do not benefit, any computational market can only
improve this kind of world utility, even if it does not achieve the
market equilibrium.  (See the discussion of ``weak triviality''
below.)

This line of reasoning provides one of the main reasons to use
computational markets when they can be applied.  Conversely, it
underscores one of the major limitations of such markets: Starting
with an arbitrary world utility function with arbitrary dynamical
restrictions, it may be quite difficult to cast that function as a
monotonically increasing $F$ taking as arguments a set of agents'
goods-vector-based utilities $f_i$, if we require that those $f_i$ be
well-enough behaved that we can reasonably expect the agents to
optimize them in a market setting.

One example of a computational economy being used for resource
allocation is Huberman and Clearwater's use of a double--blind auction
to solve the complex task of controlling the temperature of a
building.  In this case, each agent (individual temperature
controller) bids to buy or sell cool or warm air. This market
mechanism leads to an equitable temperature distribution in the
system~\cite{hucl95}.  Other domains where market mechanisms were
successfully applied include purchasing memory in an operating
systems~\cite{chha95}, allocating virtual circuits~\cite{feni89},
``stealing'' unused CPU cycles in a network of
computers~\cite{elli75,waho92}, predicting option futures in financial
markets~\cite{poco97}, and numerous scheduling and distributed
resource allocation
problems~\cite{kusi89,lase98,siao96,stao94,well93,well94}.

Computational economics can also be used for tasks not tightly coupled
to resource allocation.  For example, following the work of Maes
~\cite{maes90} and Ferber ~\cite{ferb96}, Baum shows how by using
computational markets a large number of agents can interact and
cooperate to solve a variant of the blocks world
problem~\cite{baum96,baum99}.

Viewed as candidate COINs, all market-based computational economics
fall short in relying on both centralized communication and
centralized control to some degree. Often that reliance is
extreme. For example, the systems investigated by Baum not only have
the centralized control of a market, but in addition have centralized
control of all other non-market aspects of the system. (Indeed, the
market is secondary, in that it is only used to decide which single
expert among a set of candidate experts gets to exert that centralized
control at any given moment). There has
also been doubt cast on how well computational economies perform in
practice ~\cite{tube96}, and they also often require extensive
hand-tailoring in practice.

Finally, return to consideration of a world utility function that is a
monotonically increasing function $f$ whose arguments are the
utilities of the agents. In general, the maximum of such a world
utility function will be a Pareto optimal point. So given the utility
functions of the agents, by considering all such $f$ we map out an
infinite set $S$ of Pareto optimal points that maximize {\it some}
such world utility function. ($S$ is usually infinite even if we only
consider maximizing those world utilities subject to an overall
conservation of goods constraint.) Now the market equilibrium is a
Pareto optimal point, and therefore lies in $S$.  But it is only one
element of $S$. Moreover, it is usually set in full by the utilities
of the agents, in concert with the agents' initial endowments. In
particular, it is independent of the world utility. In general then,
given the utilities of the agents and a world utility $f$, there is no
{\it a priori} reason to believe that the particular element in $S$
picked out by the auction is the point that maximizes that particular
world utility.  This subtlety is rarely addressed in the work on using
computational markets to achieve a global goal. It need not be
uncircumventable however. For example, one obvious idea would be to to
try to distort the agents' {\it perceptions} of their utility
functions and/or initial endowments so that the resultant market
equilibrium has a higher value of the world utility at
hand.\footnote{In fact, the second theorem of welfare economics
~\cite{star97} states that given any world utility such that: i) its
global maximum can be written as a Pareto optimal point for agents'
utilities all of whose level sets are convex; ii) no other maximum of
that world utility is Pareto optimal for those agents' utilities; then
one can always set initial endowments of the agents so that that
Pareto optimal point corresponds to the price-clearing point for those
endowments. Note though that that setting of endowments requires a
centralized process. Moreover, even if we are allowed such a process
to set the endowments, we still may not be able to successfully
exploit this theorem to arrive at the world utility maximum, if we use
markets involving iterative trading with dynamic associated prices,
like those in the real world.  This is because an intermediate trade
with ``incorrect'' prices may have resulted in some particular agent's
having its utility rise beyond the level it has at the point
maximizing world utility, and since that agent will never afterward engage
in a trade that diminishes its utility, the system will never arrive
at the world utility maximum.}

\subsubsection{Perfect Rationality Noncooperative Game Theory}
\label{sec:gt}

Game theory is the branch of mathematics concerned with formalized
versions of ``games'', in the sense of chess, poker, nuclear arms
races, and the like
~\cite{binm92,futi91,felt97,seth98,pal89,lura85,baol82,auha92}. It is
perhaps easiest to describe it by loosely defining some of its
terminology, which we do here and in the next subsection.

The simplest form of a game is that of `non-cooperative single-stage
extensive-form' game, which involves the following
situation: There are two or more agents (called `players' in the
literature), each of which has a pre-specified set of possible actions
that it can follow. (A `finite' game has finite sets of possible
actions for all the players.) In addition, each agent $i$ has a
utility function (also called a `payoff matrix' for finite
games). This maps any `profile' of the action choices of all agents to
an associated utility value for agent $i$. (In a `zero-sum' game, for
every profile, the sum of the payoffs to all the agents is zero.)

The agents choose their actions in a sequence, one after the other.
The structure determining what each agent knows concerning the action
choices of the preceding agents is known as the `information
set'.\footnote{While stochastic choices of actions is central to game
theory, most of the work in the field assumes the information in
information sets is in the form of definite facts, rather than a
probability distribution. Accordingly, there has been relatively
little work incorporating Shannon information theory into the analysis
of information sets.} Games in which each agent knows exactly what the
preceding (`leader') agent did are known as `Stackelberg games'. (A
variant of such a game is considered in our experiments below. See
also ~\cite{kola97}.)

In a `multi-stage' game, after all the agents choose their first
action, each agent is provided some information concerning what the
other agents did. The agent uses this information to choose its next
action. In the usual formulation, each agent gets its payoff at the
end of all of the game's stages.

An agent's `strategy' is the rule it elects to follow mapping the
information it has at each stage of a game to its associated
action. It is a `pure strategy' if it is a deterministic rule. If
instead the agent's action is chosen by randomly sampling from a
distribution, that distribution is known a `mixed strategy'. Note that
an agent's strategy concerns $all$ possible sequences of provided
information, even any that cannot arise due to the strategies of the
other agents.

Any multi-stage extensive-form game can be converted into a `normal
form' game, which is a single-stage game in which each agent is
ignorant of the actions of the other agents, so that all agents choose
their actions ``simultaneously''. This conversion is acieved by having
the ``actions'' of each agent in the normal form game correspond to an
entire strategy in the associated multi-stage extensive-form game.
The payoffs to all the agents in the normal form game for a particular
strategy profile is then given by the associated payoff matrices of
the multi-stage extensive form-game.

A `solution' to a game, or an `equilibrium', is a profile in which
every agent behaves ``rationally''.  This means that every agent's
choice of strategy optimizes its utility subject to a pre-specified
set of conditions. In conventional game theory those conditions
involve, at a minimum, perfect knowledge of the payoff matrices of all
other players, and often also involve specification of what strategies
the other agents adopted and the like.  In particular, a `Nash
equilibrium' is a a profile where each agent has chosen the best
strategy it can, {\it given the choices of the other agents}. A game
may have no Nash equilibria, one equilibrium, or many equilibria in
the space of pure strategies. A
beautiful and seminal theorem due to Nash proves that every game has
at least one Nash equilibrium in the space of mixed strategies
~\cite{nash50}.

There are several different reasons one might expect a game to result
in a Nash equilibrium. One is that it is the point that perfectly
rational Bayesian agents would adopt, assuming the probability
distributions they used to calculate expected payoffs were consistent
with one another ~\cite{auma87,kale93}. A related reason, arising even in a
non-Bayesian setting, is that a Nash equilibrium equilibrium provides
``consistent'' predictions, in that if all parties predict that the
game will converge to a Nash equilibrium, no one will benefit by
changing strategies. Having a consistent prediction does not ensure
that all agents' payoffs are maximized though.  The study of small
perturbations around Nash equilibria from a stochastic dynamics
perspective is just one example of a `refinement' of Nash equilibrium,
that is a criterion for selecting a single equilibrium state when more
than one is present~\cite{mazh98}.

In cooperative game theory the agents are able to enter binding
contracts with one another, and thereby coordinate their
strategies. This allows the agents to avoid being ``stuck'' in Nash
equilibria that are Pareto inefficient, that is being stuck at
equilibrium profiles in which all agents would benefit if only they
could agree to all adopt different strategies, with no possibility of
betrayal. The {\it characteristic function} of a game involves subsets
(`coalitions') of agents playing the game. For each such subset, it
gives the sum of the payoffs of the agents in that subset that those
agents can guarantee if they coordinate their strategies. An
$imputation$ is a division of such a guaranteed sum among the members
of the coalition. It is often the case that for a subset of the agents
in a coalition one imputation $dominates$ another, meaning that under
threat of leaving the coalition that subset of agents can demand the
first imputation rather than the second. So the problem each agent $i$
is confronted with in a cooperative game is which set of other agents
to form a coalition with, given the characteristic function of the
game and the associated imputations $i$ can demand of its partners.
There are several different kinds of solution for cooperative games
that have received detailed study, varying in how the agents address
this problem of who to form a coalition with. Some of the more popular
are the `core', the `Shapley value', the `stable set solution', and
the `nucleolus'.

In the real world, the actual underlying game the agents are playing
does not only involve the actions considered in cooperative game
theory's analysis of coalitions and imputations. The strategies of
that underlying game also involve bargaining behavior, considerations
of trying to cheat on a given contract, bluffing and threats, and the
like. In many respects, by concentrating on solutions for coalition
formation and their relation with the characteristic function,
cooperative game theory abstracts away these details of the true
underlying game.  Conversely though, progress has recently been made
in understanding how cooperative games can arise from non-cooperative
games, as they must in the real world ~\cite{auha92}.

\subsection{Evolution and Learning in Games}

Not surprisingly, game theory has come to play a large role in the
field of multi-agent systems. In addition, due to Darwinian natural
selection, one might expect game theory to be quite important in
population biology, in which the ``utility functions'' of the
individual agents can be taken to be their reproductive fitness. As it
turns out, there is an entire subfield of game theory concerned with
this connection with population biology, called `evolutionary game
theory' ~\cite{mayn82,mest96}.

To introduce evolutionary game theory, consider a game in which all
players share the same space of possible strategies, and there is an
additional space of possible `attribute vectors' that characterize an
agent, along with a probability distribution $g$ across that new
space. (Examples of attributes in the physical world could be things
like size, speed, etc.) We select a set of agents to play a game by
randomly sampling $g$. Those agents' attribute vectors jointly
determine the payoff matrices of each of the individual
agents. (Intuitively, what benefit accrues to an agent for taking a
particular action depends on its attributes and those of the other
agents.) However each agent $i$ has limited information concerning
both its attribute vector and that of the other players in the game,
information encapsulated in an `information structure'. The
information structure specifies how much each agent knows concerning
the game it is playing.

In this context, we enlarge the meaning of the term ``strategy'' to
not just be a mapping from information sets and the like to actions,
but from entire information structures to actions. In addition to the
distribution $g$ over attribute vectors, we also have a distribution
over strategies, $h$. A strategy $s$ is a `population strategy' if $h$
is a delta function about $s$.  Intuitively, we have a population
strategy when each animal in a population ``follows the same
behavioral rules'', rules that take as input what the animal is able
to discern about its strengths and weakness relative to those other
members of the population, and produce as output how the animal will
act in the presence of such animals.

Given $g$, a population strategy centered about $s$, and its own
attribute vector, any player $i$ in the support of $g$ has an expected
payoff for any strategy it might adopt. When $i$'s payoff could not
improve if it were to adopt any strategy other than $s$, we say that
$s$ is `evolutionary stable'. Intuitively, an evolutionary stable
strategy is one that is stable with respect to the introduction of
mutants into the population.

Now consider a sequence of such evolutionary games. Interpret the
payoff that any agent receives after being involved in such a game as
the `reproductive fitness' of that agent, in the biological sense. So
the higher the payoff the agent receives, in comparison to the
fitnesses of the other agents, the more ``offspring'' it has that get
propagated to the next game. In the continuum-time limit, where games
are indexed by the real number $t$, this can be formalized by a
differential equation. This equation specifies the derivative of $g_t$
evaluated for each agent $i$'s attribute vector, as a montonically
increasing function of the relative difference between the payoff of
$i$ and the average payoff of all the agents. (We also have such an
equation for $h$.) The resulting dynamics is known as `replicator
dynamics', with an evolutionary stable population strategy, if it
exists, being one particular fixed point of the dynamics.

Now consider removing the reproductive aspect of evolutionary game
theory, and instead have each agent propagate to the next game, with
``memory'' of the events of the preceding game. Furthermore, allow
each agent to modify its strategy from one game to the next by
``learning'' from its memory of past games, in a bounded rational
manner. The field of learning in games is concerned with exactly such
situations~\cite{fule98,axel84,bank94,besw98,epst98,kere98,nosi98,neym85}. Most
of the formal work in this field involves simple models for the
learning process of the agents. For example, in `ficticious play'
\cite{fule98}, in each successive game, each agent $i$ adopts what
would be its best strategy if its opponents chose their strategies
according to the empirical frequency distribution of such strategies
that $i$ has encountered in the past. More sophisticated versions of
this work employ simple Bayesian learning algorithms, or re-inventions
of some of the techniques of the RL community
~\cite{roer95}. Typically in learning in games one defines a payoff to
the agent for a sequence of games, for example as a discounted sum of
the payoffs in each of the constituent games.  Within this framework
one can study the long term effects of strategies such as cooperation
and see if they arise naturally and if so, under what circumstances.

Many aspects of real world games that do not occur very naturally
otherwise arise spontaneously in these kinds of games. For example,
when the number of games to be played is not pre-fixed, it may behoove
a particular agent $i$ to treat its opponent better than it would
otherwise, since $i$ $may$ have to rely on that other agent's treating
it well in the future, if they end up playing each other again.  This
framework also allows us to investigate the dependence of evolving
strategies on the amount of information available to the
agents~\cite{mill96a}; the effect of communication on the evolution of
cooperation~\cite{mill96b,mibu98}; and the parallels between auctions
and economic theory~\cite{homi91,mian95}.

In many respects, learning in games is even more relevant to the study
of COINs than is traditional game theory. However it suffers from the
same major shortcoming; it is almost exclusively focused on the
forward problem rather than the inverse problem. In essence, COIN
design is the problem of $inverse$ game theory.

\subsubsection{El Farol Bar Problem}
\label{sec:bar}

The ``El Farol'' bar problem and its variants provide a clean and
simple testbed for investigating certain kinds of interactions among
agents~\cite{arth94,chzh97,sebe98}.  In the original version of the
problem, which arose in economics, at each time step (each ``night''),
each agent needs to decide whether to attend a particular bar.  The
goal of the agent in making this decision depends on the total
attendance at the bar on that night. If the total attendance is below
a preset capacity then the agent should have attended.  Conversely, if
the bar is overcrowded on the given night, then the agent should not
attend.  (Because of this structure, the bar problem with capacity set
to $50\%$ of the total number of agents is also known as the `minority
game'; each agent selects one of two groups at each time step, and
those that are in the minority have made the right choice). The agents
make their choices by predicting ahead of time whether the attendance
on the current night will exceed the capacity and then taking the
appropriate course of action.

What makes this problem particularly interesting is that it is
impossible for each agent to be perfectly ``rational'', in the sense of
correctly predicting the attendance on any given night.  This is
because if most agents predict that the attendance will be low (and
therefore decide to attend), the attendance will actually high, while
if they predict the attendance will be high (and therefore decide not
to attend) the attendance will be low. (In the language of game
theory, this essentially amounts to the property that there are no
pure strategy Nash equilibria ~\cite{chen97,zamb99}.)  Alternatively,
viewing the overall system as a COIN, it has a Prisoner's Dilemma-like
nature, in that ``rational'' behavior by all the individual agents thwarts
the global goal of maximizing total enjoyment (defined as the sum of
all agents' enjoyment and maximized when the bar is exactly at
capacity).

This frustration effect is similar to what occurs in spin glasses in
physics, and makes the bar problem closely related to the physics of
emergent behavior in distributed systems~\cite{cava98,chzh97,chzh98,zhan98}.
Researchers have also studied the dynamics of the bar problem to
investigate economic properties like competition, cooperation and
collective behavior and especially their relationship to market
efficiency~\cite{capl98,joja98,sama97}.

\subsection{Biologically Inspired Systems}

Properly speaking, biological systems do not involve utility functions
and searches across them with RL algorithms. However it has long been
appreciated that there are many ways in which viewing biological
systems as involving searches over such functions can lead to deeper
understanding of them ~\cite{schu90, wrig32}. Conversely, some have
argued that the mechanism underlying biological systems can be used to
help design search algorithms ~\cite{holl93}.\footnote{See
~\cite{mawo98,woma97} though for some counter-arguments to the
particular claims most commonly made in this regard.} 

These kinds of reasoning which relate utility functions and biological
systems have traditionally focussed on the case of a single biological
system operating in some external environment. If we extend this kind
of reasoning, to a set of biological systems that are co-evolving with
one another, then we have essentially arrived at biologically-based
COINs. This section discusses some of how previous work in the
literature bears on this relationship between COINs and biology.

\subsubsection{Population Biology and Ecological Modeling}
\label{sec:popbio}
The fields of population biology and ecological modeling are concerned
with the large-scale ``emergent'' processes that govern the systems
that consist of many (relatively) simple entities interacting
 with one another ~\cite{beth96,hast97}.  As usually cast, the ``simple
entities'' are members of one or more species, and the interactions
are some mathematical abstraction of the process of natural selection
as it occurs in biological systems (involving processes like genetic
reproduction of various sorts, genotype-phenotype mappings, inter and
intra-species competitions for resources, etc.).  Population Biology
and ecological modeling in this context addresses questions concerning
the dynamics of the resultant ecosystem, and in particular how its
long-term behavior depends on the details of the interactions between
the constituent entities. Broadly construed, the paradigm of ecological
modeling can even be broadened to study how natural selection and
self-regulating feedback creates a stable planet-wide ecological
environment---Gaia~\cite{lent98}.

The underlying mathematical models of other fields can often be
usefully modified to apply to the kinds of systems population biology
is interested in~\cite{basn93}. (See also the discussion in the game
theory subsection above.) Conversely, the underlying
mathematical models of population biology and ecological modeling can
be applied to other non-biological systems. In particular, those models shed
light on social issues such as the emergence of language or culture,
warfare, and economic competition ~\cite{epst97,epax96,gabo98}. They also can
be used to investigate more abstract issues concerning the behavior of
large complex systems with many interacting components
~\cite{galu98,hans97,mcfa94,niik97,poll98}.

Going a bit further afield, an approach that is related in spirit to
ecological modeling is `computational ecologies'. These are large
distributed systems where each component of the system's acting
(seemingly) independently results in complex global behavior.  Those
components are viewed as constituting an ``ecology'' in an abstract
sense (although much of the mathematics is not derived from the
traditional field of ecological modeling). In particular, one can
investigate how the dynamics of the ecology is influenced by the
information available to each component and how cooperation and
communication among the components affects that
dynamics~\cite{hube88,huho88}.

Although in some ways the most closely related to COINs of the current
ecology-inspired research, the field of computational ecologies has
some significant shortcomings if one tries to view it as a full
science of COINs. In particular, it suffers from not being designed to
solve the inverse problem of how to configure the system so as to
arrive at a particular desired dynamics. This is a difficulty endemic
to the general program of equating ecological modeling and population
biology with the science of COINs.  These fields are primarily
concerned with the ``forward problem'' of determining the dynamics
that arises from certain choices of the underlying system. Unless
one's desired dynamics is sufficiently close to some dynamics that
was previously catalogued (during one's investigation of the forward
problem), one has very little information on how to set up the
components and their interactions to achieve that desired dynamics.
In addition, most of the work in these fields does not involve RL
algorithms, and viewed as a context in which to design COINs suffers
from a need for hand-tailoring, and potentially lack of robustness and
scalability.

\subsubsection{Swarm Intelligence}

The field of `swarm intelligence' is concerned with systems that are
modeled after social insect colonies, so that the different
components of the system are queen, worker, soldier, etc. It can be
viewed as ecological modeling in which the individual entities have
extremely limited computing capacity and/or action sets, and in which
there are very few types of entities.  The premise of the field is
that the rich behavior of social insect colonies arises not from the
sophistication of any individual entity in the colony, but from the
interaction among those entities.  The objective of current research is to
uncover kinds of interactions among the entity types that lead to pre-specified
behavior of some sort.

More speculatively, the study of social insect colonies may also
provide insight into how to achieve learning in large distributed
systems.  This is because at the level of the individual insect in a
colony, very little (or no) learning takes place. However across
evolutionary time-scales the social insect species as a whole
functions as if the various individual types in a colony had
``learned'' their specific functions.  The ``learning'' is the direct
result of natural selection. (See the discussion on this topic in the
subsection on ecological modeling.)

Swarm intelligences have been used to adaptively allocate tasks in a
mail company~\cite{boso99}, solve the traveling salesman
problem~\cite{doga97b,doga97a} and route data efficiently in dynamic
networks~\cite{bohe99,scho97,sudr97} among others.  Despite this, such
intelligences do not really constitute a general approach to designing
COINs. There is no general framework for adapting swarm intelligences
to maximize particular world utility functions. Accordingly, such
intelligences generally need to be hand-tailored for each
application. And after such tailoring, it is often quite a stretch to
view the system as ``biological'' in any sense, rather than just a
simple and {\it a priori} reasonable modification of some previously
deployed system.

\subsubsection{Artificial Life}

The two main objectives of Artificial Life, closely related to one
another, are understanding the abstract functioning and especially the
origin of terrestrial life, and creating organisms that can
meaningfully be called ``alive''~\cite{lang92a}. 

The first objective involves formalizing and abstracting the
mechanical processes underpinning terrestrial life. In particular,
much of this work involves various degrees of abstraction of the
process of self-replication ~\cite{brac98,smit92,vonn66}. Some of the
more real-world-oriented work on this topic involves investigating how
lipids assemble into more complex structures such as vesicles and
membranes, which is one of the fundamental questions concerning the
origin of life~\cite{deor80,edpe98,orsh78,poch96,nepo99}. Many
computer models have been proposed to simulate this process, though
most suffer from overly simplifying the molecular morphology.

More generally, work concerned with the origin of life can constitute
an investigation of the functional self-organization that gives rise
to life ~\cite{mcva99}. In this regard, an important early work on
functional self-organization is the {\em lambda calculus}, which
provides an elegant framework (recursively defined functions, lack of
distinction between object and function, lack of architectural
restrictions) for studying computational systems~\cite{chur41}.  This
framework can be used to develop an artificial chemistry ``function
gas'' that displays complex cooperative properties~\cite{font92}.

The second objective of the field of Artificial Life is less concerned
with understanding the details of terrestrial life per se than of
using terrestrial life as inspiration for how to design living
systems.  For example, motivated by the existence (and persistence) of
computer viruses, several workers have tried to design an immune
system for computers that will develop ``antibodies'' and handle
viruses both more rapidly and more efficiently than other
algorithms~\cite{fope94,keph94,soho99}.  More generally, because we
only have one sampling point (life on Earth), it is very difficult to
precisely formulate the process by which life emerged. By creating an
artificial world inside a computer however, it is possible to study
far more general forms of life~\cite{ray92,ray94,ray95}. 
See also ~\cite{woma99} where the argument is presented that the
richest way of approaching the issue of defining ``life'' is
phenomenologically, in terms of self-$dis$similar scaling properties
of the system.

\subsubsection{Training cellular automata with genetic algorithms}
\label{sec:ga}
Cellular automata can be viewed as digital abstractions of physical
gases~\cite{boon91,codd68,wolf83,wolf94}. 
Formally, they are discrete-time recurrent neural
nets where the neurons live on a grid, each neuron has a finite number
of potential states, and inter-neuron connections are (usually) purely
local. (See below for a discussion of recurrent neural nets.) So the
state update rule of each neuron is fixed and local, the next state of
a neuron being a function of the current states of it and of its
neighboring elements.

The state update rule of (all the neurons making up) any particular
cellular automaton specifies the mapping taking the initial
configuration of the states of all of its neurons to the final,
equilibrium (perhaps strange) attractor configuration of all those
neurons.  So consider the situation where we have a desired such
mapping, and want to know an update rule that induces that mapping.
This is a search problem, and can be viewed as similar to the inverse
problem of how to design a COIN to achieve a pre-specified global
goal, albeit a ``COIN'' whose nodal elements do not use RL
algorithms.

Genetic algorithms are a special kind of search algorithm, based on
analogy with the biological process of natural selection via
recombination and mutation of a genome~\cite{mitc96}. Although genetic
algorithms (and `evolutionary computation' in general) have been
studied quite extensively, there is no formal theory justifying
genetic algorithms as search algorithms \cite{mawo96,woma97} and few
empirical comparisons with other search techniques.  One example of a
well-studied application of genetic algorithms is to (try to) solve
the inverse problem of finding update rules for a cellular automaton
that induce a pre-specified mapping from its initial configuration to
its attractor configuration.  To date, they have used this way only
for extremely simple configuration mappings, mappings which can be
trivially learned by other kinds of systems. Despite the simplicity of
these mappings, the use of genetic algorithms to try to train cellular
automata to exhibit them has achieved little
success~\cite{crha93,dami94, mitc98,micr94}.

\subsection{Physics-Based Systems}
\subsubsection{Statistical Physics}
Equilibrium statistical physics is concerned with the stable state
character of large numbers of very simple physical objects,
interacting according to well-specified local deterministic laws, with
probabilistic noise processes superimposed~\cite{asme76,reif65}. 
Typically there is no sense in which such systems can be said
to have centralized control, since all particles contribute comparably
to the overall dynamics.

Aside from mesoscopic statistical physics, the numbers of particles
considered are usually huge ({\it e.g.}, $10^{23}$), and the particles
themselves are extraordinarily simple, typically having only a few
degrees of freedom. Moreover, the noise processes usually considered
are highly restricted, being those that are formed by ``baths'', of
heat, particles, and the like. Similarly, almost all of the field
restricts itself to deterministic laws that are readily encapsulated
in Hamilton's equations (Schrodinger's equation and its
field-theoretic variants for quantum statistical physics). In fact,
much of equilibrium statistical physics isn't even concerned with the
dynamic laws by themselves (as for example is stochastic Markov
processes).  Rather it is concerned with invariants of those laws
({\em e.g.}, energy), invariants that relate the states of all of the
particles. Trivially then, deterministic laws without such
readily-discoverable invariants are outside of the purview of much of
statistical physics.

One potential use of statistical physics for COINs involves taking the
systems that statistical physics analyzes, especially those analyzed
in its condensed matter variant ({\em e.g.}, spin glasses
\cite{stei96a,stei96b}), as simplified models of a class of
COINs. This approach is used in some of the analysis of the Bar
problem (see above). It is used more overtly in (for example) the work
of Galam \cite{gala99}, in which the equilibrium coalitions of a set
of ``countries'' are modeled in terms of spin glasses. This approach
cannot provide a general COIN framework though. In addition to the
restrictions listed above on the kinds of systems it considers, this
is due to its not providing a general solution to arbitrary COIN
inversion problems, and to its not employing RL
algorithms.\footnote{In regard to the latter point however, it's
interesting to speculate about recasting statistical physics as a
COIN, by viewing each of the particles in the physical system as
running an ``RL algorithm'' that perfectly optimizes the ``utility
function'' of its Lagrangian, given the ``actions'' of the other
particles. In this perspective, many-particle physical systems are
multi-stage games that are at Nash equilibrium in each stage. So for
example, a frustrated spin glass is such a system at a Nash equilibrium
that is not Pareto optimal.}

Another contribution that statistical physics can make is with the
mathematical techniques it has developed for its own purposes, like
mean field theory, self-averaging approximations, phase transitions,
Monte Carlo techniques, the replica trick, and tools to analyze the
thermodynamic limit in which the number of particles goes to
infinity. Although such techniques have not yet been applied to COINs,
they have been successfully applied to related fields. This is
exemplified by the use of the replica trick to analyze two-player
zero-sum games with random payoff matrices in the thermodynamic limit
of the number of strategies in \cite{been98}. Other examples
are the numeric investigation of iterated prisoner's dilemma played on
a lattice \cite{szto98}, the analysis of stochastic games by
expressing of deviation from rationality in the form of a ``heat
bath'' \cite{mazh98}, and the use of topological entropy
to quantify the complexity of a voting system studied in \cite{mebr98}.

Other quite recent work in the statistical physics literature is
formally identical to that in other fields, but presents it from a
novel perspective. A good example of this is \cite{sigu98}, which is
concerned with the problem of controlling a spatially extended system
with a single controller, by using an algorithm that is identical to a
simple-minded proportional RL algorithm (in essence, a rediscovery of
RL).

\subsubsection{Action Extremization} Much of the theory of physics can
be cast as solving for the extremization of an actional, which is a
functional of the worldline of an entire (potentially many-component)
system across all time. The solution to that extremization problem
constitutes the actual worldline followed by the system. In this way
the calculus of variations can be used to solve for the worldline of
a dynamic system. As an example, simple Newtonian dynamics can be cast
as solving for the worldline of the system that extremizes a quantity
called the `Lagrangian', which is a function of that worldline and
of certain parameters ({\em e.g.}, the `potential energy') governing the
system at hand. In this instance, the calculus of variations simply
results in Newton's laws.

If we take the dynamic system to be a COIN, we are assured that its
worldline automatically optimizes a ``global goal'' consisting of the
value of the associated actional. If we change physical aspects of the
system that determine the functional form of the actional ({\em e.g.},
change the system's potential energy function), then we change the
global goal, and we are assured that our COIN optimizes that new
global goal. Counter-intuitive physical systems, like those that
exhibit Braess' paradox ~\cite{bass92}, are simply systems for which
the ``world utility'' implicit in our human intuition is extremized at
a point different from the one that extremizes the system's actional.

The challenge in exploiting this to solve the COIN design problem is
in translating an arbitrary provided global goal for the COIN into a
parameterized actional. Note that that actional must govern the
dynamics of the physical COIN, and the parameters of the actional must
be physical variables in the COIN, variables whose values we can
modify.

\subsubsection{Active Walker Models}
The field of active walker models \cite{batt97,heke97,hesc97} is
concerned with modeling ``walkers'' (be they human walkers or instead
simple physical objects) crossing fields along trajectories, where
those trajectories are a function of several factors, including in
particular the trails already worn into the field. Often the kind of
trajectories considered are those that can be cast as solutions to
actional extremization problems so that the walkers can be explicitly
viewed as agents optimizing a private utility.

One of the primary concerns with the field of active walker models is
how the trails worn in the field change with time to reach a final
equilibrium state.  The problem of how to design the cement pathways
in the field (and other physical features of the field) so that the
final paths actually followed by the walkers will have certain
desirable characteristics is then one of solving for parameters of the
actional that will result in the desired worldline. This is a special
instance of the inverse problem of how to design a COIN.

Using active walker models this way to design COINs, like action
extremization in general, probably has limited applicability. Also, it
is not clear how robust such a design approach might be, or whether it
would be scalable and exempt from the need for hand-tailoring.

\subsection{Other Related Subjects}
This subsection presents a ``catch-all'' of other fields that have
little in common with one another except that they bear some relation
to COINs.

\subsubsection{Stochastic Fields} An extremely well-researched body of
work concerns the mathematical and numeric behavior of systems for
which the probability distribution over possible future states
conditioned on preceding states is explicitly provided. This work
involves many aspects of Monte Carlo numerical algorithms~\cite{neal96},
all of Markov Chains \cite{free83,norr98,stew95}, and especially Markov
fields, a topic that encompasses the Chapman-Kolmogorov equations
~\cite{gard85} and its variants: Liouville's equation, the
Fokker-Plank equation, and the Detailed-balance equation in
particular. Non-linear dynamics is also related to this body of work
(see the synopsis of iterated function systems below and the synopsis
of cellular automata above), as is Markov competitive decision
processes (see the synopsis of game theory above).

Formally, one can cast the problem of designing a COIN as how to fix
each of the conditional transition probability distributions of the
individual elements of a stochastic field so that the aggregate
behavior of the overall system is of a desired form.\footnote{In
contrast, in the field of Markov decision processes, discussed in
~\cite{caka94}, the full system may be a Markov field, but the system
designer only sets the conditional transition probability distribution
of a few of the field elements at most, to the appropriate ``decision
rules''. Unfortunately, it is hard to imagine how to use the results
of this field to design COINs because of major scaling problems. Any
decision process must accurately model likely future modifications to
its own behavior --- often an extremely daunting task
~\cite{mawo98}. What's worse, if multiple such decision processes are
running concurrently in the system, each such process must also model
the others, potentially needing to model them in their full
complexity.}  Unfortunately, almost all that is known in this area
instead concerns the forward problem, of inferring aggregate behavior
from a provided set of conditional distributions. Although such
knowledge provides many ``bits and pieces'' of information about how
to tackle the inverse problem, those pieces collectively cover only a
very small subset of the entire space of tasks we might want the COIN
to perform. In particular, they tell us very little about the case
where the conditional distribution encapsulates RL algorithms.

\subsubsection{Iterated Function Systems}
The technique of iterated function systems ~\cite{barn86} grew out of
the field of nonlinear dynamics ~\cite{rosa92,stro94,tabo89}. In such
systems a function is repeatedly and recursively applied to
itself. The most famous example is the logistic map, $x_{n+1} =
rx_n(1-x_n)$ for some $r$ between 0 and 4 (so that $x$ stays between 0
and 1). More generally the function along with its arguments can be
vector-valued. In particular, we can construct such functions out of
affine transformations of points in a Euclidean plane.

Iterated functions systems have been applied to image data. In this
case the successive iteration of the function generically generates a
fractal, one whose precise character is determined by the initial
iteration-1 image. Since fractals are ubiquitous in natural images, a
natural idea is to try to encode natural images as sets of iterated
function systems spread across the plane, thereby potentially
garnering significant image compression. The trick is to manage the
inverse step of starting with the image to be compressed, and
determining what iteration-1 image(s) and iterating function(s) will
generate an accurate approximation of that image.

In the language of nonlinear dynamics, we have a dynamic system that
consists of a set of iterating functions, together with a desired
attractor (the image to be compressed). Our goal is to determine what
values to set certain parameters of our dynamic system to so that the
system will have that desired attractor. The potential relationship
with COINs arises from this inverse nature of the problem tackled by
iterated function systems. If the goal for a COIN can be cast as its
relaxing to a particular attractor, and if the distributed
computational elements are isomorphic to iterated functions, then the
tricks used in iterated functions theory could be of use.

Although the techniques of iterated function systems might prove of
use in designing COINs, they are unlikely to serve as a generally
applicable approach to designing COINs. In addition, they do not
involve RL algorithms, and often involve extensive hand-tuning.

\subsubsection{Recurrent Neural Nets}

A recurrent neural net consists of a finite set of ``neurons'' each of
which has a real-valued state at each moment in time. Each neuron's
state is updated at each moment in time based on its current state and
that of some of the other neurons in the system. The topology of such
dependencies constitute the ``inter-neuronal connections'' of the net,
and the associated parameters are often called the ``weights'' of the
net. The dynamics can be either discrete or continuous ({\em i.e.}, given by
difference or differential equations).

Recurrent nets have been investigated for many
purposes~\cite{giku94,hogi95,pear89,zami99}.  One of the more
famous of these is associative memories. The idea is that given a
pre-specified pattern for the (states of the neurons in the) net,
there may exist inter-neuronal weights which result in a basin of
attraction focussed on that pattern.  If this is the case, then the
net is equivalent to an associative memory, in that a complete
pre-specified pattern across all neurons will emerge under the net's
dynamics from any initial pattern that partially matches the full
pre-specified pattern. In practice, one wishes the net to
simultaneously possess many such pre-specified associative
memories. There are many schemes for ``training'' a recurrent net to
have this property, including schemes based on spin glasses
\cite{hopf82,hota85,hota86} and schemes based on gradient descent
\cite{rumc86}.

As can the fields of cellular automata and iterated function systems,
the field of recurrent neural nets can be viewed as concerning certain
variants of COINs. Also like those other fields though, recurrent
neural nets has shortcomings if one tries to view it as a general
approach to a science of COINs. In particular, recurrent neural nets
do not involve RL algorithms, and training them often suffers from
scaling problems. More generally, in practice they can be hard to
train well without hand-tailoring.

\subsubsection{Network Theory}

Packet routing in a data network ~\cite{bega92,hsla87,stal94,wava96,kell96,grov97}
presents a particularly interesting domain for the investigation of
COINs. In particular, with such routing: \\
(i) the problem is inherently distributed; \\
(ii) for all but the most
trivial networks it is impossible to employ global control ;  \\ 
(iii) the routers have only access to local information 
(routing tables); \\
(iv) it constitutes a relatively clean and easily modified
experimental testbed; and \\
(v) there are potentially major bottlenecks induced by `greedy' 
behavior on the part of the individual routers, which behavior 
constitutes a readily investigated instance of the Tragedy Of the
Commons (TOC).

Many of the approaches to packet routing incorporate a variant on
RL~\cite{boli94,brto99,chye96,libo93,mami98}.  Q--routing is perhaps
the best known such approach and is based on routers using
reinforcement learning to select the best path~\cite{boli94}.
Although generally successful, Q--routing is not a general scheme for
inverting a global task. This is even true if one restricts attention
to the problem of routing in data networks --- there exists a global
task in such problems, but that task is directly used to construct the
algorithm.

A particular version of the general packet routing problem that is
acquiring increased attention is the Quality of Service (QoS) problem,
where different communication packets (voice, video, data) share the
same bandwidth resource but have widely varying importances both to
the user and (via revenue) to the bandwidth provider. Determining
which packet has precedence over which other packets in such cases is
not only based on priority in arrival time but more generally on the
potential effects on the income of the bandwidth provider.  In this
context, RL algorithms have been used to determine routing policy,
control call admission and maximize revenue by allocating the
available bandwidth efficiently~\cite{brto99,mami98}.

Many researchers have exploited the noncooperative game theoretic
understanding of the TOC in order to explain the bottleneck character
of empirical data networks' behavior and suggest potential alternatives
to current routing
schemes~\cite{bese99,ecsi91,kola97,kola98,laan99,laor97,olaf97,orro93a,shen95}.
Closely related is work on various ``pricing''-based resource
allocation strategies in congestable data networks
~\cite{mava95}. This work is at least partially based upon current
understanding of pricing in toll lanes, and traffic flow in general
(see below).  All of these approaches are particularly of interest
when combined with the RL-based schemes mentioned just above. Due to
these factors, much of the current research on a general framework for
COINs is directed toward the packet-routing domain (see next section).

\subsubsection{Traffic Theory}

Traffic congestion typifies the TOC public good problem: everyone
wants to use the same resource, and all parties greedily trying to
optimize their use of that resource not only worsens global behavior,
but also worsens {\em their own} private utility ({\em e.g.}, if everyone
disobeys traffic lights, everyone gets stuck in traffic jams).
Indeed, in the well-known Braess' paradox ~\cite{bass92,coje97,coke90,kola99}, 
keeping
everything else constant --- including the number and destinations of
the drivers --- but opening a new traffic path can {\it increase}
everyone's time to get to their destination. (Viewing the overall
system as an instance of the Prisoner's dilemma, this paradox in
essence arises through the creation of a novel `defect-defect' option
for the overall system.) Greedy behavior on the part of individuals
also results in very rich global dynamic patterns, such as stop and go
waves and clusters~\cite{hetr98a,hetr98b}.

Much of traffic theory employs and investigates tools that have
previously been applied in statistical
physics~\cite{hetr98a,keko95,kere96,pipe53,quxi96} (see subsection
above).  In particular, the spontaneous formation of traffic jams
provides a rich testbed for studying the emergence of complex activity
from seemingly chaotic states~\cite{hetr98a,heke98}.  Furthermore, the
dynamics of traffic flow is particular amenable to the application and
testing of many novel numerical methods in a controlled
environment~\cite{baha95,bimi92,scsc95}. Many experimental studies
have confirmed the usefulness of applying insights gleaned from such
work to real world traffic scenarios ~\cite{hetr98a,nast97,naig97}.

\subsubsection{Topics from further afield}

Finally, there are a number of other fields that, while either still
nascent or not extremely closely related to COINs, are of interest in
COIN design:

{\bf Amorphous computing:} Amorphous computing grew out of the idea of
replacing traditional computer design, with its requirements for high
reliability of the components of the computer, with a novel approach
in which widespread unreliability of those components would not
interfere with the computation~\cite{abfo98}. Some of its more
speculative aspects are concerned with ``how to program'' a massively
distributed, noisy system of components which may consist in part of
biochemical and/or biomechanical components~\cite{knsu98,weho98}. Work
here has tended to focus on schemes for how to robustly induce desired
geometric dynamics across the physical body of the amorphous computer
--- issue that are closely related to morphogenesis, and thereby lend
credence to the idea that biochemical components are a promising
approach. Especially in its limit of computers with very small
constituent components, amorphous computing also is closely related to
the fields of nanotechnology~\cite{drex92} and control of smart
matter (see below).

{\bf Control of smart matter:}. As the prospect of
nanotechnology-driven mechanical systems gets more concrete, the
daunting problem of how to robustly control, power, and sustain
protean systems made up of extremely large sets of nano-scale devices
looms more important~\cite{guho97b,guho97,hohu98}. If this problem were
to be solved one would in essence have ``smart matter''. For example,
one would be able to ``paint'' an airplane wing with such matter and
have it improve drag and lift properties significantly.

{\bf Morphogenesis:} How does a leopard embryo get its spots, or a
zebra embryo its stripes? More generally, what are the processes
underlying morphogenesis, in which a body plan develops among a
growing set of initially undifferentiated cells? These questions,
related to control of the dynamics of chemical reaction waves, are
essentially special cases of the more general question of how ontogeny
works, of how the genotype-phenotype mapping is carried out in
development. The answers involve homeobox (as well as many other)
genes ~\cite{bard92,dubo94,kash93,flei98,turi52}.  Under the presumption that
the functioning of such genes is at least in part designed to
facilitate genetic changes that increase a species' fitness, that
functioning facilitates solution of the inverse problem, of finding
small-scale changes (to DNA) that will result in ``desired'' large
scale effects (to body plan) when propagated across a growing
distributed system.  

{\bf Self Organizing systems} The concept of self-organization and
self-organized criticality~\cite{bata88} was originally developed to
help understand why many distributed physical systems are attracted to
critical states that possess long-range dynamic correlations in the
large-scale characteristics of the system. It provides a powerful
framework for analyzing both biological and economic systems.  For
example, natural selection (particularly punctuated equilibrium
~\cite{elgo72,goel77}) can be likened to self-organizing dynamical system,
and some have argued it shares many the properties ({\em e.g.}, scale
invariance) of such systems \cite{bode94}.  Similarly, one can view
the economic order that results from the actions of human agents as a
case of self-organization~\cite{deva99}.  The relationship between
complexity and self-organization is a particularly important one, in
that it provides the potential laws that allow order to arise from
chaos~\cite{kauf95}.

{\bf Small worlds (6 Degrees of Separation):} In many distributed
systems where each component can interact with a small number of
``neighbors'', an important problem is how to propagate information
across the system quickly and with minimal overhead.  On the one
extreme the neighborhood topology of such systems can exist on a
completely regular grid-like structure. On the other, the topology can
be totally random. In either case, certain nodes may be effectively
`cut-off' from other nodes if the information pathways between them
are too long. Recent work has investigated ``small worlds'' networks
(sometimes called 6 degrees of separation) in which underlying
grid-like topologies are ``doped'' with a scattering of long-range,
random connections. It turns out that very little such doping is
necessary to allow for the system to effectively circumvent the
information propagation problem ~\cite{milg67,wast98}.

{\bf Control theory:} Adaptive control
~\cite{aswi94,sabo89}, and in particular adaptive control
involving locally weighted RL algorithms ~\cite{atsh97,moat97},
constitute a broadly applicable framework for controlling small,
potentially inexactly modeled systems. Augmented by techniques in the
control of chaotic systems ~\cite{chco95,dira90,dish97}, they
constitute a very successful way of solving the ``inverse problem''
for such systems. Unfortunately, it is not clear how one could even
attempt to scale such techniques up to the massively distributed
systems of interest in COINs. The next section discusses in detail
some of the underlying reasons why the purely model-based versions of
these approaches are inappropriate as a framework for COINs.

\section{A FRAMEWORK DESIGNED FOR COINs}
\label{sec:math}
Summarizing the discussion to this point, it is hard to see how any
already extant scientific field can be modified to encompass systems
meeting all of the requirements of COINs listed at the beginning of
Section~\ref{sec:lit}. This is not too surprising, since none of those
fields were explicitly designed to analyze COINs.  This section first
motivates in general terms a framework that is explicitly designed for
analyzing COINs. It then presents the formal nomenclature of that
framework. This is followed by derivations of some of the central
theorems of that framework.\footnote{A much more detailed discussion,
including intuitive arguments, proofs and fully formal definitions of
the concepts discussion in this section, can be found in
~\cite{wolp99}.} Finally, we present experiments that illustrate the
power the framework provides for ensuring large world utility in a
COIN.

\subsection{Problems with a model-based approach}

What mathematics might one employ to understand and design COINs?
Perhaps the most natural approach, related to the stochastic fields
work reviewed above, involves the following three steps:

1) First one constructs a detailed stochastic model of the COIN's
dynamics, a model parameterized by a vector $\theta$.  As an example,
$\theta$ could fix the utility functions of the individual agents of
the COIN, aspects of their RL algorithms, which agents communicate
with each other and how, etc.

2) Next we solve for the function $f(\theta)$
which maps the parameters of the model to the resulting stochastic
dynamics.

3) Cast our goal for the system as a whole as achieving a high
expected value of some ``world utility''. Then as our final step we
would have to solve the inverse problem: we would have to search for a
$\theta$ which, via $f$, results in a high value of E(world utility
$\mid \theta$).

Let's examine in turn some of the challenges each of these three steps
entrain:

I) We are primarily interested in very large, very complex systems,
which are noisy, faulty, and often operate in a non-stationary
environment. Moreover, our ``very complex system'' consists of many RL
algorithms, all potentially quite complicated, all running
simultaneously. Clearly coming up with a detailed model that captures
the dynamics of all of this in an accurate manner will often be
extraordinarily difficult. Moreover, unfortunately, given that the
modeling is highly detailed, often the level of verisimilitude
required of the model will be quite high. For example, unless the
modeling of the faulty aspects of the system were quite accurate, the
model would likely be ``brittle'', and overly sensitive to which
elements of the COIN were and were not operating properly at any given
time.

II) Even for models much simpler than the ones called for in (I),
solving explicitly for the function $f$ can be extremely
difficult. For example, much of Markov Chain theory is an attempt to
broadly characterize such mappings. However as a practical matter,
usually it can only produce potentially useful characterizations when
the underlying models are quite inaccurate simplifications of the
kinds of models produced in step (I).

III) Even if one can write down an $f$, solving the associated inverse
problem is often impossible in practice.

IV) In addition to these difficulties, there is a more general problem
with the model-based approach. We wish to perform our analysis on a
``high level''.  Our thesis is that due to the robust and adaptive
nature of the individual agents' RL algorithms, there will be very
broad, easily identifiable regions of $\theta$ space all of which
result in excellent E(world utility $\mid \theta$), and that these
regions will not depend on the precise learning algorithms used to
achieve the low-level tasks (cf. the list at the beginning of
Section~\ref{sec:lit}). To fully capitalize on this, one would want to
be able to slot in and out different learning algorithms for achieving
the low-level tasks without having to redo our entire analysis each
time. However in general this would be possible with a model-based
analysis only for very carefully designed models (if at all). The
problem is that the result of step (3), the solution to the inverse
problem, would have to concern aspects of the COIN that are (at least
approximately) invariant with respect to the precise low-level
learning algorithms used. Coming up with a model that has this
property while still avoiding problems (I-III) is usually an extremely
daunting challenge.

Fortunately, there is an alternative approach which avoids the
difficulties of detailed modeling.  Little modeling of any sort ever
is used in this alternative, and what modeling does arise has little to
do with dynamics. In addition, any such modeling is extremely
high-level, intented to serve as a decent approximation to almost any
system having ``reasonable'' RL algorithms, rather than as an accurate
model of one particular system.

We call any framework based on this alternative a {\bf descriptive
framework}. In such a framework one identifies certain {\bf salient
characteristics} of COINs, which are characteristics of a COIN's
entire worldline that one strongly expects to find in COINs that have
large world utility.  Under this expectation, one makes the assumption
that if a COIN is explicitly modified to have the salient
characteristics (for example in response to observations of its
run-time behavior), then its world utility will benefit. So long as
the salient characteristics are (relatively) easy to induce in a COIN,
then this assumption provides a ready indirect way to cause that COIN
to have large world utility.

An assumption of this nature is the central leverage point that a
descriptive framework employs to circumvent detailed modeling.  Under
it, if the salient characteristics can be induced with little or no
modeling (e.g., via heuristics that aren't rigorously and formally
justified), then they provide an indirect way to improve world utility
without recourse to detailed modeling. In fact, since one does not use
detailed modeling in a descriptive framework, it may even be that one
does not have a fully rigorous mathematical proof that the central
assumption holds in a particular system for one's choice of salient
characteristics. One may have to be content with reasonableness
arguments not only to justify one's scheme for inducing the salient
characteristics, but for making the assumption that characteristics
are correlated with large world utility in the first
place.\footnote{Despite only being implicit, such reasonableness
arguments are all that underpins fields like non-Bayesian learning.
See ~\cite{wolp95b,wolp96a,wolp96b}.}  Of course, the trick in the
descriptive framework is to choose salient characteristics that both
have a beneficial relationship with world utility and that one expects
to be able to induce with relatively little detailed modeling of the
system's dynamics.

\subsection{Nomenclature}

There exist many ways one might try to design a descriptive
framework. In this subsection we present nomenclature needed for a
(very) cursory overview of one of them. (See \cite{wolp99} for a more
detailed exposition, including formal proofs.)

This overview concentrates on the four salient characteristics of
intelligence, learnability, factoredness, and the wonderful life
utility, all defined below. Intelligence is a quantification of how
well an RL algorithm performs.  We want to do whatever we can to help
those algorithms achieve high values of their utility functions.
Learnability is a characteristic of a utility function that one would
expect to be well-correlated with how well an RL algorithm can learn
to optimize it. A utility function is also factored if whenever its
value increases, the overall system benefits. Finally, wonderful life
utility is an example of a utility function that is both learnable and
factored.

After the preliminary definitions below, this section formalizes these
four salient characteristics, derives several theorems relating them,
and illustrates in some computer experiments how those theorems can be used
to help the system achieve high world utility.

\subsubsection{Preliminary Definitions}

{\bf 1)} We refer to an RL algorithm by which an individual component of the
COIN modifies its behavior as a {\bf microlearning} algorithm. We
refer to the initial construction of the COIN, potentially based upon
salient characteristics, as the COIN {\bf initialization}. We use the
phrase {\bf macrolearning} to refer to externally imposed run-time
modifications to the COIN which are based on statistical inference
concerning salient characteristics of the running COIN.

{\bf 2)} For convenience, we take time, $t$, to be discrete and confined to
the integers, {\it Z}. When referring to COIN initialization, we
implicitly have a lower bound on $t$, which without loss of generality
we take to be less than or equal to $0$.

{\bf 3)} All variables that have any effect on the COIN are identified
as components of Euclidean-vector-valued {\bf states} of various
discrete {\bf nodes}.  As an important example, if our COIN consists
in part of a computational ``agent'' running a microlearning
algorithm, the precise configuration of that agent at any time $t$,
including all variables in its learning algorithm, all actions
directly visible to the outside world, all internal parameters, all
values observed by its probes of the surrounding environment, etc.,
all constitute the state vector of a node representing that agent. We
define $\underline{\zeta}_{\eta,t}$ to be a vector in the Euclidean
vector space $\underline{\bf Z}_{\eta,t}$, where the components of
$\underline{\zeta}_{\eta,t}$ give the state of node $\eta$ at time
$t$. The $i$'th component of that vector is indicated by
$\underline{\zeta}_{\eta,t;i}$.

\noindent
{\bf Observation 3.1:} In practice, many COINs will involve variables that are
most naturally viewed as discrete and symbolic. In such cases, we must
exercise some care in how we choose to represent those variables as
components of Euclidean vectors. There is nothing new in this; the
same issue arises in modern work on applying neural nets to inherently
symbolic problems. In our COIN framework, we will usually employ the same
resolution of this issue employed in neural nets, namely representing
the possible values of the discrete variable with a unary
representation in a Euclidean space. Just as with neural nets, values
of such vectors that do not lie on the vertices of the unit hypercube
are not meaningful, strictly speaking. Fortunately though, just as
with neural nets, there is almost always a most natural way to extend
the definitions of any function of interest (like world utility) so
that it is well-defined even for vectors not lying on those
vertices. This allows us to meaningfully define partial derivatives of
such functions with respect to the components of $\underline{\zeta}$,
partial derivatives that we will evaluate at the corners of the unit
hypercube.

{\bf 4)} For notational convenience, we define $\underline{\zeta}_{,t}
\in \underline{\bf{Z}}_{,t}$ to be the vector of the states of all
nodes at time $t$; $\underline{\zeta}_{\;\hat{}\eta,t} \in
\underline{\bf{Z}}_{\;\hat{}\eta,t}$ to be the vector of the states of
all nodes other than $\eta$ at time $t$; and $\underline{\zeta} \equiv
\underline{\zeta}_{,} \in \underline{\bf{Z}} \equiv
\underline{\bf{Z}}'$ to be the entire vector of the states of all
nodes at all times. $\underline{\bf{Z}}$ is infinite-dimensional in
general, and usually assumed to be a Hilbert space. We will often
assume that all spaces $\underline{\bf{Z}}_{,t}$ over all times $t$
are isomorphic to a space $\underline{\bf{Z}}^{(0)}$, i.e.,
$\underline{\bf{Z}}$ is a Cartesian product of copies of
$\underline{\bf{Z}}^{(0)}$.

Also for notational convenience, we define gradients using
$\partial$-shorthand. So for example,
$\partial_{\underline{\zeta}_{,t}} F(\underline{\zeta}_{,})$ is the
vector of the partial derivative of $F(\underline{\zeta}_{,})$ with
respect to the components of $\underline{\zeta}_{,t}$. Also, we will
sometimes treat the symbol ``$t$'' specially, as delineating a range
of components of $\underline{\zeta}$. So for example an expression
like ``$\underline{\zeta}_{,t<t'}$'' refers to all components
$\underline{\zeta}_{,t}$ with $t < t'$.

{\bf 5)} To avoid confusion with the other uses of the comma operator,
we will often use $\bf{x} \bullet \bf{y}$ rather than $(\bf{x},
\bf{y})$ to indicate the vector formed by concatenating the two
ordered sets of vector components $\bf{x}$ and $\bf{y}$. For example,
$\underline{\zeta}_{\eta,t<0} \bullet \underline{\zeta}_{\eta',t>0}$
refers to the vector formed by concatenating those components of the
worldline $\underline{\zeta}$ involving node $\eta$ for times less
than 0 with those components involving node $\eta'$ that have times
greater than 0.

{\bf 6)} We take the universe in which our COIN operates to be
completely deterministic. This is certainly the case for any COIN that
operates in a digital system, even a system that emulates analog
and/or stochastic processes ({\em e.g.}, with a pseudo-random number
generator). More generally, this determinism reflects the fact that
since the real world obeys (deterministic) physics, $any$ real-world
system, be it a COIN or something else, is, ultimately, embedded in
a deterministic system.\footnote{This determinism holds even for
systems with an explicitly quantum mechanical character. Quantum
mechanical systems evolve according to Schrodinger's equation, which
is purely deterministic; as is now well-accepted, the ``stochastic''
aspect of quantum mechanics can be interpreted as an epiphenomenon of
Schrodinger's equation that arises when the Hamiltonian has an
``observational'' or ``entangling'' coupling between some of its
variables ~\cite{ever57,zupa94,geha93}, a coupling that does not obviate
the underlying determinism.}

The perspective to be kept in mind here is that of nonlinear
time-series analysis. A physical time series typically reflects a few
degrees of freedom that are projected out of the underlying space in
which the full system is deterministically evolving, an underlying
space that is actually extremely high-dimensional. This projection
typically results in an illusion of stochasticity in the time
series.

{\bf 7)} Formally, to reflect this determinism, first we bundle all
variables we are not directly considering --- but which nonetheless
affect the dynamics of the system --- as components of some catch-all
{\bf environment node}. So for example any ``noise processes'' and the
like affecting the COIN's dynamics are taken to be inputs from a
deterministic, very high-dimensional environment that is potentially
chaotic and is never directly observed ~\cite{fasi88}. Given such an
environment node, we then stipulate that for all $t, t'$ such that
$t'> t$, $\underline{\zeta}_{,t}$ sets $\underline{\zeta}_{,t'}$
uniquely.

\noindent

{\bf Observation 7.1:} When nodes are ``computational devices'', often
we must be careful to specify the physical extent of those
devices. Such a node may just be the associated CPU, or it may be that
CPU together with the main RAM, or it may include an external storage
device. Almost always, the border of the device $\eta$ will end before
any external system that $\eta$ is ``observing'' begins. This means
that since at time $t$ $\eta$ only knows the value of
$\underline{\zeta}_{\eta,t}$, its ``observational knowledge'' of that
external system is indirect. That knowledge reflects a coupling between
$\underline{\zeta}_{\eta,t}$ and $\underline{\zeta}_{^\eta,t}$, a
coupling that is induced by the dynamical evolution of the system from
preceding moments up to the time $t$. If the dynamics does not force
such a coupling, then $\eta$ has no observational knowledge of the
outside world.

{\bf 8)} We express the dynamics of our system by writing
$\underline{\zeta}_{,t'\ge t} = C(\underline{\zeta}_{,t})$. (In this
paper there will be no need to be more precise and specify the precise
dependency of $C(.)$ on $t$ and/or $t'$.) We define $\{C\}$ to be a
set of constraint equations enforcing that dynamics, and also, more
generally, fixing the entire manifold $C$ of vectors
$\underline{\zeta} \in \underline{\bf{Z}}$ that we consider to be
`allowed'. So $C$ is a subset of the set of all $\underline{\zeta} \in
\underline{\bf{Z}}$ that are consistent with the deterministic laws
governing the COIN, {\em i.e.}, that obey $\underline{\zeta}_{,t'\ge t} =
C(\underline{\zeta}_{,t}) \; \forall \; t, t'$.  We generalize this
notation in the obvious way, so that (for example) $C_{,t\ge t_0}$ is
the manifold consisting of all vectors $\underline{\zeta}_{,t \ge t_0}
\in \underline{\bf Z}_{,t \ge t_0}$ that are projections of a vector
in $C$.

\noindent
{\bf Observation 8.1:} Note that $C_{,t\ge t_0}$ is parameterized by
$\underline{\zeta}_{,t_0}$, due to determinism. Note also that whereas
$C(.)$ is defined for any argument of the form $\underline{\zeta}_{,t}
\in {\bf{\underline{Z}}}_{,t}$ for some $t$ ({\em i.e.}, we can evolve any
point forward in time), in general not all $\underline{\zeta}_{,t} \in
{\bf{\underline{Z}}}_{,t}$ lie in $C_{,t}$. In particular, there may be
extra restrictions constraining the possible states of the system
beyond those arising from its need to obey the relevant dynamical laws
of physics. Finally, whenever trying to express a COIN in terms of
the framework presented here, it is a good rule to try to write out
the constraint equations explicitly to check that what one
has identified as the space $\underline{\bf{Z}}_{,t}$ contains all
quantities needed to uniquely fix the future state of the system.

\noindent
{\bf Observation 8.2:} We do not want to have $\bf{\underline{Z}}$ be the phase space of
every particle in the system. We will instead usually have
$\bf{\underline{Z}}$ consist of variables that, although still
evolving deterministically, exist at a larger scale of granularity
than that of individual particles ({\em e.g.}, thermodynamic variables in
the thermodynamic limit). However we will often be concerned with
physical systems obeying entropy-driven dynamic processes that are
contractive at this high level of granularity. Examples are any of the
many-to-one mappings that can occur in digital computers, and, at a
finer level of granularity, any of the error-correcting processes in
the electronics of such a computer that allow it to operate in a
digital fashion. Accordingly, although the dynamics of our system will
always be deterministic, it need not be invertible.

\noindent
{\bf Observation 8.3:} Intuitively, in our mathematics, all behavior across time is
pre-fixed.  The COIN is a single fixed worldline through
$\underline{\bf{Z}}$, with no ``unfolding of the future'' as the die
underlying a stochastic dynamics get cast. This is consistent with the
fact that we want the formalism to be purely descriptive, relating
different properties of any single, fixed COIN's history. We will
often informally refer to ``changing a node's state at a particular
time'', or to a microlearner's ``choosing from a set of options'', and
the like. Formally, in all such phrases we are really comparing
different worldlines, with the indicated modification distinguishing
those worldlines.

\noindent
{\bf Observation 8.4:} Since the dynamics of any real-world COIN is deterministic, so is the
dynamics of any component of the COIN, and in particular so is any
learning algorithm running in the COIN, ultimately.  However that does
not mean that those deterministic components of the COIN are not
allowed to be ``based on'', or ``motivated by'' stochastic
concepts. The {\it motivation} behind the algorithms run by the
components of the COIN does not change their underlying
nature. Indeed, in our experiments below, we explicitly have the
reinforcement learning algorithms that are trying to maximize private
utility operate in a (pseudo-) probabilistic fashion, with
pseudo-random number generators and the like.

More generally, the deterministic nature of our framework does not
preclude our superimposing probabilistic elements on top of that
framework, and thereby generating a stochastic extension of our
framework.  Exactly as in statistical physics, a stochastic nature can
be superimposed on our space of deterministic worldlines, potentially
by adopting a degree of belief perspective on ``what probability
means'' ~\cite{wolp96c, jayn57}. Indeed, the macrolearning algorithms
we investigate below implicitly involve such a superimposing; they
implicitly assume a probabilistic coupling between the (statistical
estimate of the) correlation coefficient connecting the states of a
pair of nodes and whether those nodes are in the one another's
``effect set''.

Similarly, while it does not salient characteristics that involve
probability distributions, the descriptive framework does not preclude
such characteristics either. As an example, the ``intelligence'' of an
agent's particular action, formally defined below, measures the
fraction of alternative actions an agent could have taken that would
have resulted in a lower utility value. To define such a fraction
requires a measure across the space of such alternative actions, even
if only implicitly. Accordingly, intelligence can be viewed as
involving a probability distribution across the space of potential actions.

In this paper though, we concentrate on the mathematics that obtains
before such probabilistic concerns are superimposed. Whereas the
deterministic analysis presented here is related to game-theoretic
structures like Nash equilibria, a full-blown stochastic extension
would in some ways be more related to structures like correlated
equilibria \cite{auma87}.

{\bf 9)} Formally, there is a lot of freedom in setting the boundary between
what we call ``the COIN'', whose dynamics is determined by $C$, and
what we call ``macrolearning'', which constitutes perturbations to the
COIN instigated from ``outside the COIN'', and which therefore is
$not$ reflected in $C$.  As an example, in much of this paper, we have
clearly specified microlearners which are provided fixed private utility
functions that they are trying to maximize. In such cases usually we
will implicitly take $C$ to be the dynamics of the system,
microlearning and all, {\it for fixed private utilities} that are
specified in $\underline{\zeta}$. For example, $\underline{\zeta}$
could contain, for each microlearner, the bits in an associated
computer specifying the subroutine that that microlearner can call to
evaluate what its private utility would be for some full worldline
$\underline{\zeta}$.

Macrolearning overrides $C$, and in this situation it refers (for
example) to any statistical inference process that modifies the
private utilities at run-time to try to induce the desired salient
characteristics.  Concretely, this would involve modifications to the
bits \{$b_i$\} specifying each microlearner $i$'s private utility,
modifications that are $not$ accounted for in $C$, and that are
potentially based on variables that are not reflected in
$\underline{\bf{Z}}$. Since $C$ does not reflect such macrolearning,
when trying to ascertain $C$ based on empirical observation (as for
example when determining how best to modify the private utilities), we
have to take care to distinguish which part of the system's observed
dynamics is due to $C$ and which part instead reflects externally
imposed modifications to the private utilities.

More generally though, other boundaries between the COIN and
macrolearning-based perturbations to it are possible, reflecting other
definitions of $\underline{\bf{Z}}$, and other interpretations of the
elements of each $\underline{\zeta} \in \underline{\bf{Z}}$. For
example, say that under the perspective presented in the previous
paragraph, the private utility is a function of some components $s$ of
$\underline{\zeta}$, components that do not include the \{$b_i$\}. Now
modify this perspective so that in addition to the dynamics of other
bits, $C$ also encapsulates the dynamics of the bits \{$b_i$\}. Having
done this, we could still view each private utility as being fixed,
but rather than take the bits \{$b_i$\} as ``encoding'' the subroutine
that specifies the private utility of microlearner $i$, we would treat
them as ``parameters'' specifying the functional dependence of the
(fixed) private utility on the components of $\underline{\zeta}$. In
other words, formally, they constitute an extra set of arguments to
$i$'s private utility, in addition to the arguments
$s$. Alternatively, we could simply say that in this situation our
private utilities are time-indexed, with $i$'s private utility at time
$t$ determined by \{$b_{i,t}$\}, which in turn is determined by
evolution under $C$. Under either interpretation of private utility,
any modification under $C$ to the bits specifying $i$'s
utility-evaluation subroutine constitutes dynamical laws by which the
parameters of $i$'s microlearner evolves in time. In this case,
macrolearning would refer to some further removed process that
modifies the evolution of the system in a way not encapsulated in $C$.

For such alternative definitions of $C$/$\underline{\bf{Z}}$, we have
a different boundary between the COIN and macrolearning, and we must
scrutinize different aspects of the COIN's dynamics to infer $C$.
Whatever the boundary, the mathematics of the descriptive framework,
including the mathematics concerning the salient characteristics, is
restricted to a system evolving according to $C$, and explicitly does
not account for macrolearning. This is why the strategy of trying to
improve world utility by using macrolearning to try to induce salient
characteristics is almost always ultimately based on an assumption
rather than a proof.

{\bf 10)} We are provided with some Von Neumann {{\bf world utility}}
$G: \underline{\bf{Z}} \rightarrow \cal{R}$ that ranks the various
conceivable worldlines of the COIN. Note that since the environment
node is never directly observed, we implicitly assume that the world
utility is not directly (!) a function of its state. Our mathematics
will not involve $G$ alone, but rather the relationship between $G$
and various sets of {\bf personal utilities} $g_{\eta,t} :
\underline{\bf{Z}} \otimes {\it Z} \rightarrow \cal{R}$.

Intuitively, as discussed below, for many purposes such personal
utilities are equivalent to arbitrary ``virtual'' versions of the
private utilities mentioned above. In particular, it is only private
utilities that will occur within any microlearning computer algorithms
that may be running in the COIN as manifested in $C$. Personal
utilities are external mathematical constructions that the COIN
framework employs to analyze the behavior of the system. They can be
involved in learning processes, but only as tools that are employed
outside of the COIN's evolution under $C$, {\it i.e.}, only in
macrolearning. (For example, analysis of them can be used to modify
the private utilities.)

\noindent
{\bf Observation 10.1:} These utility definitions are very broad. In particular,
they do not require casting of the utilities as discounted sums.  Note
also that our world utility is not indexed by $t$. Again reflecting
the descriptive, worldline character of the formalism, we simply
assign a single value to an entire worldline of the system, implicitly
assuming that one can always say which of two candidate worldlines are
preferable. So given some ``present time'' $t_0$, issues like which of
two ``potential futures'' $\underline{\zeta}_{, t > t_0}$,
$\underline{\zeta}'_{, t > t_0}$ is preferable are resolved by
evaluating the relevant utility at two associated points
$\underline{\zeta}$ and $\underline{\zeta}'$ , where the $t > t_0$
components of those points are the futures indicated, and the two
points share the same (usually implicit) $t \le t_0$ ``past''
components.

This time-independence of $G$ automatically avoids formal problems
that can occur with general ({\em i.e.}, not necessarily discounted sum)
time-indexed utilities, problems like having what's optimal at one
moment in time conflict with what's optimal at other moments in
time.\footnote{Such conflicts can be especially troublesome when they
interfere with our defining what we mean by an ``optimal'' set of
actions by the nodes at a particular time $t$.  The effects of the
actions by the nodes, adn therefore whether those actions are
``optimal'' or not, depends on the future actions of the
nodes. However if they too are to be ``optimal'', according to their
world-utility, those future actions will depend on {\it their}
futures. So we have a potentially infinite regress of differing
stipulations of what ``optimal'' actions at time $t$ entails.} For
personal utilities such formal problems are often irrelevant
however. Before we begin our work, we as COIN designers must be able
to rank all possible worldlines of the system at hand, to have a
well-defined design task. That is why world utility cannot be
time-indexed. However if a particular microlearner's goal keeps
changing in an inconsistent way, that simply means that that
microlearner will grow ``confused''. From our perspective as COIN
designers, there is nothing {\it a priori} unacceptable about such
confusion. It may even result in better performance of the system as a
whole, in whic case we would actually want to induce it. Nonetheless,
for simplicity, in most of this paper we will have all $g_{\eta,t}$ be
independent of $t$, just like world utility.

World utility is defined as that function that we are ultimately
interested in optimizing. In conventional RL it is a discounted sum,
with the sum starting at time $t$. In other words, conventional RL has
a time-indexed world utility. It might seem that in this at least,
conventional RL considers a case that has more generality than that of
the COIN framework presented here. (It obviously has less generality
in that its world utility is restricted to be a discounted sum.)  In
fact though, the apparent time-indexing of conventional RL is
illusory, and the time-dependent discounted sum world utilty of
conventional RL is actually a special case of the non-time-indexed
world utility of our COIN framework. To see this formally, consider
any (time-independent) world utility $G(\underline{\zeta})$ that
equals $\sum_{t=0}^{\infty} \gamma^t r(\underline{\zeta}_{,t})$ for
some function $r(.)$ and some positive constant $\gamma$ with
magnitude less than 1. Then for any $t' > 0$ and any
$\underline{\zeta}'$ and $\underline{\zeta}''$ where
$\underline{\zeta}'_{,t<t'} = \underline{\zeta}''_{,t<t'}$,
$sgn[G(\underline{\zeta}') - G(\underline{\zeta}'')] =
sgn[\sum_{t=0}^{\infty} \gamma^t r(\underline{\zeta}'_{,t}) -
\sum_{t=0}^{\infty} \gamma^t
r(\underline{\zeta''}_{,t})]$. Conventional RL merely expresses this
in terms of time-dependent utilities
$u_{t'}(\underline{\zeta}_{,t>t'}) \equiv \sum_{t=t'}^{\infty}
\gamma^{t-t'} r(\underline{\zeta}_{,t})$ by writing
$sgn[G(\underline{\zeta}') - G(\underline{\zeta}'')] =
sgn[u_{t'}(\underline{\zeta}') - u_{t'}(\underline{\zeta}'')]$ for all
$t'$. Since utility functions are, by definition, only unique up to
the relative orderings they impose on potential values of their
arguments, we see that conventional RL's use of a time-dependent
discounted sum world utility $u_{t'}$ is identical to use of a
particular time-independent world utility in our COIN framework.

{\bf 11)} As mentioned above, there may be variables in each node's state which,
under one particular interpretation, represent the ``utility
functions'' that the associated microlearner's computer program is
trying to extremize.  When there are such components of
$\underline{\zeta}$, we refer to the utilities they represent as {\bf
private utilities}. However even when there are private utilities,
formally we allow the personal utilities to differ from them. The
personal utility functions \{$g_\eta$\} do not exist ``inside the
COIN''; they are not specified by components of
$\underline{\zeta}$. This separating of the private utilities from the
\{$g_\eta$\} will allow us to avoid the teleological problem that one
may not always be able to explicitly identify ``the'' private utility
function reflected in $\underline{\zeta}$ such that a particular
computational device can be said to be a microlearner ``trying to
increase the value of its private utility''. To the degree that we can
couch the theorems purely in terms of personal rather than private
utilities, we will have successfully adopted a purely behaviorist
approach, without any need to interpret what a computational device is
``trying to do''.

Despite this formal distinction though, often we will implicitly have
in mind deploying the personal utilities onto the microlearners as
their private utilities, in which case the terms can usually be used
interchangeably. The context should make it clear when this is the
case.

\subsubsection{Intelligence}

We will need to quantify how well the entire system performs in terms
of $G$. To do this requires a measure of the performance of an
arbitrary worldline $\underline{\zeta}$, for an arbitrary utility
function, under arbitrary dynamic laws $C$. Formally, such a measure
is a mapping from three arguments to $\cal{\bf{R}}$. 

Such a measure will also allow us to quantify how well each
microlearner performs in purely behavioral terms, in terms of its
personal utility.  (In our behaviorist approach, we do not try to make
specious distinctions between whether a microlearner's performance is
due to its level of ``innate sophistication'', or rather due to dumb
luck --- all that matters is the quality of its behavior as reflected
in its utility value for the system's worldline.)  This behaviorism in
turn will allow us to avoid having private utilities explicitly arise
in our theorems (although they still arise frequently in pedagogical
discussion). Even when private utilities exist, there will be no
formal need to explicitly identify some components of
$\underline{\zeta}$ as such utilities.  Assuming a node's microlearner
is competent, the fact that it is trying to optimize some particular
private utility $U$ will be manifested in our performance measure's
having a large value at $\underline{\zeta}$ for $C$ for that utility
$U$.

The problem of how to formally define such a performance measure is
essentially equivalent to the problem of how to quantify bounded
rationality in game theory. Some of the relevant work in game theory,
for example that involving `trembling hand equilibria' or `$\epsilon$
equilibria' \cite{fr86} is concerned with refinements or modifications
of Nash equilibria (see also ~\cite{kurz96}). Rather than a
behaviorist approach, such work adopts a strongly teleological
perspective on rationality. In general, such work is only applicable
to those situations where the rationality is bounded due to the
precise causal mechanisms investigated in that work. Most of the other
game-theoretic work first models (!) the microlearner, as some
extremely simple computational device ({\em e.g.}, a deterministic
finite automaton (DFA). One then assumes that the microlearner
performs perfectly for that device, so that one can measure that
learner's performance in terms of some computational capacity measure
of the model ({\em e.g.}, for a DFA, the number of states of that DFA)
\cite{fule98,neym85,sale97}.\footnote{Some of the more popular
model-based scenarios for investigating bounded rationality, like
`ficticious play' (see the game theory section above), do not
even stipulate one particular way to quantify that rationality.}
However, if taken as renditions of real-world computer-based
microlearners --- never mind human microlearners--- the models in this
approach are often extremely abstracted, with many important
characteristics of the real learners absent or distorted. In addition,
there is little reason to believe that any results arising from this
approach would not be highly dependent on the model choice and on the
associated representation of computational capacity. Yet another
disadvantage is that this approach concentrates on perfect, fully
rational behavior of the microlearners, within their computational
restrictions.

We would prefer a less model-dependent approach, especially given our
wish that the performance measure be based solely on the utility
function at hand, $\underline{\zeta}$, and $C$. Now we don't want our
performance measure to be a ``raw'' utility value like
$g_{\eta}(\underline{\zeta})$, since that is not invariant with
respect to monotonic transformations of $g_{\eta}$. Similarly, we
don't want to penalize the microlearner for not achieving a certain
utility value if that value was impossible to achieve not due to the
microlearner's shortcomings, but rather due to $C$ and the actions of
other nodes.  A natural way to address these concerns is to generalize
the game-theoretic concept of ``best-response strategy'' and consider
the problem of how well $\eta$ performs {\it given the actions of the
other nodes}.  Such a measure would compare the utility ultimately
induced by each of the possible states of $\eta$ at some particular
time, which without loss of generality we can take to be 0, to that
induced by the actual state $\underline{\zeta}_{\eta,0}$. In other
words, we would compare the utility of the actual worldline
$\underline{\zeta}$ to those of a set of alternative worldlines
$\underline{\zeta}'$, where $\underline{\zeta}_{\;\hat{ }\eta,0} =
\underline{\zeta}'_{\;\hat{ }\eta,0}$, and use those comparisons to
quantify the quality of $\eta$'s performance.

Now we are only concerned with comparing the effects of replacing
$\underline{\zeta}$ with $\underline{\zeta}'$ on $future$
contributions to the utility. But if we allow arbitrary
$\underline{\zeta}'_{,t<0}$, then in and of themselves the difference
between those past components of $\underline{\zeta}'$ and those of
$\underline{\zeta}$ can modify the value of the utility, regardless of
the effects of any difference in the future components. Our
presumption is that for many COINs of interest we can avoid this
conundrum by restricting attention to those $\underline{\zeta}'$ where
$\underline{\zeta}'_{,t<0}$ differs from $\underline{\zeta}_{,t<0}$
only in the internal parameters of $\eta$'s microlearner, differences
that only at times $t \ge 0$ manifest themselves in a form the utility
is concerned with. (In game-theoretic terms, such ``internal
parameters'' encode full extensive-form strategies, and we only
consider changes to the vertices at or below the $t= 0$ level in the
tree of an extensive-form strategy.)

Although this solution to our conundrum is fine when we can apply it,
we don't want to restrict the formalism so that it can only concern
systems having computational algorithms which involve a clearly
pre-specified set of extensive strategy ``internal parameters'' and
the like.  So instead, we formalize our presumption behaviorally, even
for computational algorithms that do not have explicit extensive
strategy internal parameters. Since changing the internal parameters
doesn't affect the $t<0$ components of $\underline{\zeta}_{\eta,}$
{\it that the utility is concerned with}, and since we are only
concerned with changes to $\underline{\zeta}$ that affect the utility,
we simply elect to not change the $t<0$ values of the internal
parameters of $\underline{\zeta}_{\eta,}$ at all. In other words, we
leave $\underline{\zeta}_{\eta,t<0}$unchanged. The advantage of this
stipulation is that we can apply it just as easily whether $\eta$ does
or doesn't have any ``internal parameters'' in the first place.

So in quantifying the performance of $\eta$ for behavior given by
$\underline{\zeta}$ we compare $\underline{\zeta}$ to a set of
$\underline{\zeta}'$, a set restricted to those $\underline{\zeta}'$
sharing $\underline{\zeta}$'s past: $\underline{\zeta}'_{,t<0} =
\underline{\zeta}_{,t<0}$, $\underline{\zeta}'_{\;\hat{ }\eta,0} =
\underline{\zeta}_{\;\hat{ }\eta, 0}$, and $\underline{\zeta}'_{,t \ge
0} \in C_{,t \ge 0}$. Since $\underline{\zeta}'_{\eta,0}$ is free to
vary (reflecting the possible changes in the state of $\eta$ at time
0) while $\underline{\zeta}'_{,t<0}$ is not, $\underline{\zeta}'
\notin C$, in general. We may even wish to allow
$\underline{\zeta}'_{,t\ge0} \notin C_{,t\ge0}$ in certain
circumstances. (Recall that $C$ may reflect other restrictions imposed
on allowed worldlines besides adherence to the underlying dynamical
laws, so simply obeying those laws does not suffice to ensure that a
worldline lies on $C$.) In general though, our presumption is that as
far as utility values are concerned, considering these dynamically
impossible $\underline{\zeta}'$ is equivalent to considering a more
restricted set of $\underline{\zeta}'$ with ``modified internal
parameters'', all of which are $\in C$.

We now present a formalization of this performance measure. Given $C$
and a measure $d\mu(\underline{\zeta}_{,0})$ demarcating what points
in $\underline{\bf Z}_{\eta,0}$ we are interested in, we define the
($t=0$) {\bf intelligence} for node $\eta$ of a point
$\underline{\zeta}$ with respect to a utility $U$ as follows:
\begin{equation} \epsilon_{\eta,U}(\underline{\zeta}) \equiv
 \int d\mu(\underline{\zeta}'_{,0}) \;\;
	\Theta[U( \underline{\zeta} )
			-
		U( \underline{\zeta}_{,t<0} \bullet
		   C(\underline{\zeta}'_{,0}) )]
\cdot \delta(\underline{\zeta}'_{\;\hat{}\eta,0} - \underline{\zeta}_{\;\hat{}\eta,0})
\end{equation}
where $\Theta(.)$ is the Heaviside theta function which equals 0 if
its argument is below 0 and equals 1 otherwise, $\delta(.)$ is the
Dirac delta function, and we assume that
$\int d\mu(\underline{\zeta}'_{\eta,0}) = 1$.

Intuitively, $\epsilon_{\eta,U}(\underline{\zeta})$ measures the
fraction of alternative states of $\eta$ which, if $\eta$ had been in
those states at time 0, would either degrade or not improve $\eta$'s
performance (as measured by $U$). Sometimes in practice we will only
want to consider changes in those components of
$\underline{\zeta}_{\eta,0}$ that we consider as ``free to vary'',
which means in particular that those changes are consistent with $C$
and the state of the external world,
$\underline{\zeta}_{^\eta,0}$. (This consistency ensures that $\eta$'s
observational information concerning the external world is correct;
see Observation 7.1 above.) Such a restriction means that even though
$\underline{\zeta}_{,0}$ may not be consistent with $C$ and
$\underline{\zeta}_{,t<0}$, by itself it is still consistent with $C$;
in quantifying the quality of a particular
$\underline{\zeta}_{\eta,0}$. So we don't compare our point to other
$\underline{\zeta}_{,0}$ that are physically impossible, no matter
what the past is. Any such restrictions on what changes we are
considering are reflected implicitly in intelligence, in the measure
$d\mu$.

As an example of intelligence, consider the situation where for each
player $\eta$, the support of the measure
$d\mu(\underline{\zeta}'_{,0})$ extends over all possible actions that
$\eta$ could take that affect the ultimate value of its personal
utility, $g_\eta$. In this situation we recover conventional full
rationality game theory involving Nash equilibria, as the analysis of
scenarios in which the intelligence of each player $\eta$ with respect
to $g_\eta$ equals 1.\footnote{As an alternative to such fully
rational games, one can define a bounded rational game as one in which
the intelligences equal some vector $\vec{\epsilon}$ whose components
need not all equal 1.  Many of the theorems of conventional game
theory can be directly carried over to such bounded-rational games
\cite{wolp99b} by redefining the utility functions of the players. In
other words, much of conventional full rationality game theory applies
even to games with bounded rationality, under the appropriate
transformation. This result has strong implications for the legitimacy
of the common criticism of modern economic theory that its assumption
of full rationality does not hold in the real world, implications that
extend significantly beyond the Sonnenschein-Mantel-Debreu Theorem
equilibrium aggregate demand theorem ~\cite{mawh95}.} As an
alternative, we could for each $\eta$ restrict
$d\mu(\underline{\zeta}'_{,0})$ to some limited ``set of actions that
$\eta$ actively considers''. This provides us with an ``effective Nash
equilibrium'' at the point $\underline{\zeta}$ where each
$\epsilon_{\eta,g_\eta}(\underline{\zeta})$ equals 1, in the sense that
{\it as far it's concerned}, each player $\eta$ has played a best
possible action at such a point. As yet another alternative, we could
restrict each $d\mu(\underline{\zeta}'_{,0})$ to some infinitesimal
neighborhood about $\underline{\zeta}_{,0}$, and thereby define a
``local Nash equilibrium'' by having
$\epsilon_{\eta,g_\eta}(\underline{\zeta}) = 1$ for each player $\eta$.

In general, competent greedy pursuit of private utility $U$ by the
microlearner controlling node $\eta$ means that the intelligence of
$\eta$ for personal utility $U$,
$\epsilon_{\eta,U}(\underline{\zeta})$, is close to 1. Accordingly, we
will often refer interchangeably to a capable microlearner's
``pursuing private utility $U$'', and to its having high intelligence
for personal utility $U$.  Alternatively, if the microlearner for node
$\eta$ is incompetent, then it may even be that ``by luck'' its
intelligence for some personal utility \{$g_{\eta}$\} exceeds its
intelligence for the different private utility that it's actually
trying to maximize, $U_{\eta}$.

Say that we expect that a particular microlearner is ``smart'', in
that it is more likely to have high rather than low intelligence. We
can model this by saying that given a particular
$\underline{\zeta}_{\;\hat{}\eta,0}$, the conditional probability that
$\underline{\zeta}_{\eta,0} = z$ is a monotonically increasing
function of $\epsilon_{\eta,g_{\eta}}(\underline{\zeta}_{,t<0} \bullet
C(z \bullet \underline{\zeta}_{\;\hat{}\eta,0}))$. Since for a given
$\underline{\zeta}_{\;\hat{}\eta,0}$ the intelligence
$\epsilon_{\eta,g_{\eta}}$ is a monotonically increasing function of
$g_{\eta}$, this modelling assumption means that the probability that
$\underline{\zeta}_{\eta,0} = z$ is a monotonically increasing
function of $g_{\eta}(\underline{\zeta}_{,t<0} \bullet C(z \bullet
\underline{\zeta}_{\;\hat{}\eta,0}))$. An alternative weaker model is to
only stipulate that the probability of having a particular pair
$(\underline{\zeta}_{\eta,0}, \underline{\zeta}_{\;\hat{}\eta,0})$ with
$\epsilon_{\eta,g_{\eta}}$ equal to $z$ is a monotonically increasing
function of $z$. (This probability is an integral over a joint
distribution, rather than a conditional distribution, as in the
original model.) In either case, the ``better'' the microlearner, the
more tightly peaked the associated probability distribution over
intelligence values is.

Any two utility functions that are related by a monotonically
increasing transformation reflect the same preference ordering over
the possible arguments of those functions. Since it is only that
ordering that we are ever concerned with, we would like to remove this
degeneracy by ``normalizing'' all utility functions. In other words,
we would like to reduce any equivalence set of utility functions that
are monotonic transformations of one another to a canonical member of
that set. To see what this means in the COIN context, fix
$\underline{\zeta}_{\;\hat{}\eta,}$. Viewed as a function from
$\underline{\bf Z}_{\eta,} \rightarrow \cal{R}$,
$\epsilon_{\eta,U}(\underline{\zeta}_{\;\hat{}\eta,}, .)$ is itself a
utility function, one that is a monotonically increasing function of
$U$. (It says how well $\eta$ would have performed for all vectors
$\underline{\zeta}_{\eta,}$.) Accordingly, the integral transform
taking $U$ to $\epsilon_{\eta,U}(\underline{\zeta}_{\;\hat{}\eta,}, .)$
is a (contractive, non-invertible) mapping from utilities to
utilities. Applied to any member of a utility in $U$'s equivalence
set, this mapping produces the same image utility, one that is also in
that equivalence set. It can be proven that any mapping from utilities
to utilities that has this and certain other simple properties must be
such an integral transform.  In this, intelligence is the unique way
of ``normalizing'' Von Neumann utility functions.

For those conversant with game theory, it is worth noting some of the
interesting aspects that ensue from this normalizing nature of
intelligences.  At any point $\underline{\zeta}$ that is a Nash
equilibrium in the set of personal utilities \{$g_\eta$\}, all
intelligences $\epsilon_{\eta,g_\eta}(\underline{\zeta})$ must equal
1. Since that is the maximal value any intelligence can take on, a
Nash equilibrium in the \{$g_\eta$\} is a Pareto optimal point in the
associated intelligences (for the simple reason that no deviation from
such a $\underline{\zeta}$ can raise any of the
intelligences). Conversely, if there exists at least one Nash
equilibrium in the \{$g_\eta$\}, then there is not a Pareto optimal
point in the \{$\epsilon_{\eta,g_\eta}(\underline{\zeta})$\} that is
not a Nash equilibrium.

Now restrict attention to systems with only a single instant of time,
{\em i.e.}, single-stage games. Also have each of the (real-valued)
components of each $\underline{\zeta}_\eta$ be a mixing component of
an associated one of $\eta$'s potential strategies for some underlying
finite game. Then have $g_\eta(\underline{\zeta})$ be the associated
expected payoff to $\eta$. (So the payoff to $\eta$ of the underlying
pure strategies is given by the values of $g_\eta(\underline{\zeta})$
when $\underline{\zeta}$ is a unit vector in the space
$\underline{\bf{Z}}_\eta$ of $\eta$'s possible states.) Then we know
that there must exist at least one Nash equilibrium in the
\{$g_\eta$\}. Accordingly, in this situation the set of Nash
equilibria in the \{$g_\eta$\} is identical to the set of points that
are Pareto optimal in the associated intelligences. (See Eq. 5 in the
discussion of factored systems below.)

\subsubsection{Learnability} 
\label{sec:learnability}
Intelligence can be a difficult quantity to work with,
unfortunately. As an example, fix $\eta$, and consider any (small
region centered about some) $\underline{\zeta}$ along with some
utility $U$, where  $\underline{\zeta}$ is not a local
maximum of $U$. Then by increasing the values $U$ takes on
in that small region we will increase the intelligence
$\epsilon_{\eta,U}(\underline{\zeta})$. However in doing this we will
also necessarily $decrease$ the intelligence at points outside that
region.  So intelligence has a non-local character, a character that
prevents us from directly modifying it to ensure that it is
simultaneously high for any and all $\underline{\zeta}$.

A second, more general problem is that without specifying the details
of a microlearner, it can be extremely difficult to predict which of
two private utilities the microlearner will be better able to
learn. Indeed, even $with$ the details, making that prediction can be
nearly impossible. So it can be extremely difficult to determine what
private utility intelligence values will accrue to various choices of
those private utilities. In other words, macrolearning that involves
modifying the private utilities to try to directly increase
intelligence with respect to those utilities can be quite difficult.

Fortunately, we can circumvent many of these difficulties by using a
proxy for (private utility) intelligence. Although we expect its value
usually to be correlated with that of intelligence in practice, this
proxy does not share intelligence's non-local nature.  In addition,
the proxy does not depend heavily on the details of the microlearning
algorithms used, {\em i.e.}, it is fairly independent of those aspects of
$C$. Intuitively, this proxy can be viewed as a ``salient
characteristic'' for intelligence.

We motivate this proxy by considering having $g_\eta = G$ for all
$\eta$.  If we try to actually use these \{$g_{\eta}$\} as the
microlearners' private utilities, particularly if the COIN is large,
we will invariably encounter a very bad signal-to-noise problem. For
this choice of utilities, the effects of the actions taken by node
$\eta$ on its utility may be ``swamped'' and effectively invisible,
since there are so many other processes going into determining $G$'s
value. This makes it hard for $\eta$ to discern the echo of its
actions and learn how to improve its private utility. It also means
that $\eta$ will find it difficult to decide how best to act once
learning has completed, since so much of what's important to $\eta$ is
decided by processes outside of $\eta$s immediate purview. In such a
scenario, there is nothing that $\eta$'s microlearner can do to
reliably achieve high intelligence.\footnote{This ``signal-to-noise''
problem is actually endemic to reinforcement learning as a whole, even
sometimes occurring when one has just a single reinforcement learner,
and only a few random variables jointly determining the value of the
rewards~\cite{wone99}.}

In addition to this ``observation-driven'' signal/noise problem, there
is an ``action-driven'' one. For reasons discussed in Observation 7.1
above, we can define a distribution
$d\mu(\underline{\zeta'}_{\;\hat{}\eta,0})$ reflecting what $\eta$
does/doesn't know concerning the actual state of the outside world
$\hat{}\eta$ at time 0. If the node $\eta$ chooses its actions in a
Bayes-optimal manner, then $\underline{\zeta}_{\eta,0;act} = {\rm
argmax}_z [\int d\mu(\underline{\zeta'}_{\;\hat{}\eta,0})
U(\underline{\zeta}_{,t<0} \bullet C(z \bullet
\underline{\zeta}_{\eta,0;act} \bullet
\underline{\zeta'}_{\;\hat{}\eta,0}))]$, where $z$ runs over the
allowed action components of $\eta$ at time 0. Since this will differ
from ${\rm argmax}_z U(\underline{\zeta}_{,t<0} \bullet C(z \bullet
\underline{\zeta}_{\eta,0;act} \bullet
\underline{\zeta}_{\;\hat{}\eta,0}))]$ in general, this Bayes-optimal
node's intelligence will be less than 1 for the particular
$\underline{\zeta}$ at hand, in general. Moreover, the less $U$'s
ultimate value (after the application of $C$, etc.) depends on
$\underline{\zeta}_{\;\hat{}\eta,0}$, the smaller the difference in
these two argmax-based $z$'s, and therefore the higher the
intelligence of $\eta$, in general.\footnote{In practice, due to computational
limitations if nothing else, the node won't be exactly
Bayes-optimal. But incorporating such a suboptimality doesn't affect
the thrust of this argument that we want $U$'s ultimate value to not
depend on $\underline{\zeta}_{\;\hat{}\eta,0}$.}

We would like a measure of $U$ that captures these efects, but without
depending on function maximization or any other detailed aspects of how the
node determines its actions.  One natural way to do this is via the
{\bf (utility) learnability}: Given a measure
$d\mu(\underline{\zeta}'_{,0})$ restricted to a manifold $C$, the ($t=
0$) utility learnability of a utility $U$ for a node $\eta$ at
$\underline{\zeta}$ is:
\begin{equation}
\Lambda_{\eta,U}(\underline{\zeta}) \equiv
\frac
  {\int d\mu(\underline{\zeta}'_{,0}) \;\; 
  |U(\underline{\zeta}_{,t<0} \bullet
     C(\underline{\zeta}_{\;\hat{}\eta,0} \bullet \underline{\zeta}'_{\eta,0})) 
				- U(\underline{\zeta})|}
  {\int d\mu(\underline{\zeta}'_{,0}) \;\; 
  |U(\underline{\zeta}_{,t<0} \bullet
     C(\underline{\zeta}'_{\;\hat{}\eta,0} \bullet \underline{\zeta}_{\eta,0})) 
				- U(\underline{\zeta})|} \; .
\end{equation}
{\bf Intelligence learnability} is defined the same way, with $U(.)$
replaced by $\epsilon_{\eta,U}(.)$.  Note that any affine transformation
of $U$ has no effect on either the utility learnability 
$\Lambda_{\eta,U}(\underline{\zeta})$ or the associated intelligence
learnability, $\Lambda_{\eta,\epsilon_{\eta,U}}(\underline{\zeta})$.

The integrand in the numerator of the definition of learnability
reflects how much of the change in $U$ that results from replacing
$\underline{\zeta}_{,0}$ with $\underline{\zeta}'_{,0}$ is due to the
change in $\eta$'s $t=0$ state (the ``signal''). The denominator
reflects how much of the change in $U$ that results from replacing
$\underline{\zeta}$ with $\underline{\zeta}'$ is due to the change in
the $t = 0$ states of nodes other than $\eta$ (the ``noise''). So
learnability quantifies how easy it is for the microlearner to discern
the ``echo'' of its behavior in the utility function $U$. Our
presumption is that the microlearning algorithm will achieve higher
intelligence if provided with a more learnable private utility.

Intuitively, the (utility) {\bf differential learnability} of $U$ at a
point $\underline{\zeta}$ is the learnability with $d\mu$ restricted
to an infinitesimal ball about $\underline{\zeta}$. We formalize it as
the following ratio of magnitudes of a pair of gradients, one
involving $\eta$, and one involving $\hat{}\eta$:
\begin{equation}
\lambda_{\eta,U}(\underline{\zeta}) \equiv
\frac{
||\partial_{\underline{\zeta}_{\eta,0}} U(\underline{\zeta}_{,t<0}
\bullet C(\underline{\zeta}_{,0}))||
}
{||\partial_{\underline{\zeta}_{\;\hat{}\eta,0}} U(\underline{\zeta}_{,t<0}
\bullet C(\underline{\zeta}_{,0}))||
} \;.
\end{equation}

Note that a particular value of differential utility learnability, by
itself, has no significance. Simply rescaling the units of
$\underline{\zeta}_{\eta,0}$ will change that value. Rather what is
important is the ratio of differential learnabilities, at the same
$\underline{\zeta}$, for different $U$'s. Such a ratio quantifies the
relative preferability of those $U$'s.

One nice feature of differential learnability is that unlike
learnability, it does not depend on choice of some measure
$d\mu(.)$. This independence can lead to trouble if one is not careful
however, and in particular if one uses learnability for purposes other
than choosing between utility functions. For example, in some
situations, the COIN designer will have the option of enlarging the
set of variables from the rest of the COIN that are ``input'' to some
node $\eta$ at $t = 0$ and that therefore can be used by $\eta$ to
decide what action to take.  Intuitively, doing so will not affect the
RL ``signal'' for $\eta$'s microlearner (the magnitude of the
potential ``echo'' of $\eta$'s actions are not modified by changing
some aspect of how it chooses among those actions). However it {\it
will} reduce the ``noise'', in that $\eta$'s microlearner now knows
more about the state of the rest of the system.

In the full integral version of learnability, this effect can be
captured by having the support of $d\mu(.)$ restricted to reflect the
fact that the extra inputs to $\eta$ at $t=0$ are correlated with the
$t=0$ state of the external system. In differential learnability
however this is not possible, precisely because no measure $d\mu(.)$
occurs in its definition.  So we must capture the reduction in noise
in some other fashion.\footnote{One way to capture this noise
reduction is to replace the noise term
$\partial_{\underline{\zeta}_{\;\hat{}\eta,0}}
U(\underline{\zeta}_{,t<0} \bullet C(\underline{\zeta}_{,0}))$
occurring in the definition of differential learnability with
something more nuanced.  For example, one may wish to replace it with
the maximum of the dot product of $\underline{u} \equiv
\partial_{\underline{\zeta}_{,}}U(\underline{\zeta})$ with any
$\underline{\bf{Z}}$ vector $\underline{v}$, subject not only to the
restrictions that $||\underline{v}|| = 1$ and $\underline{v}_{\eta,0}
=$ {\bf 0}, but also subject to the restriction that $\underline{v}$
must lie in the tangent plane of $C$ at $\underline{\zeta}$. The first
two restrictions, in concert with the extra restriction that
$\underline{v}_{,t<0} =$ {\bf 0}, give the original definition of the
noise term. If they are instead joined with the third, new
restriction, they will enforce any applicable coupling between the
state of $\eta$ at time 0 and the rest of the system at time
0. Solving with Lagrange multipliers, we get $\underline{v} \propto
\underline{u} - \lambda_2 \underline{\alpha} - \lambda_3
\underline{\beta}$, where $\underline{\alpha}$ is the normal to $C$ at
$\underline{\zeta}$, $\underline{\beta}_{\eta',t'} \equiv
\delta_{\eta',\eta} \; \delta_{t',t}$, and $\lambda_2 = \frac
{\underline{\alpha} \cdot \underline{u} - (\underline{\beta} \cdot
\underline{u}) (\underline{\alpha} \cdot \underline{\beta})} {1 -
(\underline{\alpha} \cdot \underline{\beta})^2}$ while $\lambda_3 =
\frac{\underline{\beta} \cdot \underline{u} - (\underline{\alpha}
\cdot \underline{u})(\underline{\alpha} \cdot \underline{\beta})} {1 -
(\underline{\alpha} \cdot \underline{\beta})^2}$. As a practical
matter though, it is often simplest to assume that the
$\underline{\zeta}_{\;\hat{}\eta,0}$ can vary arbitrarily, independent
of $\underline{\zeta}_{\eta,0}$, so that the noise term takes the form
in Eq. 3.}

Alternatively, if the extra variables are being input to $\eta$ for
all $t \ge 0$, not just at $t = 0$, and if $\eta$ ``pays attention''
to those variables for all $t \ge 0$, then by incorporating those
changes into our system $C$ itself has changed, $\forall \; t \ge
0$. Hypothesize that at those $t$ the node $\eta$ is capable of
modifying its actions to ``compensate'' for what (due to our
augmentation of $\eta$'s inputs) $\eta$ now knows to be going on
outside of it. Under this hypothesis, those changes in those external
events will have less of an effect on the ultimate value of $g_\eta$
than they would if we had not made our modification. In this
situation, the noise term has been reduced, so that the differential
learnabiliity properly captures the effect of $\eta$'s having more
inputs.

Another potential danger to bear in mind concerning differential
learnability is that it is usually best to consider its average over a
region, in particular over points with less than maximal
intelligence. It is really designed for such points; in fact, at the
intelligence-maximizing $\underline{\zeta}$,
$\lambda_{\eta,U}(\underline{\zeta}) = 0$.

Whether in its differential form or not, and whether
referring to utilities or intelligence, learnability is not meant to
capture all factors that will affect how high an intelligence value a
particular microlearner will achieve. Such an all-inclusive definition
is not possible, if for no other reason the fact that there are many
such factors that are idiosyncratic to the particular microlearner
used. Beyond this though, certain more general factors that affect
most popular learning algorithms, like the curse of dimensionality,
are also not (explicitly) designed into learnability. Learnability is
not meant to provide a full characterization of performance --- that
is what intelligence is designed to do.  Rather (relative)
learnability is ony meant to provide a {\it guide} for how to improve
performance.

A system that has infinite (differential, intelligence) learnability
for all its personal utilities is said to be ``perfectly''
(differential, intelligence) learnable. It is straight-forward to
prove that a system is perfectly learnable $\forall \zeta \in C$ iff
$\forall \eta, \forall \underline{\zeta} \in C,
g_\eta(\underline{\zeta})$ can be written as
$\psi_\eta(\underline{\zeta}_{\eta,0})$ for some function
$\psi_\eta(.)$. (See the discussion below on the general condition for
a system's being perfectly factored.)

\subsection{A descriptive framework for COINs}
With these definitions in hand, we can now present (a portion of) one
descriptive framework for COINs. In this subsection, after
discussing salient characteristics in general, we present some theorems
concerning the relationship between personal utilities and the
salient characteristic we choose to concentrate on. We then discus how
to use these theorems to induce that salient characteristic in a COIN.

\subsubsection{Candidate salient characteristics of a COIN}
The starting point with a descriptive framework is the identification
of ``salient characteristics of a COIN which one strongly expects to
be associated with its having large world utility''. In this chapter
we will focus on salient characteristics that concern the relationship
between personal and world utilities.  These characteristics are
formalizations of the intuition that we want COINs in which the
competent greedy pursuit of their private utilities by the
microlearners results in large world utility, without any bottlenecks,
TOC, ``frustration'' (in the spin glass sense) or the like.

One natural candidate for such a characteristic, related to Pareto
optimality ~\cite{fule98,futi91}, is {{\bf weak triviality}}. It is
defined by considering any two worldlines $\underline{\zeta}$ and
$\underline{\zeta}'$ both of which are consistent with the system's
dynamics ({\em i.e.}, both of which lie on $C$), where for every node
$\eta$, $g_{\eta}(\underline{\zeta}) \ge
g_{\eta}(\underline{\zeta}')$.\footnote{An obvious variant is to
restrict $\underline{\zeta}'_{,t<0} = \underline{\zeta}_{,t<0}$, and
require only that both of the ``partial vectors''
$\underline{\zeta}'_{,t\ge0}$ and $\underline{\zeta}_{,t\ge0}$ obey
the relevant dynamical laws, and therefore lie in $C_{,t\ge0}$.}  If
for any such pair of worldlines where one ``Pareto dominates'' the
other it is necessarily true that $G(\underline{\zeta}) \ge
G(\underline{\zeta}')$, we say that the system is weakly trivial. We
might expect that systems that are weakly trivial for the
microlearners' private utilities are configured correctly for inducing
large world utility. After all, for such systems, if the microlearners
collectively change $\underline{\zeta}$ in a way that ends up helping
all of them, then necessarily the world utility also rises. More
formally, for a weakly trivial system, the maxima of $G$ are
Pareto-optimal points for the personal utilities (although the reverse
need not be true).

As it turns out though, weakly trivial systems can readily evolve to a
world utility $minimum$, one that often involves TOC. To see this,
consider automobile traffic in the absence of any traffic control
system. Let each node be a different driver, and say their private
utilities are how quickly they each individually get to their
destination. Identify world utility as the sum of private
utilities. Then by simple additivity, for all $\underline{\zeta}$ and
$\underline{\zeta}'$, whether they lie on $C$ or not, if
$g_\eta(\underline{\zeta}) \ge g_\eta(\underline{\zeta}') \;\;\forall
\eta$ it follows that $G(\underline{\zeta}) \ge
G(\underline{\zeta}')$; the system is weakly trivial. However as any
driver on a rush-hour freeway with no carpool lanes or metering lights
can attest, every driver's pursuing their own goal definitely does not
result in acceptable throughput for the system as a whole;
modifications to private utility functions (like fines for violating
carpool lanes or metering lights) would result in far better global
behavior. A system's being weakly trivial provides no assurances
regarding world utility.

This does not mean weak triviality is never of use. For example, say
that for a set of weakly trivial personal utilities each agent can
guarantee that {\it regardless of what the other agents do}, its
utility is above a certain level. Assume further that, being
risk-averse, each agent chooses an action with such a guarantee. Say
it is also true that the agents are provided with a relatively large
set of candidate guaranteed values of their utilities. Under these
circumstances, the system's being weakly trivial provides some
assurances that world utility is not too low. Moreover, if the
overhead in enforcing such a future-guaranteeing scheme is small, and
having a sizable set of guaranteed candidate actions provided to each
of the agents does not require an excessively centralized
infrastructure, we can actually employ this kind of scheme in
practice.  Indeed, in the extreme case, one can imagine that every
agent is guaranteed exactly what its utility would be for every one of
its candidate actions. (See the discussion on General Equilibrium in
the Background Section above.) In this situation, Nash equilibria and
Pareto optimal points are identical, which due to weak triviality
means that the point maximizing $G$ is a Nash equilibrium. However in
any less extreme situation, the system may not achieve a value of
world utility that is close to optimal. This is because even for
weakly trivial systems a Pareto optimal point may have poor world
utility, in general.

Situations where one has guarantees of lower bounds on one's utility
are not too common, but they do arise.  One important example is a
round of trades in a computational market (see the Background Section
above). In that scenario, there is an agent-indexed set of functions
\{$f_\eta(\bf{z} \in \underline{\bf{Z}}^{(0)})$\} and the personal
utility of each agent $\eta \in \{1, 2, ...\}$ is given by
$f_\eta(\underline{\zeta}_{\eta,t^*})$, where $t^*$ is the end of the
round of trades. There is also a function $F(\bf{z} \in
{\underline{\bf{Z}}}^{(0)})$ $\equiv F(f_1(\bf{z}), f_2(\bf{z}), ...)$
that is a monotonically increasing function of its arguments, and
world utility $G$ is given by $F(\underline{\zeta}_{,t^*}) =
F(f_1(\underline{\zeta}_{,t^*}), f_2(\underline{\zeta}_{,t^*}), ...)$.
So the system is weakly trivial.  In turn, each $f_\eta({\bf{z}})$ is
determined solely by the ``allotment of goods'' possessed by $\eta$,
as specified in the appropriate components of $z_\eta$.  To be able to
remove uncertainty about its future value of $f_\eta$ in this kind of
system, in determining its trading actions each agent $\eta$ must
employ some scheme like inter-agent contracts. This is because without
such a scheme, no agent can be assured that if it agrees to a proposed
trade with another agent that the full proposed transaction of that
trade actually occurs. Given such a scheme, if in each trade round $t$
each agent $\eta$ myopically only considers those trades that are
assured of increasing the corresponding value of $f_\eta$, then we are
guaranteed that the value of the world utility is not less than the
initial value $F(\underline{\zeta}_{,0})$.

The problem with using weak triviality as a general salient
characteristic is precisely the fact that the individual microlearners
$are$ greedy. In a COIN, there is no system-wide incentive to replace
$\underline{\zeta}$ with a different worldline that would improve
everybody's private utility, as in the definition of weak
triviality. Rather the incentives apply to each microlearner
individually and motivate the learners to behave in a way that may
well hurt some of them. So weak triviality is, upon examination, a
poor choice for the salient characteristic of a COIN.

One alternative to weak triviality follows from consideration of the
stricture that we must `expect' a salient characteristic to be coupled
to large world utility in a running real-world COIN.  What can we
reasonably expect about a running real-world COIN?  We cannot assume
that all the private utilities will have large values --- witness the
traffic example. But we $can$ assume that if the microlearners are
well-designed, each of them will be doing close to as well it can {\it
given the behavior of the other nodes}. In other words, within broad
limits we can assume that the system is more likely to be in
$\underline{\zeta}$ than $\underline{\zeta}'$ if for all $\eta$,
$\epsilon_{\eta,g_\eta}(\underline{\zeta}) \ge
\epsilon_{\eta,g_\eta}(\underline{\zeta}')$. We define a system to be
{{\bf coordinated}} iff for any such $\underline{\zeta}$ and
$\underline{\zeta}'$ lying on $C$, $G(\underline{\zeta}) \ge
G(\underline{\zeta}')$. (Again, an obvious variant is to restrict
$\underline{\zeta}'_{,t<0} = \underline{\zeta}_{,t<0}$, and require
only that both $\underline{\zeta}_{,t\ge0}$ and
$\underline{\zeta}'_{,t\ge0}$ lie in $C_{,t\ge0}$.) Traffic systems
are $not$ coordinated, in general. This is evident from the simple
fact that if all drivers acted as though there were metering lights
when in fact there weren't any, they would each be behaving with lower
intelligence given the actions of the other drivers (each driver would
benefit greatly by changing its behavior by no longer pretending there
were metering lights, etc.). But nonetheless, world utility would be
higher.

\subsubsection{The Salient Characteristic of Factoredness}

Like weak triviality, coordination is intimately related to the
economics concept of Pareto optimality. Unfortunately, there is not
room in this chapter to present the mathematics associated with
coordination and its variants. We will instead discuss a third
candidate salient characteristic of COINs, one which like coordination
(and unlike weak triviality) we can reasonably expect to be associated
with large world utility. This alternative fixes weak triviality not
by replacing the personal utilities \{$g_\eta$\} with the
intelligences \{$\epsilon_{\eta,g_\eta}$\} as coordination does, but
rather by only considering worldlines whose difference at time 0
involves a single node. This results in this alternative's being
related to Nash equilibria rather than Pareto optimality.

Say that our COIN's worldline is $\underline{\zeta}$. Let
$\underline{\zeta}'$ be any other worldline where
$\underline{\zeta}_{,t<0} = \underline{\zeta}'_{,t<0}$, and where
$\underline{\zeta}'_{,t \ge 0} \in C_{,t \ge 0}$. Now restrict
attention to those $\underline{\zeta}'$ where at $t = 0$
$\underline{\zeta}$ and $\underline{\zeta}'$ differ only for node
$\eta$. If for all such $\underline{\zeta}'$
\begin{equation}
sgn[g_\eta(\underline{\zeta}) - g_\eta(\underline{\zeta}_{,t<0} \bullet
C(\underline{\zeta}'_{,0}))] = sgn[G(\underline{\zeta}) -
G(\underline{\zeta}_{,t<0} \bullet C(\underline{\zeta}'_{,0}))] \; ,
\end{equation}
\noindent
and if this is true for all nodes $\eta$, then we say that the COIN is
{{\bf factored}} for all those utilities \{$g_{\eta}$\} (at $\underline{\zeta}$,
with respect to time 0 and the utility $G$).

For a factored system, for any node $\eta$, {\it given the rest of the
system}, if the node's state at $t = 0$ changes in a way that improves
that node's utility over the rest of time, then it necessarily also improves world
utility. Colloquially, for a system that is factored for a particular
microlearner's private utility, if that learner does something that
improves that personal utility, then everything else being equal, it has
also done something that improves world utility. Of two potential
microlearners for controlling node $\eta$ ({\em i.e.}, two potential
$\underline{\zeta}_\eta$) whose behavior until $t=0$ is identical but
which differ there, the microlearner that is smarter with respect to
$g$ will always result in a larger $g$, by definition of
intelligence. Accordingly, for a factored system, the smarter
microlearner is also the one
that results in better $G$. So as long as we have deployed a
sufficiently smart microlearner on $\eta$, we have assured a good $G$
(given the rest of the system). Formally, this is expressed in the
fact \cite{wolp99} that for a factored system, for all nodes $\eta$,
\begin{equation}\label{eq_epsilons}
\epsilon_{\eta,g_\eta}(\underline{\zeta}) =
\epsilon_{\eta,G}(\underline{\zeta}) \; .
\end{equation}
\noindent

One can also prove that Nash equilibria of a factored system are local
maxima of world utility. Note that in keeping with our behaviorist
perspective, nothing in the definition of factored requires the
existence of private utilities. Indeed, it may well be that a system
having private utilities \{$U_{\eta}$\} is factored, but for personal
utilities \{$g_{\eta}$\} that differ from the \{$U_{\eta}$\}.

A system's being factored does $not$ mean that a change to
$\underline{\zeta}_{\eta,0}$ that improves $g_\eta(\underline{\zeta})$
cannot also hurt $g_{\eta'}(\underline{\zeta})$ for some $\eta' \neq
\eta$. Intuitively, for a factored system, the side effects on the
rest of the system of $\eta$'s increasing its own utility do not end
up decreasing world utility --- but can have arbitrarily adverse
effects on other private utilities. (In the language of economics, no
stipulation is made that $\eta$'s ``costs are endogenized.'') For
factored systems, the separate microlearners successfully pursuing
their separate goals do not frustrate each other {\it as far as world
utility is concerned}.

In addition, if $g_{\eta,t'}$ is factored with respect to $G$, then a
change to $\underline{\zeta}_{\eta,t'}$ that improves
$g_{\eta,t'}(\underline{\zeta}_{,t<t'}, C(\underline{\zeta}_{,t'}))$
improves $G(\underline{\zeta}_{,t<t'},
C(\underline{\zeta}_{,t'}))$. But it may $hurt$ some $g_{\eta,t'' \neq
t'}(\underline{\zeta}_{,t<t'}, C(\underline{\zeta}_{,t'}))$ and/or
$\epsilon_{(\eta,t''),g_{\eta,t''}} (\underline{\zeta}_{,t<t'},
C(\underline{\zeta}_{,t'}))$. (This is even true for a discounted sum
of rewards personal utility, so long as $t'' > t'$.) An example of this
would be an economic system cast as a single individual, $\eta$,
together with an environment node, where $G$ is a steeply discounted
sum of rewards $\eta$ receives over his/her lifetime, $t'' > t'$, and
$\forall t$, $g_{\eta,t}(\underline{\zeta}) =
G({\mbox{CL}}_{,t<t'}(\underline{\zeta}))$. For such a situation, it
may be appropriate for $\eta$ to live extravagantly at the time $t'$,
and ``pay for it'' later.

As an instructive example of the ramifications of Eq. 5, say node
$\eta$ is a conventional computer. We want
$\epsilon_{\eta,G}(\underline{\zeta})$ to be as high as possible,
i.e., given the state of the rest of the system at time 0, we want
computer $\eta$'s state then to be the best possible, as far as the
resultant value of $G$ is concerned. Now a computer's ``state''
consists of the values of all its bits, including its code segment,
i.e., including the program it is running. So for a factored personal
utility $g_{\eta}$, if the program running on the computer is better
than most others as far as $g_{\eta}$ is concerned, then it is also
better than most other programs as far as $G$ is concerned.

Our task as COIN designers engaged in COIN initialization or
macrolearning is to find such a program and such an associated
$g_{\eta}$.  One way to approach this task is to restrict attention to
programs that consist of RL algorithms with private utility specified
in the bits \{$b_i$\} of $\eta$. This reduces the task to one of
finding a private utility \{$b_i$\} (and thereby fully specifying
$\underline{\zeta}_{\eta,0}$) such that our RL algorithm working with
that private utility has high $\epsilon_{\eta,g_{\eta}}$, i.e., such
that that algorithm outperforms most other programs as far as the
personal utility $g_\eta$ is concerned.

Perhaps the simplest way to address this reduced task is to exploit
the fact that for a good enough RL algorithm $\epsilon_{\eta,\{b_i\}}$
will be large, and therefore adopt such an RL algorithm and fix the
private utility to equal $g_\eta$. In this way we further reduce the
original task, which was to search over all personal utilities
$g_\eta$ and all programs $R$ to find a pair such that both $g_\eta$
is factored with respect to $G$ and there are relatively few programs
that outperform $R$, as far as $g_\eta$.  The task is now instead to
search over all private utilities \{$b_i$\} such that both \{$b_i$\}
is factored with respect to $G$ and such that there are few programs
({\it of any sort}, RL-based or not) that outperform our RL algorithm
working on \{$b_i$\}, as far as that self-same private utility is
concerned. The crucial assumption being leveraged in this approach is
that our RL algorithm is ``good enough'', and the reason we want
learnable \{$b_i$\} is to help effect this assumption.

In general though, we can't have both perfect learnability and perfect
factoredness. As an example, say that $\forall t,
\underline{\bf{Z}}_{\eta,t} = \underline{\bf{Z}}_{\;\hat{}\eta,t} =
\cal{R}$, and that the dynamics is the identity operator: $\forall t$,
$C(\underline{\zeta}_{,0})_{,t} = \underline{\zeta}_{,0}$. Then if
$G(\underline{\zeta}_{,0}) = \underline{\zeta}_{\eta,0} \cdot
\underline{\zeta}_{\;\hat{}\eta,0}$ and the system is perfectly
learnable, it is not perfectly factored. This is because perfect
learnability requires that $\forall \underline{\zeta} \in C,
g_\eta(\underline{\zeta}) = \psi_\eta(\underline{\zeta}_{\eta,0})$ for
some function $\psi_\eta(.)$.  However any change to
$\underline{\zeta}_{\eta,0}$ that improves such a $g_\eta$ will either
help or $hurt$ $G(\underline{\zeta})$, depending on the sign of
$\underline{\zeta}_{\;\hat{}\eta,0}$. For the ``wrong'' sign of
$\underline{\zeta}_{\;\hat{}\eta,0}$, this means the system is
actually ``anti-factored''. Due to such incompatibility between
perfect factoredness and perfect learnability, we must usually be
content with having high degree of factoredness and high
learnability. In such situations, the emphasis of the macrolearning
process should be more and more on having high degree of factoredness
as we get closer and closer to a Nash equilibrium. This way the system
won't relax to an incorrect local maximum.

In practice of course, a COIN will often not be perfectly
factored. Nor in practice are we always interested only in whether the
system is factored at one particular point (rather than across a
region say).  These issues are discussed in ~\cite{wolp99}, where in
particular a formal definition of of the {\bf degree of factoredness}
of a system is presented.

If a system is factored for utilities $\{g_\eta\}$, then it is also
factored for any utilities  $\{g'_\eta\}$ where for each $\eta$
$g'_\eta$ is a monotonically increasing function of $g_\eta$. More
generally, the following result characterizes the set of all factored
personal utilities:

\noindent
{\bf Theorem 1:} A system is factored at all $\underline{\zeta}
\in C$ iff for all those $\underline{\zeta}$, $\forall \eta$, we can
write
\begin{equation}
g_\eta(\underline{\zeta}) =
\Phi_\eta(\underline{\zeta}_{,t<0}, \underline{\zeta}_{\;\hat{}\eta,0}, G(\underline{\zeta}))
\end{equation}
\noindent
for some function $\Phi_\eta(., ., .)$ such that $\partial_G
\Phi_\eta(\underline{\zeta}_{,t<0},
\underline{\zeta}_{\;\hat{}\eta,0}, G) > 0$ for all $\underline{\zeta}
\in C$ and associated $G$ values. (The form of the \{$g_\eta$\} off of
$C$ is arbitrary.)

\vspace{5mm}
\noindent
{\bf Proof:}
For fixed $\underline{\zeta}_{\eta,0}$ and $\underline{\zeta}_{,t<0}$,
any change to $\underline{\zeta}_{\eta,0}$ which keeps
$\underline{\zeta}_{,t \ge 0}$ on $C$ and which at the same time
increases $G(\underline{\zeta}) = G(\underline{\zeta}_{,t<0} \bullet 
C(\underline{\zeta}_{\;\hat{}\eta,0} \bullet \underline{\zeta}_{\eta,0} ))$
must increase $\Phi_\eta(\underline{\zeta}_{,t<0},
\underline{\zeta}_{\;\hat{}\eta,0}, G(\underline{\zeta}))$, due to the
restriction on $\partial_G \Phi_\eta(\underline{\zeta}_{,t<0},
\underline{\zeta}_{\;\hat{}\eta,0}, G)$. This establishes the backwards
direction of the proof.

\noindent
For the forward direction, write $g_{\eta}(\underline{\zeta}) =
g_{\eta}(\underline{\zeta}, G(\underline{\zeta})) =
g_{\eta}(\underline{\zeta}_{,t<0} \bullet 
C(\underline{\zeta}_{\;\hat{}\eta,0} \bullet \underline{\zeta}_{\eta,0}),
G(\underline{\zeta})) \;\; \forall \; \underline{\zeta} \in C$. Define
this formulation of $g_\eta$ as $\Phi_{\eta}(\underline{\zeta}_{,t<0},
\underline{\zeta}_{,0}, G(\underline{\zeta}))$, which we can
re-express as
$\Phi_{\eta}(\underline{\zeta}_{,t<0},\underline{\zeta}_{\;\hat{}\eta,0} \bullet 
\underline{\zeta}_{\eta,0}, G(\underline{\zeta}))$. Now since the
system is factored, $\forall \underline{\zeta} \in C,
\forall \underline{\zeta}'_{,t \ge 0} \in C_{,t \ge 0}$,
\begin{eqnarray}
\Phi_{\eta}(\underline{\zeta}_{,t<0}, \underline{\zeta}_{\;\hat{}\eta,0} \bullet 
\underline{\zeta}_{\eta,0}, 
G(\underline{\zeta}_{,t<0} \bullet C(\underline{\zeta}_{\;\hat{}\eta,0} \bullet 
\underline{\zeta}_{\eta,0}))) \;\;= \;\;
\Phi_{\eta}(\underline{\zeta}_{,t<0}, \underline{\zeta}_{\;\hat{}\eta,0} \bullet 
\underline{\zeta'}_{\eta,0},  
G(\underline{\zeta}_{,t<0} \bullet C(\underline{\zeta}_{\;\hat{}\eta,0} \bullet 
\underline{\zeta'}_{\eta,0}))) \nonumber 
\end{eqnarray}
\begin{eqnarray}
\iff  \nonumber
\end{eqnarray}
\begin{eqnarray}
G(\underline{\zeta}_{,t<0} \bullet  C(\underline{\zeta}_{\;\hat{}\eta,0} \bullet
\underline{\zeta}_{\eta,0}))) \;\;=\;\; 
G(\underline{\zeta}_{,t<0} \bullet  C(\underline{\zeta}_{\;\hat{}\eta,0} \bullet
\underline{\zeta'}_{\eta,0})) \; . \nonumber
\end{eqnarray}

\noindent
So consider any situation where the system is factored, and the values
of $G$, $\underline{\zeta}_{, t<0}$, and
$\underline{\zeta}_{\;\hat{}\eta,0}$ are specified. Then we can find
{\it any} $\underline{\zeta}_{\eta,0}$ consistent with those values
({\em i.e.}, such that our provided value of $G$ equals
$G(\underline{\zeta}_{,t<0} \bullet C(\underline{\zeta}_{\;\hat{}\eta,0}
\bullet \underline{\zeta}_{\eta,0}))$), evaluate the resulting value
of $\Phi_{\eta}(\underline{\zeta}_{,t<0},
\underline{\zeta}_{\;\hat{}\eta,0} \bullet \underline{\zeta}_{\eta,0},
G)$, and know that we would have gotten the same value if we had found
a different consistent $\underline{\zeta}_{\eta,0}$. This is true for
all $\underline{\zeta} \in C$. Therefore the mapping
$(\underline{\zeta}_{,t<0}, \underline{\zeta}_{\;\hat{}\eta,0}, G)
\rightarrow \Phi_{\eta}$ is single-valued, and we can write
$\Phi_{\eta}(\underline{\zeta}_{,t<0},
\underline{\zeta}_{\;\hat{}\eta,0}, G(\underline{\zeta}))$.  {\bf
QED.}

\vspace{5mm}

By Thm. 1, we can ensure that the system is factored without any
concern for $C$, by having each $g_\eta(\underline{\zeta}) =
\Phi_\eta(\underline{\zeta}_{,t<0},\underline{\zeta}_{\;\hat{}\eta,0},
G(\underline{\zeta})) \;\; \forall \underline{\zeta} \in
\underline{\bf{Z}}$.  Alternatively, by only requiring that $\forall
\underline{\zeta} \in C$ does $g_\eta(\underline{\zeta}) =
\Phi_\eta(\underline{\zeta}_{,t<0},
\underline{\zeta}_{\;\hat{}\eta,0}, G(\underline{\zeta}))$ ({\em i.e.}, does
$g_\eta(\underline{\zeta}_{,t<0} \bullet C(\underline{\zeta}_{,0})) =
\Phi_\eta(\underline{\zeta}_{,t<0}, \underline {\zeta}_{\;\hat{}\eta,0},
G(\underline{\zeta}_{,t<0} \bullet C(\underline{\zeta}_{,0})))$), we can
access a broader class of factored utilities, a class that $does$
depend on $C$. Loosely speaking, for those utilities, we only need the
projection of $\partial_{\underline{\zeta}_{,t\ge0}}
G(\underline{\zeta})$ onto $C_{\eta,0}$ to be parallel to the
projection of $\partial_{\underline{\zeta}_{,t\ge0}}
g_\eta(\underline{\zeta})$ onto $C_{\eta,0}$. Given $G$ and $C$, there
are infinitely many $\partial_{\underline{\zeta}_{,t \ge 0}}
g_\eta(\underline{\zeta})$ having this projection (the set of such
$\partial_{\underline{\zeta}_{,t \ge 0}} g_\eta(\underline{\zeta})$
form a linear subspace of $\underline{\bf{Z}}$).  The partial
differential equations expressing the precise relationship  are
discussed in ~\cite{wolp99}.

As an example of the foregoing, consider a `team game' (also known
as an `exact potential game' \cite{erfl97,mosh96}) in which $g_\eta
= G$ for all $\eta$. Such COINs are factored, trivially, regardless of
$C$; if $g_\eta$ rises, then $G$ must as well, by
definition. (Alternatively, to confirm that team games are factored
just take $\Phi_\eta(\underline{\zeta}_{,t<0},
\underline{\zeta}_{\;\hat{}\eta,0}, G) = G \;\;\forall \eta$ in Thm. 1.)
On the other hand, as discussed below, COINs with `wonderful life'
personal utilities are also factored, but the definition of such
utilities depends on $C$.

\subsubsection{Wonderful life utility}

Due to their often having poor learnability and requiring centralized
communication (among other infelicities), in practice team game
utilities often are poor choices for personal utilities. Accordingly,
it is often preferable to use some other set of factored utilities.
To present an important example, first define the ($t=0$) {\bf effect
set} of node $\eta$ at $\underline{\zeta}$,
$C^{eff}_\eta(\underline{\zeta})$, as the set of all components
$\underline{\zeta}_{\eta',t}$ for which
$\partial_{\underline{\zeta}_{\eta,0}}
(C(\underline{\zeta}_{,0}))_{\eta',t} \neq \vec{0}$.  Define the
effect set $C^{eff}_\eta$ with no specification of $\underline{\zeta}$
as $\cup_{\underline{\zeta} \in C}
C^{eff}_\eta(\underline{\zeta})$. (We take this latter definition to
be the default meaning of ``effect set''.) We will also find it useful
to define $\hat{}C^{eff}_\eta$ as the set of components of the space
$\underline{\bf{Z}}$ that are not in $C^{eff}_\eta$.

Intuitively, $\eta$'s effect set is the set of all components
$\underline{\zeta}_{\eta',t}$ which would be affected by a change in
the state of node $\eta$ at time 0. (They may or may not be affected
by changes in the $t=0$ states of the other nodes.) Note that the
effect sets of different nodes may overlap. The extension of the
definition of effect sets for times other than 0 is immediate. So is
the modification to have effect sets only consist of those components
$\underline{\zeta}_{\eta,t;i}$ that vary with with the state of node
$\eta$ at time 0, rather than consist of the full vectors
$\underline{\zeta}_{\eta,t}$ possessing such a component. These
modifications will be skipped here, to minimize the number of
variables we must keep track of.

Next for any set $\sigma$ of components ($\eta', t$), define
$\mbox{CL}_\sigma(\underline{\zeta})$ as the ``virtual'' vector formed
by clamping the $\sigma$-components of $\underline{\zeta}$ to an
arbitrary fixed value. (In this paper, we take that fixed value to be
$\vec{0}$ for all components listed in $\sigma$.) Consider in
particular a {\bf Wonderful Life} set $\sigma$. The value of the {\bf
wonderful life utility} (WLU for short) for $\sigma$ at
$\underline{\zeta}$ is defined as:
\begin{equation}
WLU_{\sigma}(\underline{\zeta}) \equiv G(\underline{\zeta}) -
G(\mbox{CL}_{\sigma} (\underline{\zeta})) \;
.
\end{equation}
\noindent
In particular, the  WLU for the effect set of node $\eta$ is
$G(\underline{\zeta}) - G(\mbox{CL}_{C^{eff}_\eta}
(\underline{\zeta}))$, which for $\underline{\zeta} \in C$ can be
written as $G(\underline{\zeta}_{,t<0} \bullet C(\underline{\zeta}_{,0})) - 
G(\mbox{CL}_{C^{eff}_\eta}
( \underline{\zeta}_{,t<0} \bullet C(\underline{\zeta}_{,0}) ))$.

We can view $\eta$'s effect set WLU as analogous to the change in
world utility that would have arisen if node $\eta$ ``had never
existed''.  (Hence the name of this utility - cf. the Frank Capra
movie.)  Note however, that $\mbox{CL}$ is a purely ``fictional'',
counter-factual operation, in the sense that it produces a new
$\underline{\zeta}$ without taking into account the system's dynamics.
Indeed, no assumption is even being made that
$\mbox{CL}_{\sigma}(\underline{\zeta})$ is consistent with the
dynamics of the system. The sequence of states the node $\eta$ is
clamped to in the definition of the WLU need not be consistent with
the dynamical laws embodied in $C$.

This dynamics-independence is a crucial strength of the WLU.  It means
that to evaluate the WLU we do {\it not} try to infer how the system
would have evolved if node $\eta$'s state were set to 0 at time 0 and
the system evolved from there. So long as we know $\underline{\zeta}$
extending over all time, and so long as we know $G$, we know the value
of WLU. This is true even if we know nothing of the dynamics of the
system.

An important example is effect set wonderful life utilities when the
set of all nodes is partitioned into `subworlds' in such a way that all
nodes in the same subworld $\omega$ share substantially the same
effect set. In such a situation, all nodes in the same subworld
$\omega$ will have essentially the same personal utilities, exactly as
they would if they used team game utilities with a ``world'' given by
$\omega$. When all such nodes have large intelligence values, this
sharing of the personal utility will mean that all nodes in the same
subworld are acting in a coordinated fashion, loosely speaking.

The importance of the WLU arises from the following results:

\vspace{.5cm}

\noindent
{\bf Theorem 2:} i) A system is factored at all $\underline{\zeta}
\in C$ iff for all those $\underline{\zeta}$, $\forall \eta$, we can
write
\begin{equation}
g_\eta(\underline{\zeta}) =
\hat{\Phi}_\eta(\underline{\zeta}_{\;\hat{}C^{eff}_{\eta}}, G(\underline{\zeta}))
\end{equation}
\noindent
for some function $\hat{\Phi}_\eta(., .)$ such that $\partial_G
\hat{\Phi}_\eta(\underline{\zeta}_{\;\hat{}C^{eff}_{\eta}},
G) > 0$ for all $\underline{\zeta}
\in C$ and associated $G$ values. (The form of the \{$g_\eta$\} off of
$C$ is arbitrary.)

\noindent
ii) In particular, a COIN is factored for personal utilities set
equal to the associated effect set wonderful life utilities.

\vspace{.5cm}
\noindent
{\bf Proof:} To prove (i), first write
$\underline{\zeta}_{\;\hat{}C^{eff}_{\eta}} = \underline{\zeta}_{,t<0}
\bullet \underline{\zeta}_{\;\hat{}\eta,0} \bullet 
\underline{\zeta}_{(\;\hat{}C^{eff}_{\eta})_{,t>0}}$. For all
$\underline{\zeta} \in C$,
$\underline{\zeta}_{(\;\hat{}C^{eff}_{\eta})_{,t > 0}}$ is independent
of $\underline{\zeta}_{\eta,0}$, and so by definition of $C(.)$ it is a
single-valued function of
$\underline{\zeta}_{\;\hat{}\eta,0}$ for such $\underline{\zeta}$. Therefore
$\underline{\zeta}_{\;\hat{}C^{eff}_{\eta}} = \underline{\zeta}_{,t<0}
\bullet \underline{\zeta}_{\;\hat{}\eta,0} \bullet
f(\underline{\zeta}_{\;\hat{}\eta,0})$ for some function
$f(.)$. Accordingly, by Thm. 1, for \{$g_\eta$\} of the
form stipulated in (i), the system is factored. Going the other way, if the
system is factored, then by Thm. 1 it can be written as
$\Phi_\eta(\underline{\zeta}_{,t<0},
\underline{\zeta}_{\;\hat{}\eta,0}, G(\underline{\zeta}))$. Since both
$\underline{\zeta}_{,t<0}$ and $\underline{\zeta}_{\;\hat{}\eta,0}
\not\in C^{eff}_{\eta}$, we can rewrite this as
$\Phi_\eta([\hat{}C^{eff}_{\eta}]_{,t<0},
[\hat{}C^{eff}_{\eta}]_{\;\hat{}\eta,0}, G(\underline{\zeta}))$. {\bf QED.}

\noindent
Part (ii) of the theorem follows immediately from part (i). For
pedagogical value though, here we instead derive it directly. First,
since $\mbox{CL}_{C^{eff}_\eta}(\underline{\zeta})$
is independent of $\underline{\zeta}_{\eta',t}$ for all $(\eta',t) \in
C^{eff}_\eta$, so is the $\underline{\bf Z}$ vector
$\mbox{CL}_{ C^{eff}_\eta } (
\underline{\zeta}_{,t<0} \bullet C(\underline{\zeta}_{,0}) )$, {\em i.e.},
$\partial_{\underline{\zeta}_{\eta,0}}[ \mbox{CL}_{
C^{eff}_\eta } ( \underline{\zeta}_{,t<0} \bullet
C(\underline{\zeta}_{,0})) ]_{\eta',t} = \vec{0} \;\; \forall (\eta',
t) \in C^{eff}_\eta$.  This
means that viewed as a  $\underline{\zeta}_{,t<0}$-parameterized
function from $C_{,0}$ to $\underline{\bf Z}$,
$\mbox{CL}_{ C^{eff}_\eta } (
\underline{\zeta}_{,t<0} \bullet C(.) )$ is a single-valued function
of the
$\underline{\zeta}_{\;\hat{},\eta,0}$ components. Therefore $G( \mbox{CL}_{
C^{eff}_\eta } ( \underline{\zeta}_{,t<0} \bullet
C(\underline{\zeta}_{,0}) ))$ can only depend on
$\underline{\zeta}_{,t<0}$ and the non-$\eta$ components of
$\underline{\zeta}_{,0}$. Accordingly, the WLU for
$C^{eff}_\eta$ is just $G$ minus a term that is a
function of $\underline{\zeta}_{,t<0}$ and
$\underline{\zeta}_{\;\hat{},\eta,0}$. By choosing $\Phi_\eta(., ., .)$
in Thm. 1 to be that difference, we see that $\eta$'s effect set WLU
is of the form necessary for the system to be factored.
{\bf QED.}

\vspace{.5cm}
\noindent
As a generalization of (ii), the system is factored if each node $\eta$'s personal
utility is (a monotonically increasing function of) the WLU for a set
$\sigma_\eta$ that contains $C^{eff}_\eta$.

For conciseness, except where explicitly needed, for the remainder of
this subsection we will suppress the argument
``$\underline{\zeta}_{,t<0}$'', taking it to be implicit. The next
result concerning the practical importance of effect set WLU is the
following:

\vspace{.5cm}

\noindent
{\bf Theorem 3:} Let $\sigma$ be a set containing $C_{\eta}^{eff}$. Then
\begin{eqnarray}
\frac{\lambda_{\eta,WLU_{\sigma}}(\underline{\zeta})}
	{\lambda_{\eta,G}(\underline{\zeta})}
		=
  \frac{|| \partial_{\underline{\zeta}_{\;\hat{}\eta,0} }
G(C(\underline{\zeta}_{,0}))  ||}
	{|| \partial_{\underline{\zeta}_{\;\hat{}\eta,0}}
G(C(\underline{\zeta}_{,0}))   -  
	\partial_{\underline{\zeta}_{\;\hat{}\eta,0}}
G(\mbox{CL}_{\sigma}(C(\underline{\zeta}_{,0}))) ||} \; . \nonumber
\end{eqnarray}

\vspace{.5cm}
\noindent
{\bf Proof:} Writing it out,
\begin{eqnarray}
\lambda_{\eta,WLU_{\sigma}}(\underline{\zeta}) =
\frac  {||\partial_{\underline{\zeta}_{\eta,0}}
G(C(\underline{\zeta}_{,0}))   -
	\partial_{\underline{\zeta}_{\eta,0}}
G( \mbox{CL}_{\sigma}(C(\underline{\zeta}_{,0}))) ||}
	{|| \partial_{\underline{\zeta}_{\;\hat{}\eta,0}}
G(C(\underline{\zeta}_{,0}))   -
	\partial_{\underline{\zeta}_{\;\hat{}\eta,0}}
G( \mbox{CL}_{\sigma}(C(\underline{\zeta}_{,0})))||} \;. \nonumber
\end{eqnarray}
The second term in the numerator equals 0, by definition of effect
set. Dividing by the similar expression for
$\lambda_{\eta,G}(\underline{\zeta})$ then gives the result claimed.
{\bf QED.}

\vspace{.5cm} \noindent So if we expect that ratio of magnitudes of
gradients to be large, effect set WLU has much higher learnability
than team game utility --- while still being factored, like team game
utility. As an example, consider the case where the COIN is a very
large system, with $\eta$ being only a relatively minor part of the
system ({\em e.g.}, a large human economy with $\eta$ being a
``typical John Doe living in Peoria Illinois''). Often in such a
system, for the vast majority of nodes $\eta' \neq \eta$, how $G$
varies with $\underline{\zeta}_{\eta',}$ will be essentially
independent of the value $\underline{\zeta}_{\eta,0}$. (For example,
how GDP of the US economy varies with the actions of our John Doe from
Peoria, Illinois will be independent of the state of some Jane Smith
living in Los Angeles, California.) In such circumstances, Thm. 3
tells us that the effect set wonderful life utility for $\eta$ will
have a far larger learnability than does the world utility.

For any fixed $\sigma$, if we change the clamping operation ({\em i.e.},
change the choice of the ``arbitrary fixed value'' we clamp each
component to), then we change the mapping $\underline{\zeta}_{,0}
\rightarrow \mbox{CL}_{\sigma}(C(\underline{\zeta}_{,0}))$, and
therefore change the mapping $(\underline{\zeta}_{\eta,0},
\underline{\zeta}_{\;\hat{}\eta, 0}) \rightarrow G(\mbox
{CL}_{\sigma}(C(\underline{\zeta}_{,0})))$. Accordingly, changing the
clamping operation can affect the value of
$\partial_{\underline{\zeta}_{\;\hat{}\eta,0}} G(\mbox
{CL}_{\sigma}(C(\underline{\zeta}_{,0}))$ evaluated at some point
$\underline{\zeta}_{,0}$. Therefore, by Thm. 3, changing the clamping
operation can affect
$\lambda_{\eta,WLU_{\sigma}}(\underline{\zeta})$. So properly
speaking, for any choice of $\sigma$, if we are going to use
$WLU_{\sigma}$, we should set the clamping operation so as to maximize
learnability. For simplicity though, in this paper we will ignore this
phenomenon, and simply set the clamping operation to the more or less
``natural'' choice of {\bf{0}}, as mentioned above.

Next consider the case where, for some node $\eta$, we can write
$G(\underline\zeta_{,})$ as $G_1(\underline{\zeta}_ {C^{eff}_\eta} ) +
G_2(\underline{\zeta}_{,t<0} \bullet
\underline{\zeta}_{\;\hat{}C^{eff}_\eta})$.  Say it is also true that
$\eta$'s effect set is a small fraction of the set of all
components. In this case it often true that the values of $G(.)$ are
much larger than those of $G_1(.)$, which means that partial
derivatives of $G(.)$ are much larger than those of $G_1(.)$. In such
situations the effect set WLU is far more learnable than the world
utility, due to the following results:

\vspace{.5cm}

\noindent
{\bf Theorem 4:} If for some node $\eta$ there is a set $\sigma$
containing $C^{eff}_{\eta}$, a function
$G_1(\underline{\zeta}_{\sigma} \in {\bf{\underline{Z}}}_{\sigma})$,
and a function $G_2(\underline{\zeta}_{\; \hat{}\sigma} \in
{\bf{\underline{Z}}}_{\;\hat{} \sigma})$, such that
$G(\underline{\zeta}) = G_1(\underline{\zeta}_{\sigma}) +
G_2(\underline{\zeta}_{\;\hat{} \sigma})$, then
\begin{eqnarray}
\frac{\lambda_{\eta,WLU_{\sigma}}(\underline{\zeta})}
	{\lambda_{\eta,G}(\underline{\zeta})}
		=
  \frac{|| \partial_{\underline{\zeta}_{\;\hat{}\eta,0}} 
G(C(\underline{\zeta}_{,0}))  ||}
	{|| \partial_{\underline{\zeta}_{\;\hat{}\eta,0}}
G(\mbox{CL}_{\;\hat{}\sigma}(C(\underline{\zeta}_{,0}))) ||}   \; . \nonumber
\end{eqnarray}

\vspace{.5cm}
\noindent
{\bf Proof:} For brevity, write $G_1$ and $G_2$ both as functions of
full $\underline{\zeta} \in {\bf{\underline{Z}}}$, just such functions
that are only allowed to depend on the components of
$\underline{\zeta}$ that lie in $\sigma$ and those components that do
not lie in $\sigma$, respectively. Then the $\sigma$ WLU for node
$\eta$ is just $g_\eta(\underline{\zeta}) = G_1(\underline{\zeta}) -
G_1(\mbox{CL}_{\sigma}(\underline{\zeta}))$.  Since in that second
term we are clamping all the components of $\underline{\zeta}$ that
$G_1(.)$ cares about, for this personal utility
$\partial_{\underline{\zeta}_{,0}} g_\eta(C(\underline{\zeta}_{,0})) =
\partial_{\underline{\zeta}_{,0}} G_1(C(\underline{\zeta}_{,0}))$.  So
in particular $\partial_{\underline{\zeta}_{\;\hat{}\eta,0}}
g_\eta(C(\underline{\zeta}_{,0})) =
\partial_{\underline{\zeta}_{\;\hat{}\eta,0}}
G_1(C(\underline{\zeta}_{,0})) =
\partial_{\underline{\zeta}_{\;\hat{}\eta,0}}
G(\mbox{CL}_{\;\hat{}\sigma}(C(\underline{\zeta}_{,0})))$. Now by
definition of effect set, $\partial_{\underline{\zeta}_{\eta,0}}
G_2(\underline{\zeta}_{,t<0} \bullet C(\underline{\zeta}_{,0})) = \vec{0}$,
since $\hat{}\sigma$ does not contain $C_{\eta}^{eff}$. So
$\partial_{\underline{\zeta}_{\eta,0}} G(C(\underline{\zeta}_{,0})) =
\partial_{\underline{\zeta}_{\eta,0}} G_1(C(\underline{\zeta}_{,0})) =
\partial_{\underline{\zeta}_{\eta,0}}
g_\eta(C(\underline{\zeta}_{,0}))$.   {\bf QED.}

\vspace{5mm}
\noindent
The obvious extensions of Thm.'s 3 and 4 to effect sets with respect 
to times other than 0 can also be proven~\cite{wolp99}.

An important special case of Thm. 4 is the following:

\vspace{5mm}
\noindent
{\bf Corollary 1:} If for some node $\eta$ we can write

i) $G(\underline{\zeta}) = G_1(\underline{\zeta}_{\sigma}) +
G_2([\underline{\zeta}_{\;\hat{}\sigma}]_{t\ge0}) +
G_3(\underline{\zeta}_{,t<0})$

\noindent
for some set $\sigma$ containing $C^{eff}_{\eta}$, and if

ii) $||\partial_{\underline{\zeta}_{\;\hat{}\eta,0}}
G(C(\underline{\zeta}_{,0}))||
\gg
||\partial_{\underline{\zeta}_{\;\hat{}\eta,0}}
G_1([C(\underline{\zeta}_{,0})]_{\sigma}) ||$,

\noindent
then

$\lambda_{\eta,WLU_{\sigma}}(\underline{\zeta})
\gg
\lambda_{\eta,G}(\underline{\zeta})$.

\vspace{5mm}
\noindent
In practice, to assure that condition (i) of this corollary is met
might require that $\sigma$ be a proper superset of
$C^{eff}_\eta$. Countervailingly, to assure that condition (ii) is
met will usually force us to keep $\sigma$ as small as possible.

One can often remove elements from an effect set and still have the
results of this section hold.  Most obviously, if ($\eta'$, t) $\in
C^{eff}_\eta$ but $\partial_{\underline{\zeta}_{\eta',t}}
G(\underline{\zeta})$ = {\bf 0}, we can remove ($\eta'$, t) from
$C^{eff}_\eta$ without invalidating our results. More generally, if
there is a set $\sigma' \in C^{eff}_\eta$ such that for each component
($\eta, 0; i)$ the chain rule term $\sum_{(\eta',t) \in
\sigma'}[\partial_{\underline{\zeta}_{\eta',t}} G(\underline{\zeta})]
\;\cdot \; [\partial_{\underline{\zeta}_{\eta,0;i}}
[C(\underline{\zeta}_{,0})]_{\eta',t}]$ = 0, then the effects on $G$ of
changes to $\underline{\zeta}_{\eta,0}$ that are ``mediated'' by the
members of $\sigma'$ cancel each other out. In this case we can
usually remove the elements of $\sigma'$ from $C^{eff}_\eta$ with no
ill effects.

\subsubsection{Inducing our salient characteristic} 

Usually the mathematics of a descriptive framework --- a formal
investigation of the salient characteristics --- will not provide
theorems of the sort, ``If you modify the COIN the following way at
time $t$, the  value of the world utility will increase.''  Rather
it provides theorems that relate a COIN's salient characteristics with
the general properties of the COIN's entire history, and in particular
with those properties embodied in $C$.  In particular, the salient
characteristic that we are concerned with in this chapter is that the
system be highly intelligent for personal utilities for which it is
factored, and our mathematics concerns the relationship between
factoredness, intelligence, personal utilities, effect sets, and the
like.

More formally, the desideratum associated with our salient
characteristic is that we want the COIN to be at a $\underline{\zeta}$
for which there is some set of \{$g_\eta$\} (not necessarily
consisting of private utilities) such that (a) $\underline{\zeta}$ is
factored for the \{$g_\eta$\}, and (b)
$\epsilon_{\eta,g_\eta}(\underline{\zeta})$ is large for all
$\eta$. Now there are several ways one might try to induce the COIN to
be at such a point. One approach is to have each algorithm controlling
$\eta$ explicitly try to ``steer'' the worldline towards such a point.
In this approach $\eta$ needn't even have a private utility in the
usual sense. (The overt ``goal'' of the algorithm controlling $\eta$
involves finding a $\underline{\zeta}$ with a good associated extremum
over the class of all possible $g_\eta$, independent of any private
utilities.)  Now initialization of the COIN, {\em i.e.}, fixing of
$\underline{\zeta}_{,0}$, involves setting the algorithm controlling
$\eta$, in this case to the steering algorithm. Accordingly, in this
approach to initialization, we fix $\underline{\zeta}_{,0}$ to a point
for which there is some special $g_\eta$ such that both
$C(\underline{\zeta}_{,0})$ is factored for $g_\eta$, and
$\epsilon_{\eta,g_\eta}(C(\underline{\zeta}_{,0}))$ is large. There is
nothing peculiar about this. What is odd though is that in this
approach we do not know what that ``special'' $g_\eta$ is when we do
that initialization; it's to be determined, by the unfolding of the
system.

In this chapter we concentrate on a different approach, which can
involve either initialization or macrolearning. In this alternative we
deploy the \{$g_\eta$\} as the microlearners' private utilities at
some $t < 0$, in a process not captured in $C$, so as to induce a
factored COIN that is as intelligent as possible. (It is with that
``deploying of the \{$g_\eta$\}'' that we are trying to induce our
salient characteristic in the COIN.)  Since in this approach we are
using private utilities, we can replace intelligence with its
surrogate, learnability. So our task is to choose \{$g_\eta$\}
which are as learnable as possible while still being factored.

Solving for such utilities can be expressed as solving a set of
coupled partial differential equations. Those equations involve the
tangent plane to the manifold $C$, a functional trading off (the
differential versions of) degree of factoredness and learnability, and any
communication constraints on the nodes we must respect. While there is
not space in the current chapter to present those equations, we can
note that they are highly dependent on the correlations among the
components of $\underline{\zeta}_{\eta,t}$.  So in this approach, in
COIN initialization we use some preliminary guesses as to those
correlations to set the initial \{$g_\eta$\}. For example, the effect
set of a node constitutes all components $\underline{\zeta}_{\eta',
t>0}$ that have non-zero correlation with
$\underline{\zeta}_{\eta,0}$.  Furthermore, by Thm. 2 the system is
factored for effect set WLU personal utilities. And by Coroll. 1, for
small effect sets, the effect set WLU has much greater differential
utility learnability than does $G$.  Extending the reasoning behind
this result to all $\underline{\zeta}$ (or at least all likely
$\underline{\zeta}$), we see that for this scenario, the descriptive
framework advises us to use Wonderful Life private utilities based on
(guesses for) the associated effect sets rather than the team game
private utilities, $g_\eta = G \; \; \forall \eta$.

In macrolearning we must instead run-time estimate an approximate
solution to our partial differential equations, based on statistical
inference.\footnote{Recall that in the physical world, it is often
useful to employ devices using algorithms that are based on
probabilistic concepts, even though the underlying system is
ultimately deterministic. (Indeed, theological Bayesians invoke a
``degree of belief'' interpretation of probability to {\it demand}
such an approach --- see ~\cite{wolp96c} for a discussion of the
legitimacy of this viewpoint.) Similarly, although we take the
underlying system in a COIN to be deterministic, it is often useful to
use microlearners or --- as here --- macrolearners that are based on
probabilistic concepts.} As an example, we might start with an initial
guess as to $\eta$'s effect set, and set its private utility to the
associated WLU. But then as we watch the system run and observe the
correlations among the components of $\underline{\zeta}$, we might
modify which components we think comprise $\eta$'s effect set, and
modify $\eta$'s personal utility accordingly.

\subsection{Illustrative Simulations of our Descriptive Framework}
As implied above, often one can perform reasonable COIN initialization
and/or macrolearning without writing down the partial differential equations
governing our salient characteristic explicitly. Simply ``hacking''
one's way to the goal of maximizing both degree of factoredness and
intelligibility, for example by estimating effect sets,
often results in dramatic improvement in performance. This is
illustrated in the experiments recounted in the next two subsections.

\subsubsection{COIN Initialization}
Even if we don't exactly
know the effect set of each node $\eta$, often we will be able to make
a reasonable guess about which components of $\underline{\zeta}$
comprise the ``preponderance'' of $\eta$'s effect set. We call such a
set a {\bf guessed effect set}. As an example,
often the primary effects of changes to $\eta$'s state will be on the
future state of $\eta$, with only relatively minor effects on the
future states of other nodes. In such situations, we would expect to
still get good results if we approximated the effect set WLU of each
node $\eta$ with a
WLU based on the guessed effect set $\underline{\zeta}_{\eta,t \ge 0}$.  In
other words, we
would expect to be able to replace WLU$_{C^{eff}_{\eta}}$ with
WLU$_{\underline{\zeta}_{\eta,t \ge 0}}$ and still get good performance.

This phenomenon was borne out in the experiments recounted in
\cite{wotu99a} that used COIN initialization for distributed control
of network packet routing.  In a conventional approach to packet
routing, each router runs what it believes (based on the information
available to it) to be a shortest path algorithm (SPA), {\em i.e.}, each
router sends its packets in the way that it surmises will get those
packets to their destinations most quickly.  Unlike with an approach
based on our COIN framework, with
SPA-based routing the routers have no concern for the possible
deleterious side-effects of their routing decisions on the global
performance ({\em e.g.}, they have no concern for whether they induce
bottlenecks).  We performed simulations in which we compared such a COIN-based
routing system to an SPA-based system. For the COIN-based system $G$
was global throughput and no macrolearning was used. The COIN
initialization was to have each router's private utility be a WLU
based on an associated guessed effect set generated {\it a priori}.  In
addition, the COIN-based system was realistic in that each router's
reinforcement algorithm had imperfect knowledge of the state of the
system. On the other hand, the SPA was an idealized ``best-possible''
system, in which each router knew exactly what the shortest paths were
at any given time. Despite the handicap that this disparity imposed on
the COIN-based system, it achieved significantly better global throughput in our
experiments than did the perfect-knowledge SPA-based system, and in
particular, avoided the Braess' Paradox that was built-in to some of those
systems~\cite{tuwo99}.

The experiments in \cite{wotu99a} were primarily concerned with the
application of packet-routing. To concentrate more precisely on the
issue of COIN initialization, we ran subsequent experiments on variants
of Arthur's famous ``El Farol bar problem'' (see
Section~\ref{sec:lit}). To facilitate the analysis we modified Arthur's
original problem to be more general, and since we were not interested
in directly comparing our results to those in the literature, we used
a more conventional (and arguably ``dumber'') machine learning
algorithm than the ones investigated
in~\cite{arth94,cama97,chzh97,capl98}.

In this formulation of the bar problem~\cite{wowh99a}, there are $N$
agents, each of whom picks one of seven nights to attend a bar the
following week, a process that is then repeated.  In each week, each
agent's pick is determined by its predictions of the associated
rewards it would receive.  These predictions in turn are based solely
upon the rewards received by the agent in preceding weeks. An agent's
``pick'' at week $t$ ({\em i.e.}, its node's state at that week) is
represented as a unary seven-dimensional vector. (See the discussion
in the definitions subsection of our representing discrete variables
as Euclidean variables.) So $\eta$'s zeroing its state in some week,
as in the CL$_{\underline{\zeta}_{\eta,t}}$ operation, essentially
means it elects not to attend any night that week.

The world utility is $$G(\underline{\zeta})~=~ \sum_t
R(\underline\zeta_{,t}),$$ where: $R(\underline{\zeta}_{,t}) =
\sum_{k=1}^7 \gamma_k(x_k(\underline{\zeta}_{,t}))$;
$x_k(\underline{\zeta}_{, t})$ is the total attendance on night $k$ at
week $t$; $\gamma_k(y) ~\equiv~ \alpha_k y \exp{(-y/c)}$; and $c$ and
each of the \{$\alpha_k$\} are real-valued parameters.  Intuitively,
the ``world reward'' $R$ is the sum of the global ``rewards'' for each
night in each week.  It reflects the effects in the bar as the
attendance profile of agents changes.  When there are too few agents
attending some night, the bar suffers from lack of activity and
therefore the global reward for that night is low.  Conversely, when
there are too many agents the bar is overcrowded and the reward for
that night is again low.  Note that $\gamma_k(\cdot)$ reaches its
maximum when its argument equals $c$.

In these experiments we investigate two different $\vec{\alpha}$'s.
One treats  all
nights equally; $\vec{\alpha}~=~[1~1~1~1~1~1~1]$.  The other is only
concerned with one night; $\vec{\alpha}~=~[0~0~0~7~0~0~0]$.  In our
experiments, $c = 6$ and $N$ is chosen to be 4 times
larger than the number of agents necessary to have $c$ agents attend
the bar on each of the seven nights, {\em i.e.}, there are $4 \cdot 6
\cdot 7 = 168$ agents (this ensures that there are no trivial 
solutions and that for the world utility to be maximized, the 
agents have to ``cooperate'').

As explained below, our microlearning algorithms worked by providing
a real-valued ``reward'' signal to each agent at each week $t$. Each
agent's reward function is a surrogate for an associated utility
function for that agent. The difference between the two functions is
that the reward function only reflects the state of the system at one
moment in time (and therefore is potentially observable), whereas the utility
function reflects the agent's ultimate goal, and therefore can depend
on the full history of that agent across time.

We investigated three agent reward functions. One was based on effect
set WLU. The other two were ``natural'' rewards included for
comparison purposes. With $d_{\eta}$ the night selected by $\eta$, the
three rewards are:
\begin{eqnarray*}
   \mbox{\rm Uniform Division (UD):} & 
   r_{\eta}(\underline{\zeta}_{,t})  & \equiv
   \gamma_{d_\eta} (x_{d_\eta} (\underline{\zeta}_{,t}))/x_{d_\eta}(\underline{\zeta}_{,t}) \\ [-.1in]
   \mbox{Global (G):} \; \; \;  \; \; \;  \; \; \;  \; \; \;  \; \; \;  \; \; &  
   r_{\eta}(\underline{\zeta}_{,t})&  \equiv 
	R(\underline{\zeta}_{,t}) =
\sum_{k=1}^7 \gamma_k(x_k(\underline{\zeta}_{,t}))  \\ [-.05in]
   \mbox{Wonderful Life (WL):} \; \; \;  & 
   r_{\eta}(\underline{\zeta}_{,t}) & \equiv 
	R(\underline{\zeta}_{,t}) -
	R(\mbox{CL}_{\underline{\zeta}_{\eta,t}}(\underline{\zeta}_{,t})) \\ [-.05in]
 &&= \gamma_{d_\eta} (x_{d_\eta} (\underline{\zeta}_{,t})) -
 \gamma_{d_\eta} (x_{d_\eta}
   (\mbox{CL}_{\underline{\zeta}_{\eta,t}} (\underline{\zeta}_{,t})))
\end{eqnarray*}
\renewcommand{\arraystretch}{1.0}

The conventional UD reward is a natural ``naive'' choice for the
agents' reward; the total reward on each night gets uniformly divided
among the agents attending that night.  If we take
$g_\eta(\underline{\zeta}) = \sum_t r_\eta (\underline{\zeta}_{,t})$
({\em i.e.}, $\eta$'s utility is an undiscounted sum of its rewards), then
for the UD reward $G(\underline{\zeta}) =
\sum_{\eta}\;g_{\eta}(\underline{\zeta})$, so that the system is
weakly trivial. The original version of the bar problem in the physics
literature~\cite{chzh97} is the special case where UD reward is used
but there are only two ``nights'' in the week (one of which
corresponds to ``staying at home''); $\vec{\alpha}$ is uniform; and
$\gamma_k(x_k) = x_k \Theta(c_kN - x_k)$ for some vector $\vec{c}$, taken to
equal (.6, .4) in the very original papers. So the reward to agent $\eta$ is
1 if it attends the bar and attendance is below capacity, or if it stays
at home and the bar is over capacity. Reward is 0 otherwise. (In
addition, unlike in our COIN-based systems, in the original work on
the bar problem the microlearners work by explicitly predicting
the bar attendance, rather than by directly
modifying behavior to try to increase a reward signal.)

In contrast to the UD reward, providing the G reward at time $t$ to
each agent results in all agents receiving the same reward.  This is
the team game reward function, investigated for example in
~\cite{crba96}. For this reward function, the system is automatically
factored if we define $g_\eta(\underline{\zeta}) \equiv \sum_t
r_\eta(\underline{\zeta}_{,t})$. However, evaluation of this reward
function requires centralized communication concerning all seven
nights.  Furthermore, given that there are 168 agents, G is likely to
have poor learnability as a reward for any individual agent.

This latter problem is obviated by using the WL reward, where the
subtraction of the clamped term removes some of the ``noise'' of the
activity of all other agents, leaving only the underlying ``signal''
of how the agent in question affects the utility.  So one would expect
that with the WL reward the agents can readily discern the effects of
their actions on their rewards. Even though the conditions in Coroll. 1
don't hold\footnote{The $t=0$ elements of $C_\eta^{eff}$ are just
$\underline{\zeta}_{\eta,t=0}$, but the contributions of
$\underline{\zeta}_{,t=0}$ to $G$ cannot be written as a sum of a
$\underline{\zeta}_{\eta,t=0}$ contribution and a
$\underline{\zeta}_{\;\hat{} \eta,t=0}$ contribution.}, this reasoning
accords with the implicit advice of Coroll. 1 under the approximation of
the $t = 0$ effect set as $C^{eff}_\eta \approx
\underline{\zeta}_{\eta,t\ge0}$. In other words, it agrees with that corollary's
implicit advice under the identification of $\underline{\zeta}_{\eta,t
\ge 0}$ as $\eta$'s $t=0$ guessed effect set.

In fact, in this very simple system, we can explicitly calculate the
ratio of the WL reward's learnability to that of the G reward, by
recasting the system as existing for only a single instant so that
$C_{\eta}^{eff} = \underline{\zeta}_{\eta,0}$ exactly and then
applying Thm. 3. So for example, say that all $\alpha_k = 1$, and that
the number of nodes $N$ is evenly divided among the seven nights. The
numerator term in Thm. 3 is a vector whose components are some of the
partials of G evaluated when $x_k(\underline{\zeta}_{,0}) = N/7$. This
vector is $7(N-1)$ dimensional, one dimension for each of the 7
components of (the unary vector comprising) each node in
$\hat{}\eta$. For any particular $\eta' \neq \eta$ and night $i$, the
associated partial derivative is $\sum_k
[e^{-x_k(\underline{\zeta}_{,0})/c} (1 -
x_k(\underline{\zeta}_{,0})/c) \cdot
\partial_{\underline{\zeta}_{\eta',0;i}}(x_k(\underline{\zeta}_{,0}))]$,
where as usual ``$\underline{\zeta}_{\eta',0;i}$'' indicates the $i$'th
component of the unary vector $\underline{\zeta}_{\eta',0}$. Since
$\partial_{\underline{\zeta}_{\eta',0;i}}(x_k(\underline{\zeta}_{,0}))
= \delta_{i,k}$, for any fixed $i$ and $\eta'$, this sum just equals
$e^{(-N/7c)} \; (1 - N/7c)$. Since there are $7(N-1)$ such terms,
after taking the norm we obtain $|e^{(-N/7c)} \; [1 - N/7c] \;
\sqrt{7(N-1})|$.

The denominator term in Thm. 3 is the difference between the gradients
of the global reward and the clamped reward. These differ on only $N -
1$ terms, one term for that component of each node $\eta' \neq \eta$
corresponding to the night $\eta$ attends. (The other $6N-6$ terms are
identical in the two partials and therefore cancel.) This yields
$|e^{(-N/7c)} \; [1 - N/7c] \; [1 - e^{1/c} (1 - \frac{7}{N-7c})] \;
\sqrt{N-1}$. Combining with the result of the previous paragraph, our
ratio is $|\sqrt{7} \; \frac{N-7c}{(N-7c)(1 - e^{1/c}) + 7e^{1/c}}|
\simeq 11$.

In addition to this learnability advantage of the WL reward, to
evaluate its WL reward each agent only needs to know the total
attendance on the night it attended, so no centralized communication
is required. Finally, although the system won't be perfectly factored
for this reward (since in fact the effect set of $\eta$'s action at
$t$ would be expected to extend a bit beyond
$\underline{\zeta}_{\eta,t}$), one might expect that it is close
enough to being factored to result in large world utility.

Each agent keeps a seven dimensional Euclidean vector representing its
estimate of the reward for attending each night of the week.  At the
end of each week, the component of this vector corresponding to the
night just attended is proportionally adjusted towards the actual
reward just received.  At the beginning of the succeeding week, the
agent picks the night to attend using a Boltzmann distribution with
energies given by the components of the vector of estimated rewards,
where the temperature in the Boltzmann distribution decays in time.
(This learning algorithm is equivalent to Claus and
Boutilier's~\cite{clbo98} independent learner algorithm for
multi-agent reinforcement learning.) We used the same parameters
(learning rate, Boltzmann temperature, decay rates, etc.) for all
three reward functions. (This is an {\it extremely} primitive RL
algorithm which we only chose for its pedagogical value; more
sophisticated RL algorithms are crucial for eliciting high
intelligence levels when one is confronted with more complicated
learning problems.)

\begin{figure}[ht]
	\vspace*{-.1in}
   \centerline{\mbox{
   	\psfig{figure={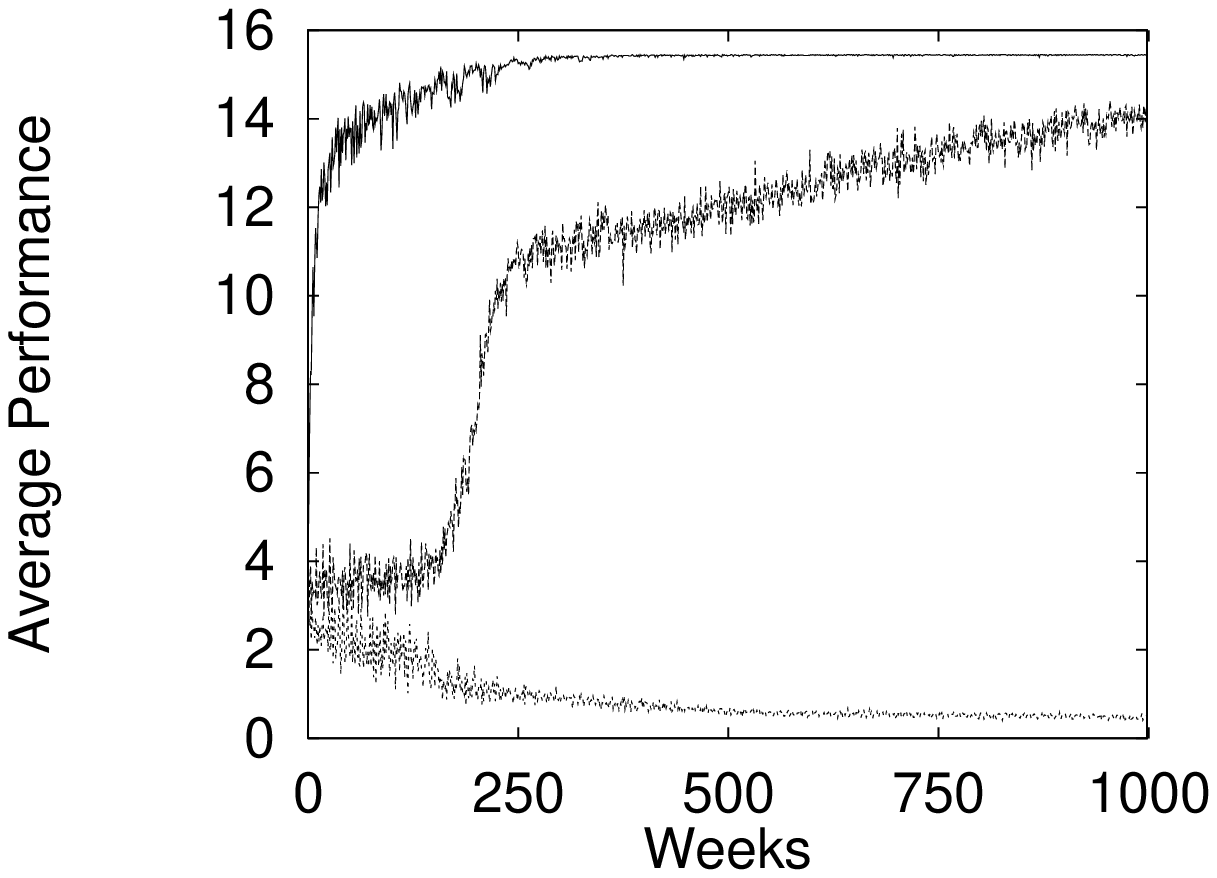},width=2.80in,height=2.0in}
	\hspace{0.10in}
   	\psfig{figure={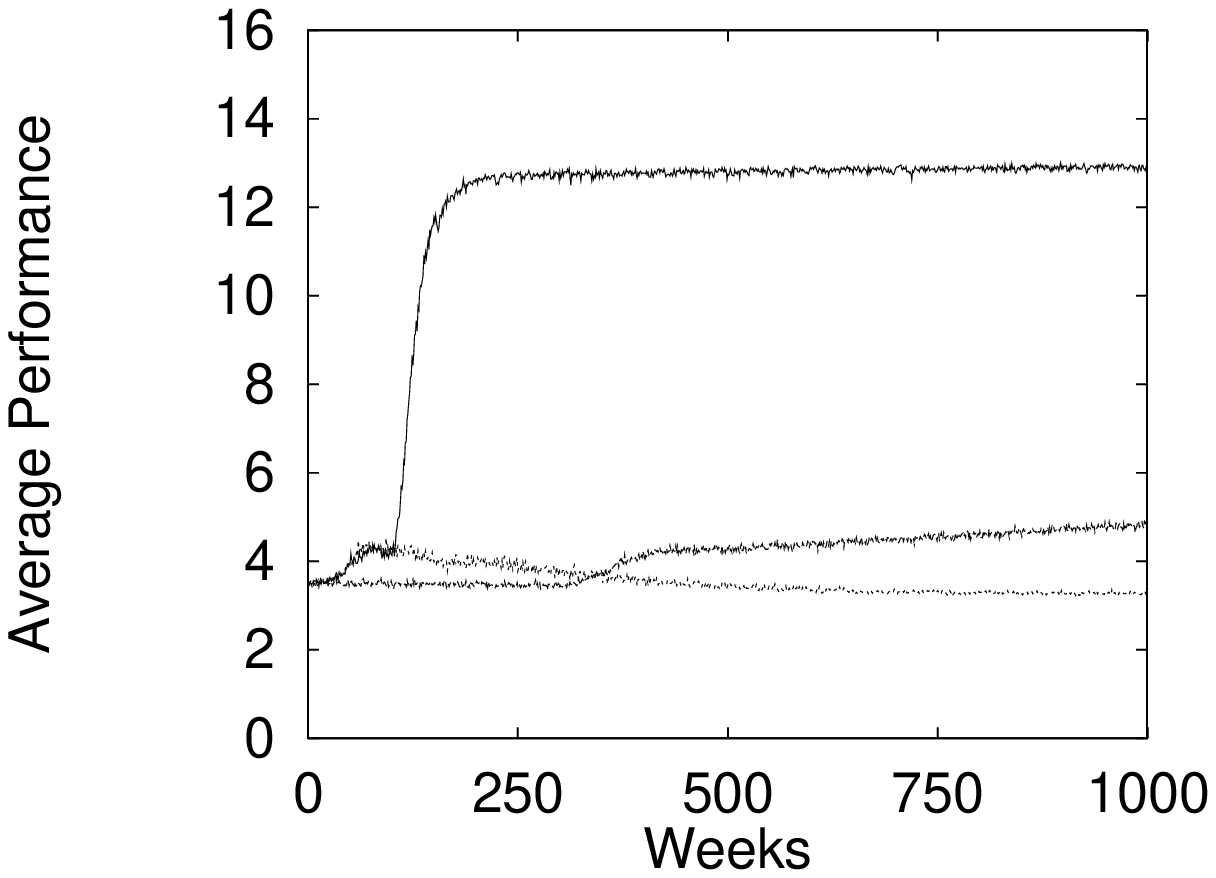},width=2.80in,height=2.0in} 
	}}
	\vspace*{-.1in}
  \caption{Average world reward when $\vec{\alpha}~=~[0~0~0~7~0~0~0]$ (left) and
	when $\vec{\alpha}~=~[1~1~1~1~1~1~1]$ (right). In both plots the top curve is
	WL, middle is G, and bottom is UD.}
\label{fig:barfig}
\end{figure}

Figure~\ref{fig:barfig} presents world reward values as a function of
time, averaged over 50 separate runs, for all three reward functions,
for both $\vec{\alpha}~=~[1~1~1~1~1~1~1]$ and
$\vec{\alpha}~=~[0~0~0~7~0~0~0]$.  The behavior with the G reward
eventually converges to the global optimum.  This is in agreement with
the results obtained by Crites~\cite{crba96} for the bank of elevators
control problem.  Systems using the WL reward also converged to
optimal performance.  This indicates that for the bar problem our
approximations of effects sets are sufficiently accurate, {\em i.e.}, that
ignoring the effects one agent's actions will have on future actions
of other agents does not significantly diminish performance.  This
reflects the fact that the only interactions between agents occurs
indirectly, via their affecting each others' reward values.

However since the WL reward is more learnable than
than the G reward,
convergence with the WL reward should be far quicker than with 
the G reward. Indeed, when $\vec{\alpha}~=~[0~0~0~7~0~0~0]$,
systems using the G reward converge in 1250 weeks, which is
5 times worse than the systems using WL reward.  When
$\vec{\alpha}~=~[1~1~1~1~1~1~1]$  systems take 6500
weeks to converge with the G reward, which is more than {\it 30 times}
worse than the time with
the WL reward.  

\begin{figure}[ht]
	\vspace*{-.1in}
   \centerline{\mbox{
   	\psfig{figure={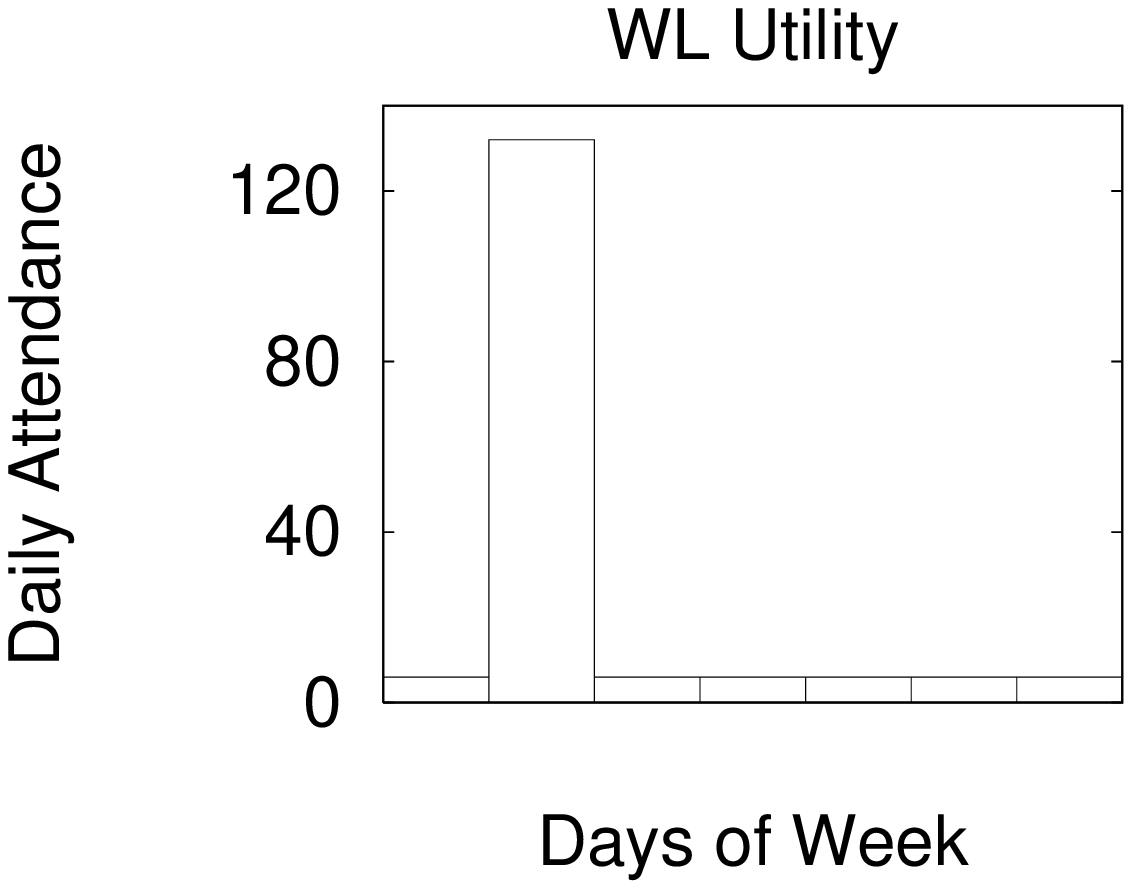},width=1.9in,height=1.5in}
	\psfig{figure={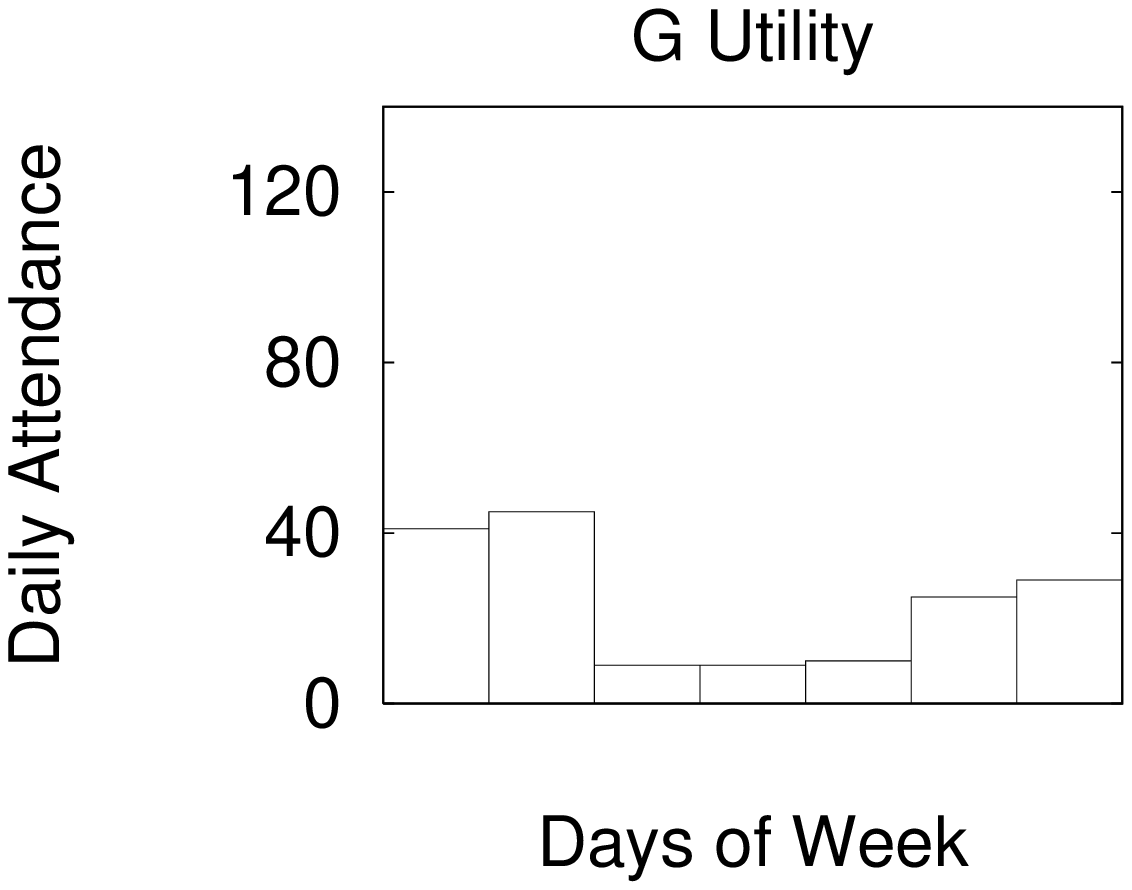},width=1.9in,height=1.5in}
	\psfig{figure={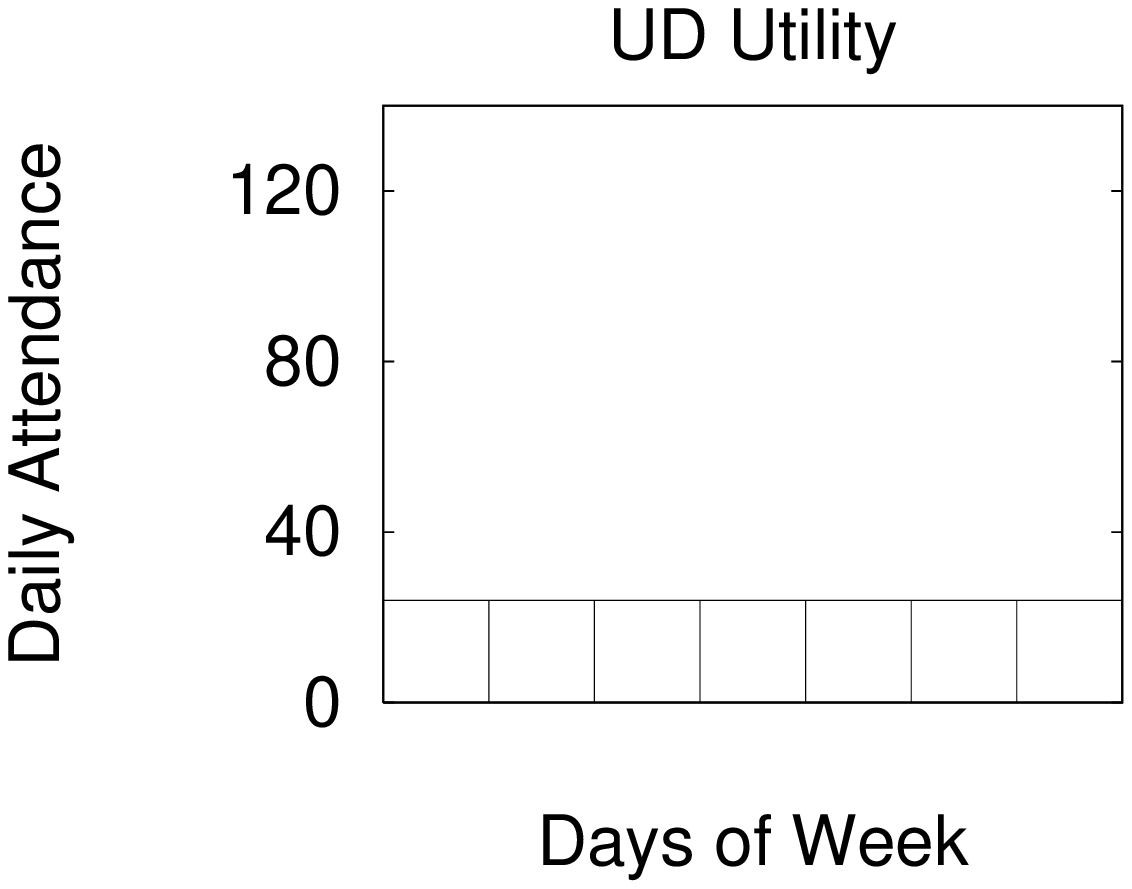},width=1.9in,height=1.5in}
   }}
	\vspace*{-.1in}
   \caption{Typical daily attendance when $\vec{\alpha}~=~[1~1~1~1~1~1~1]$ for 
   WL (left), G (center), and UD (right).} 
\label{fig:attend}
\end{figure}

In contrast to the behavior for reward functions based on our COIN
framework, use of the conventional UD reward results in very poor
world reward values, values that deteriorated as the learning
progressed. This is an instance of the TOC. For example, for the case
where $\vec{\alpha}~=~[0~0~0~7~0~0~0]$, it is in every agent's
interest to attend the same night --- but their doing so shrinks the
world reward ``pie'' that must be divided among all agents.  A similar
TOC occurs when $\vec{\alpha}$ is uniform.  This is illustrated in
fig.~\ref{fig:attend} which shows a typical example of daily
attendance figures (\{$x_k(\underline{\zeta}_{,t})$\}) for each of the
three reward functions for $t~=~2000$.  In this example optimal
performance (achieved with the WL reward) has 6 agents each on 6
separate nights, (thus maximizing the reward on 6 nights), and the
remaining 132 agents on one night.

\begin{figure}[ht]
	\vspace*{-.1in}
   \centerline{\mbox{
   	\psfig{figure={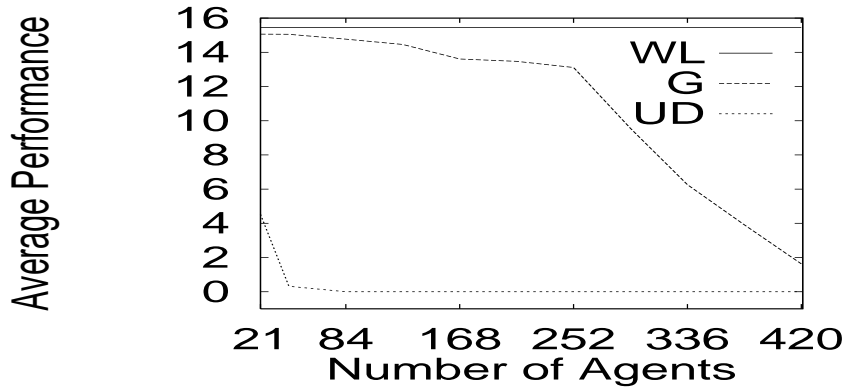},width=4.8in,height=2.0in}
   }}
	\vspace*{-.1in}
   \caption{Behavior of each reward function with respect to the number of
   agents for  $\vec{\alpha}~=~[0~0~0~7~0~0~0]$.} 
\label{fig:numagents}
\end{figure}

Figure~\ref{fig:numagents} shows how $t = 2000$ performance
scales with $N$ for each of the reward signals
for $\vec{\alpha}~=~[0~0~0~7~0~0~0]$.  Systems using the UD
reward perform poorly regardless of $N$. Systems
using the G reward perform well when $N$
is low. As $N$ increases however, it becomes
increasingly difficult for the agents to extract the 
information they need from the G reward. (This problem is significantly worse
for uniform $\vec{\alpha}$.)
Because of their superior learnability,
systems using the WL reward overcome this
signal-to-noise problem ({\em i.e.}, because the WL reward is based on the
{\em difference} between the actual state and the state where 
one agent is clamped, it is much less affected by the
total number of agents).

\subsection{Macrolearning} 
\label{sec:leader} 
In the experiments
recounted above, the agents were sufficiently independent that
assuming they did not affect each other's actions (when forming
guesses for effect sets) allowed the resultant WL reward signals to
result in optimal performance.  In this section we investigate the
contrasting situation where we have initial guesses of effect sets
that are quite poor and that therefore result in bad global
performance when used with WL rewards. In particular, we investigate
the use of macrolearning to correct those guessed effect sets at
run-time, so that with the corrected guessed effect sets WL rewards
will instead give optimal performance. This models real-world
scenarios where the system designer's initial guessed effect sets are
poor approximations of the actual associated effect sets and need to
be corrected adaptively.

In these experiments the bar problem is significantly modified to
incorporate constraints designed to result in poor $G$ when the WL
reward is used with certain initial guessed effect sets.  To do this
we forced the nights actually attended by some of the agents
(followers) to agree with those attended by other agents (leaders),
regardless of what night those followers ``picked'' via their
microlearning algorithms.  (For leaders, picked and actually attended
nights were always the same.) We then had the world utility be the
sum, over all leaders, of the values of a triply-indexed reward matrix
whose indices are the  nights that each leader-follower set
attends: $G(\underline{\zeta}) ~=~ \sum_t \sum_i
R_{l_i(t),f1_{i}(t),f2_{i}(t)}$ where $l_i(t)$ is the night the
$i^{th}$ leader attends in week $t$, and $f1_{i}(t)$ and $f2_i(t)$ are
the nights attended by the followers of leader $i$, in week $t$ (in
this study, each leader has two followers). We also had the states of
each node be one of the integers \{0, 1, ..., 6\} rather than (as in
the bar problem) a unary seven-dimensional vector. This was a bit of
a contrivance, since constructions like
$\partial_{\underline{\zeta}_{\eta,0}}$ aren't meaningful for such
essentially symbolic interpretations of the possible states
$\underline{\zeta}_{\eta,0}$. As elaborated below, though, it was helpful
for constructing a scenario in which guessed effect set WLU results in
poor performance, {\em i.e.}, a scenario in which we can explore the
application of macrolearning.

To see how this setup can result in poor world utility, first note
that the system's dynamics is what restricts all the members of each
triple $(l_i(t), f1_i(t), f2_i(t))$ to equal the night picked by
leader $i$ for week $t$.  So $f1_i(t)$ and $f2_i(t)$ are both in
leader $i$'s actual effect set at week $t$ --- whereas the initial
guess for $i$'s effect set may or may not contain nodes other than
$l_i(t)$. (For example, in the bar problem experiments, the guessed
	       effect set does not contain
any nodes beyond $l_i(t)$.)  On the other hand, $G$ and $R$ are
defined for all possible triples ($l_i(t), f1_i(t), f2_i(t)$).  So in
particular, $R$ is defined for the dynamically unrealizable triples
that can arise in the clamping operation.  This fact, combined with
the leader-follower dynamics, means that for certain $R$'s there exist
guessed effect sets such that the dynamics assures poor world utility
when the associated WL rewards are used.  This is precisely the type
of problem that macrolearning is designed to correct.

As an example, say each week only contains two nights, 0 and 1. Set
$R_{111} = 1$ and $R_{000} = 0$. So the contribution to $G$ when a
leader picks night 1 is 1, and when that leader picks night 0 it is 0,
independent of the picks of that leader's followers (since the actual
nights they attend are determined by their leader's
picks). Accordingly, we want to have a private utility for each leader
that will induce that leader to pick night 1. Now if a leader's
guessed effect set includes both of its followers (in addition to the
leader itself), then clamping all elements in its effect set to 0
results in an $R$ value of $R_{000} = 0$. Therefore the associated
guessed effect set WLU will reward the leader for choosing night 1,
which is what we want. (For this case WL reward equals $R_{111} -
R_{000} = 1$ if the leader picks night 1, compared to reward $R_{000}
- R_{000} = 0$ for picking night 0.)

However consider having two leaders, $i_1$ and $i_2$, where $i_1$'s
guessed effect set consists of $i_1$ itself together with the two
followers of $i_2$ (rather than together with the two followers of
$i_1$ itself). So neither of leader $i_1$'s followers are in its
guessed effect set, while $i_1$ itself is. Accordingly, the three
indices to $i_1$'s $R$ need not have the same value. Similarly,
clamping the nodes in its guessed effect set won't affect the values
of the second and third indices to $i_1$'s $R$, since the values of
those indices are set by $i_1$'s followers. So for example, if $i_2$
and its two followers go to night 0 in week 0, and $i_1$ and its two
followers go to night 1 in that week, then the associated guessed effect 
set wonderful life reward for $i_1$ for week 0 is $G(\underline{\zeta}_{,t=0}) -
G(\mbox{CL}_{l_{i_1}(0),f1_{i_2}(0),f2_{i_2}(0)}(\underline{\zeta}_{,t=0}))
= R_{l_{i_1}(0),f1_{i_1}(0),f2_{i_1}(0)} +
R_{l_{i_2}(0),f1_{i_2}(0),f2_{i_2}(0)} -
[R_{0,f1_{i_1}(0),f2_{i_1}(0)} + R_{l_{i_2}(0),0,0}]$. This equals
$R_{111} + R_{000} - R_{011} - R_{000} = 1 - R_{011}$. Simply by
setting $R_{011} > 1$ we can ensure that this is
negative. Conversely, if leader $i_1$ had gone to night 0, its guessed
effect WLU would have been 0. So in this situation leader $i_1$ will
get a greater reward for going to night 0 than for going to night 1.
In this situation, leader $i_1$'s using its guessed effect set WLU
will lead it to make the wrong pick.

To investigate the efficacy of the macrolearning, two sets of separate
experiments were conducted.  In the first one the reward matrix $R$
was chosen so that if each leader is maximizing its WL reward, but for
guessed effect sets that contain none of its followers, then the
system evolves to $minimal$ world reward.  So if a leader incorrectly
guesses that some $\sigma$ is its effect set even though $\sigma$
doesn't contain both of that leader's followers, and if this is true
for all leaders, then we are assured of worst possible performance.
In the second set of experiments, we investigated the efficacy of
macrolearning for a broader spectrum of reward matrices by generating
those matrices randomly. We call these two kinds of reward matrices
{\em worst-case} and {\em random} reward matrices, respectively.  In
both cases, if it can modify the initial guessed effect sets of the
leaders to include their followers, then macrolearning will induce the
system to be factored. 

The microlearning in these experiments was the same as in the bar
problem. All experiments used the WL personal reward with some
(initially random) guessed effect set.  When macrolearning was used, it
was implemented starting after the microlearning had run for a
specified number of weeks. The macrolearner worked by estimating the
correlations between the agents' selections of which nights to
attend. It did this by examining the attendances of the agents over
the preceding weeks.  Given those estimates, for each agent $\eta$ the
two agents whose attendances were estimated to be the most correlated
with those of agent $\eta$ were put into agent $\eta$'s guessed effect
set.  Of course, none of this macrolearning had any effect on global
performance when applied to follower agents, but the macrolearning
algorithm cannot know that ahead of time; it applied this procedure to
each and every agent in the system.

\begin{figure}[ht]
   \centerline{\mbox{
      \psfig{figure={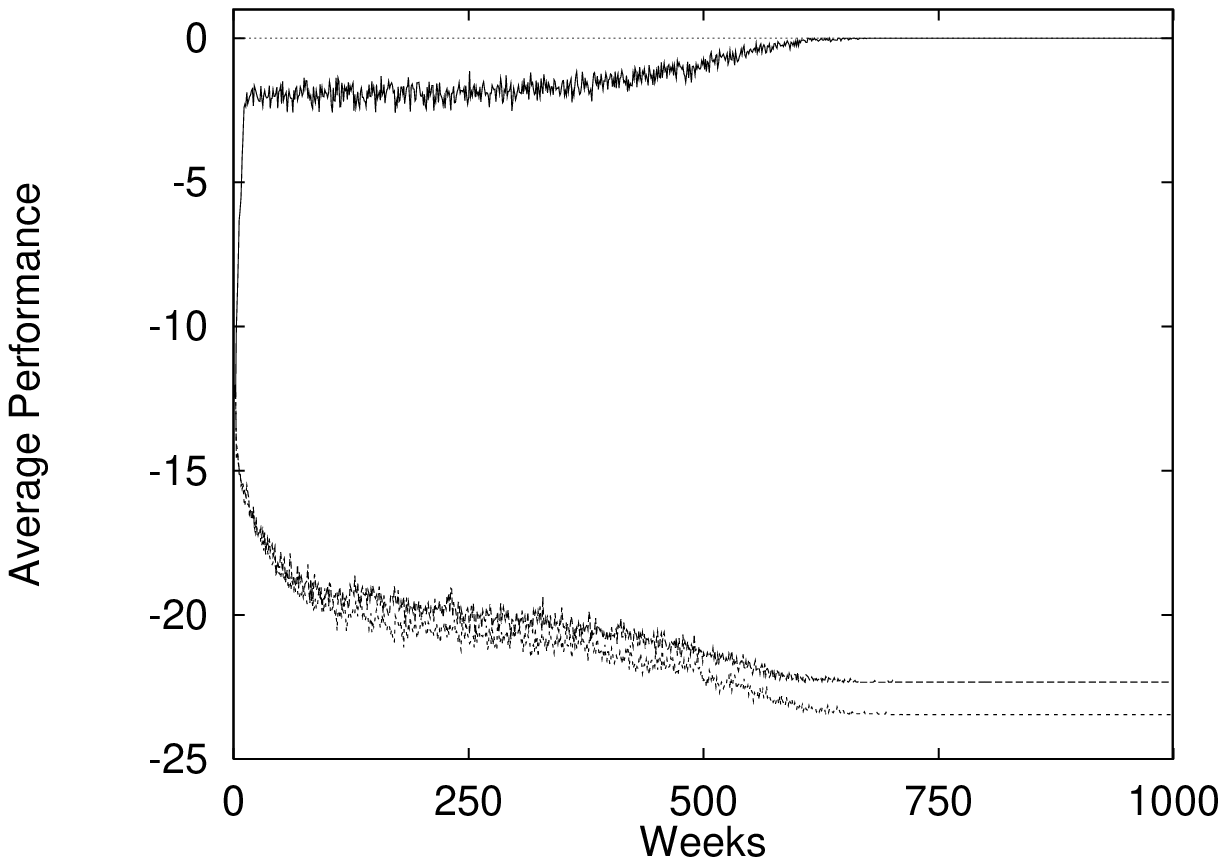},width=3.0in,height=2.25in}
      \hspace{0.1in}
      \psfig{figure={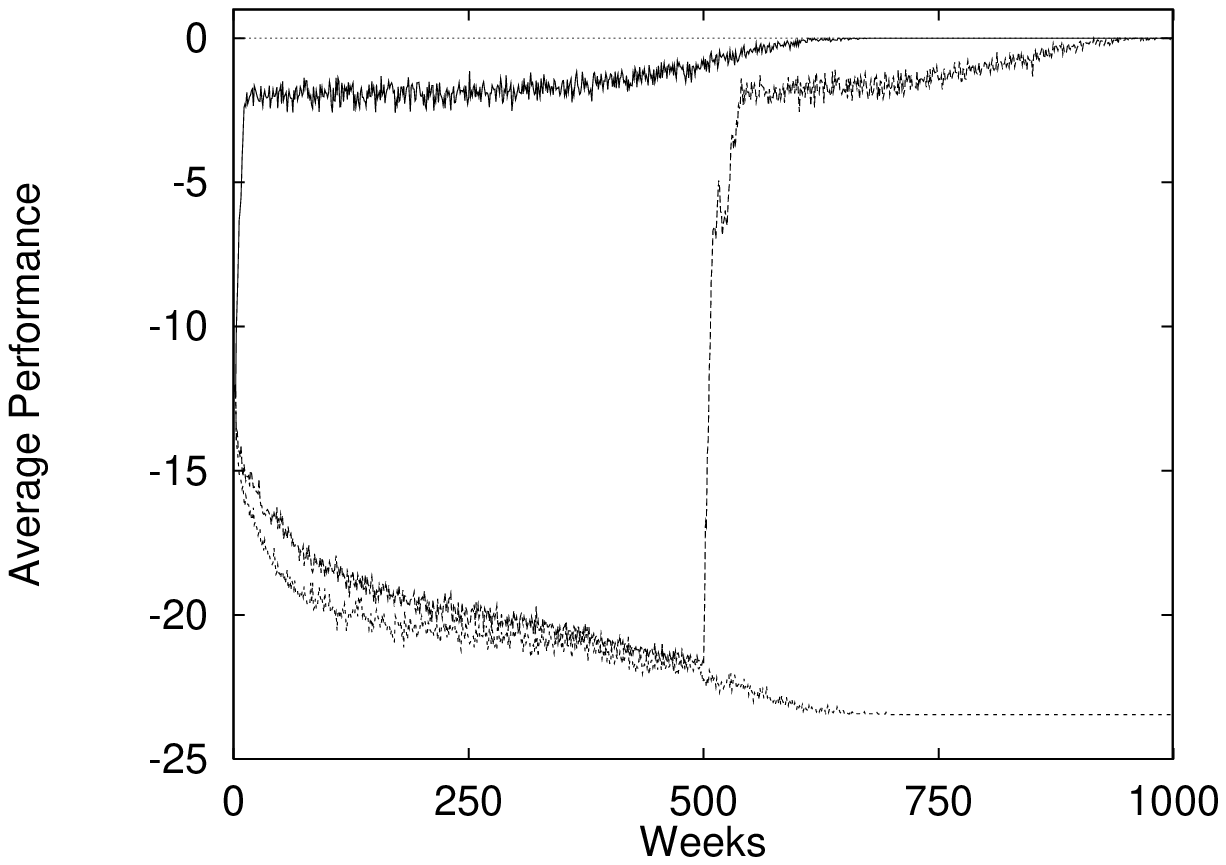},width=3.0in,height=2.25in}
   }}
   \caption{Leader-follower problem with worst case reward matrix. 
   In both plots, every follower is in its leader's guessed effect set
   in the top curve, no
   follower is in its leader's guessed effect set in the bottom curve,
   and followers are randomly assigned to guessed effect sets of the
   leaders in the middle curve. The two plots are without
   (left)  and with (right) macrolearning at 500 weeks.} 
   \label{fig:worstreward}
\end{figure}

Figure~\ref{fig:worstreward} presents averages over 50 runs of world reward
as a function of weeks using the worst-case reward matrix.  For
comparison purposes, in both plots the top curve represents the case
where the followers are in their leader's guessed effect sets. The
bottom curve in both plots represents the other extreme where no
leader's guessed effect set contains either of its followers. In both
plots, the middle curve is performance when the leaders' guessed
effect sets are initially random, both with (right) and without (left)
macrolearning turned on at week 500.

The performance for random guessed effect sets differs only slightly
from that of having leaders' guessed effect sets contain none of their
followers; both start with poor values of world reward that
deteriorates with time.  However, when macrolearning is performed on
systems with initially random guessed effect sets, the system quickly
rectifies itself and converges to optimal performance.  This is
reflected by the sudden vertical jump through the middle of the right
plot at 500 weeks, the point at which macrolearning changed the
guessed effect sets.  By changing those guessed effect sets
macrolearning results in a system that is factored for the associated WL
reward function, so that those reward functions quickly induced the
maximal possible world reward.

\begin{figure}[ht]
   \centerline{\mbox{
      \psfig{figure={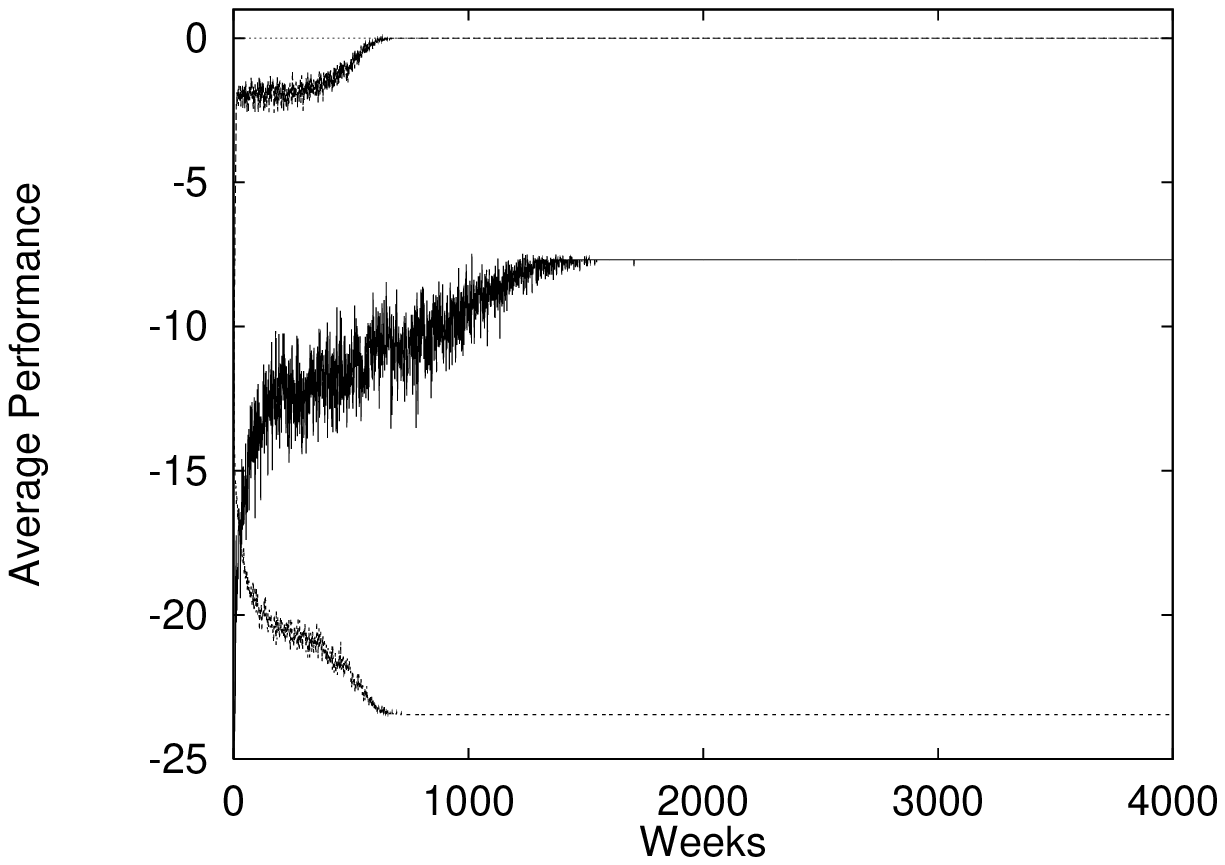},width=3.0in,height=2.25in}
      \hspace{0.1in}
      \psfig{figure={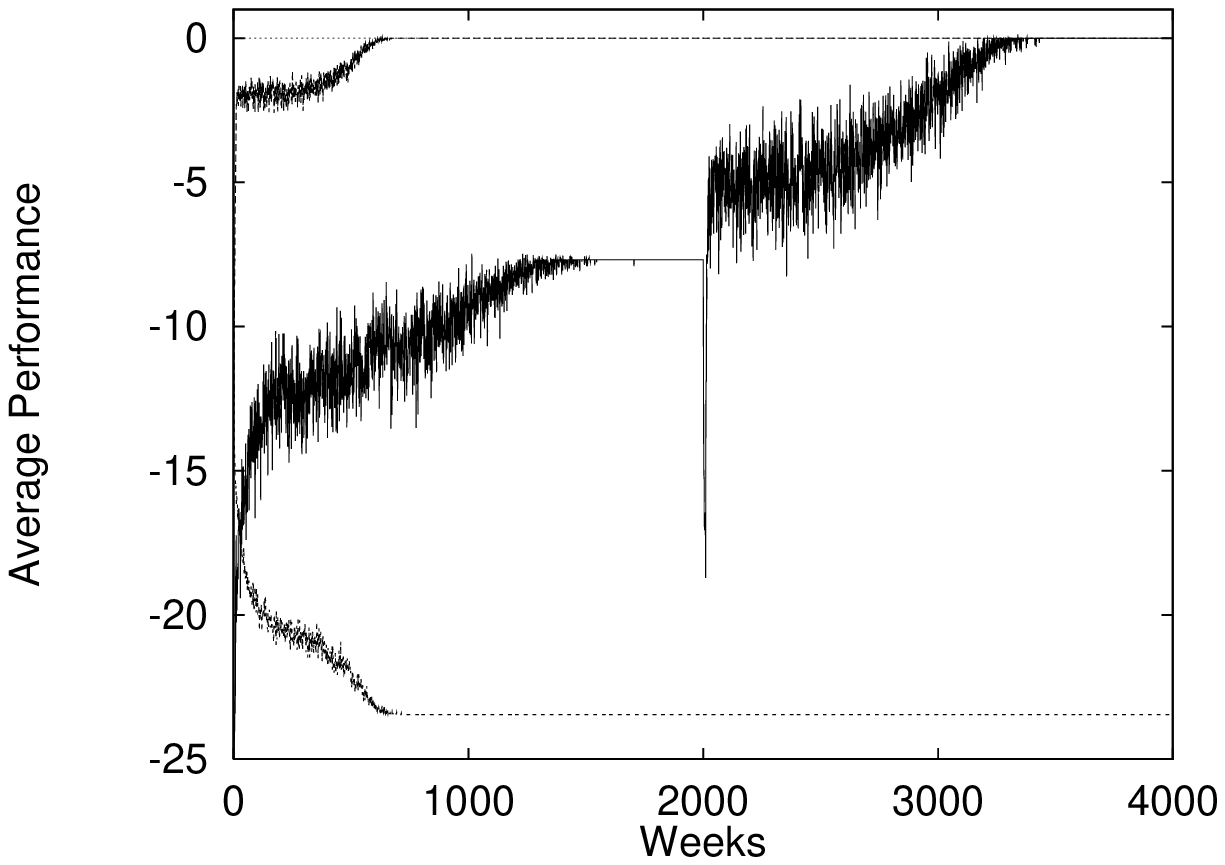},width=3.0in,height=2.25in}
   }}
   \caption{Leader-follower problem for random reward matrices. 
   The ordering of the plots is exactly as in Figure 4. 
   Macrolearning is applied at 2000 weeks, in the right plot.} 
   \label{fig:randomreward}
\end{figure}

Figure~\ref{fig:randomreward} presents performance averaged over 
50 runs for world reward as a function of weeks using a spectrum 
of reward matrices selected at random.  
The ordering of the plots is exactly as in Figure~\ref{fig:worstreward}.
Macrolearning is applied at 2000 weeks, in the right plot.
The simulations in Figure~\ref{fig:randomreward}  were lengthened from
those in Figure~\ref{fig:worstreward} because the convergence time of 
the full spectrum of reward matrices case was longer.

In figure~\ref{fig:randomreward} the macrolearning resulted in a
transient degradation in performance at 2000 weeks followed by
convergence to the optimal.  Without macrolearning the system's
performance no longer varied after 2000 weeks.  Combined with the
results presented in Figure~\ref{fig:worstreward}, these experiments
demonstrate that macrolearning induces optimal performance by aligning
the agents' guessed effect sets with those agents that they actually do
influence the most.

\section{CONCLUSION}
\label{sec:conc}
Many distributed computational tasks cannot be addressed by direct
modeling of the underlying dynamics, or are at best poorly addressed
that way due to robustness and scalability concerns. Such tasks should
instead be addressed by model-independent machine learning
techniques. In particular, Reinforcement Learning (RL) techniques are
often a natural choice for how to address such tasks. When --- as is
often the case --- we cannot rely on centralized control and
communication, such RL algorithms have to be deployed locally,
throughout the system.

This raises the important and profound question of how to configure
those algorithms, and especially their associated utility functions,
so as to achieve the (global) computational task. In particular we
must ensure that the RL algorithms do not ``work at cross-purposes''
as far as the global task is concerned, lest phenomena like tragedy of
the commons occur. How to initialize a system to do this is a
novel kind of inverse problem, and how to adapt a system at run-time
to better achieve such a global task is a novel kind of learning
problem. We call any distributed computational system analyzed from
the perspective of such an inverse problem a COllective INtelligence
(COIN).

As discussed in the literature review section of this chapter, there
are many approaches/fields that address aspects of COINs. These range
from multi-agent systems through conventional economics and on to
computational economics. (Human economies are a canonical model of a
functional COIN.) They range onward to game theory, various aspects of
distributed biological systems, and on through physics, active walker
models, and recurrent neural nets.  Unfortunately, none of these
fields seems appropriate as a general approach to understanding COINs.

After this literature review we present a mathematical theory for
COINs.  We then present experiments on two test problems that validate
the predictions of that theory for how best to design a COIN to
achieve a global computational task. The first set of experiments
involves a variant of Arthur's famous El Farol Bar problem.  The
second set instead considers a leader-follower problem that is
hand-designed to cause maximal difficulty for the advice of our theory
on how to initialize a COIN. This second set of experiments is
therefore a test of the on-line learning aspect of our approach to
COINs. In both experiments the procedures derived from our theory,
procedures using only local information, vastly outperformed natural
alternative approaches, even such approaches that exploited global
information.  Indeed, in both problems, following the theory
summarized in this chapter provides good solutions even when the exact
conditions required by the associated theorems hold only approximately.

There are many directions in which future work on COINs will proceed;
it is a vast and rich area of research. We are already successfully
applying our current understanding of COINs, tentative as it is, to
internet packet routing problems. We are also investigating COINs in a
more general optimization context where economics-inspired market
mechanisms are used to guide some of the interactions among the
agents of the distributed system. The goal in this second body of work
is to parallelize and solve numerical optimization problems where the
concept of an ``agent'' may not be in the natural definition of the
problem. We also intend to try to apply our current COIN framework to
the problem of designing high-occupancy toll lanes in vehicular
traffic, and to help understand the ``design space'' necessary for
distributed biochemical entities like pre-genomic cells.

\noindent{\bf Acknowledgements:} The authors would like to thank Ann
Bell, Michael New, Peter Norvig and Joe Sill for their comments.

\vspace{-.1in}

\bibliographystyle{plain}

\end{document}